\documentclass[preprint,3p,times,twocolumn]{elsarticle} 
\usepackage{epstopdf}
\usepackage{graphicx}
\usepackage{amssymb}
\usepackage{tabularx}
\usepackage{subcaption} 
\usepackage{amsmath}
\usepackage{pifont}
\usepackage{supertabular}
\usepackage{eurosym}
\usepackage{enumitem}
\usepackage{multirow}
\usepackage{appendix}
\usepackage{xcolor}
\newcommand{\forallt}{\ensuremath{\forall t \in \mathcal{T}}}
\newcommand{\forallw}{\ensuremath{\forall \omega \in \Omega}}
\newcommand{\PVgeneration}  {\ensuremath{y^\text{pv}}} 
\newcommand{\PVforecast} {\ensuremath{\hat{y}^\text{pv}}} 
\newcommand{\Wgeneration}  {\ensuremath{y^\text{w}}} 
\newcommand{\Wforecast}  {\ensuremath{\hat{y}^\text{w}}} 
\newcommand{\Load} {\ensuremath{y^\text{l}}} 
\newcommand{\Loadforecast} {\ensuremath{\hat{y}^\text{l}}} 
\newcommand{\Discharge} {\ensuremath{y^\text{dis}}} 
\newcommand{\Charge} {\ensuremath{y^\text{cha}}} 
\newcommand{\Soc} {\ensuremath{s}} 

\usepackage{color}

\usepackage[us]{datetime} 

\newcommand{\daddate}{\formatdate{6}{2}{2020}}

\newcount\myloopcounter
\newcommand{\nstars}[1][4]{%
  \myloopcounter0
  \loop\ifnum\myloopcounter < 3
  \ifthenelse{\myloopcounter < #1}{
    \textcolor{black}{\star}
  }{
    \textcolor{black!22}{\star}
  }
  \advance\myloopcounter by 1 %
  \repeat 
}

\journal{Applied Energy}

\usepackage[draft=False]{hyperref} 
\hypersetup{
     colorlinks=true,
     linkcolor=blue,
     filecolor=blue,
     citecolor=black,      
     urlcolor=cyan,
     }

\begin{document}

\begin{frontmatter}

\author[ADRESS]{Jonathan Dumas\corref{JDUMAS}}
\ead{jdumas@uliege.be}
\author[ADRESS]{Antoine Wehenkel}
\author[ADRESS2]{Damien Lanaspeze}
\author[ADRESS]{Bertrand Corn\'elusse}
\author[ADRESS]{Antonio Sutera}
\address[ADRESS]{Li\`ege University, Departments of Computer Science and Electrical Engineering, Belgium}
\address[ADRESS2]{Mines ParisTech, France}
\cortext[JDUMAS]{Corresponding author}

\title{A deep generative model for probabilistic energy forecasting in power systems: normalizing flows}

\begin{abstract}
Greater direct electrification of end-use sectors with a higher share of renewables is one of the pillars to power a carbon-neutral society by 2050. However, in contrast to conventional power plants, renewable energy is subject to uncertainty raising challenges for their interaction with power systems. Scenario-based probabilistic forecasting models have become a vital tool to equip decision-makers.
This paper presents to the power systems forecasting practitioners a recent deep learning technique, the \textit{normalizing flows}, to produce accurate scenario-based probabilistic forecasts that are crucial to face the new challenges in power systems applications. The strength of this technique is to directly learn the stochastic multivariate distribution of the underlying process by maximizing the likelihood.
Through comprehensive empirical evaluations using the open data of the Global Energy Forecasting Competition 2014, we demonstrate that this methodology is competitive with other state-of-the-art deep learning generative models: generative adversarial networks and variational autoencoders. 
The models producing weather-based wind, solar power, and load scenarios are properly compared in terms of forecast value by considering the case study of an energy retailer and quality using several complementary metrics.
The numerical experiments are simple and easily reproducible. Thus, we hope it will encourage other forecasting practitioners to test and use normalizing flows in power system applications such as bidding on electricity markets, scheduling power systems with high renewable energy sources penetration, energy management of virtual power plan or microgrids, and unit commitment.
\noindent
\end{abstract} 


\end{frontmatter}

\section{Introduction}\label{introduction}

To limit climate change and achieve the ambitious targets prescribed by the Intergovernmental Panel on Climate Change \citep{allen2019technical}, the transition towards a carbon-free society goes through an inevitable increase of the share of renewable generation in the energy mix. However, in contrast to conventional power plants, renewable energy is subject to uncertainty. Therefore, the operational predictability of modern power systems has been challenging as the total installed capacity of renewable energy sources (RES) increases and new distributed energy resources are introduced into the existing power networks. To address this challenge \textit{point} forecasts are widely used in the industry as inputs to decision-making tools. However, they are inherently uncertain and in the context of decision-making, a point forecast plus an uncertainty interval is of genuine added value. In this context, \textit{probabilistic} forecasts \citep{gneiting2014probabilistic}, which aim at modeling the distribution of all possible future realizations, have become an important tool \citep{hong2016probabilistic} to equip decision-makers \citep{morales2013integrating}, hopefully leading to better decisions in energy applications \citep{hong2020energy}.

The various types of probabilistic forecasts range from quantile to density forecasts, scenarios, and through prediction intervals \citep{morales2013integrating}. This paper focuses on \textit{scenario generation}, a popular probabilistic forecasting method to capture the uncertainty of load, photovoltaic (PV) generation, and wind generation. It consists of producing sequences of possible load or power generation realizations for one or more locations. 

Forecasting methodologies can typically be classified into two groups: \textit{statistical} and \textit{machine learning} models. On the one hand, statistical approaches are more interpretable than machine learning techniques, sometimes referred to as black-box models. On the other hand, they are generally more robust, user-friendly, and successful in addressing the non-linearity in the data than statistical techniques. We provide in the following a few examples of statistical approaches. More references can be found in \citet{khoshrou2019short} and \citet{mashlakov2021assessing}. 

Multiple linear regression models \citep{wang2016electric} and autoregressive integrated moving average \citep{de200625} are among the most fundamental and widely-used models. The latter generates spatiotemporal scenarios with given power generation profiles at each renewables generation site \citep{morales2010methodology}. These models mostly learn a relationship between several explanatory variables and a dependent target variable. However, they require some expert knowledge to formulate the relevant interaction between different variables. Therefore, the performance of such models is only satisfactory if the dependent variables are well formulated based on explanatory variables.
Another class of statistical approaches consists of using simple parametric distributions, \textit{e.g.}, the Weibull distribution for wind speed \citep{karaki2002probabilistic}, or the beta distribution for solar irradiance \citep{karaki1999probabilistic} to model the density associated with the generative process. In this line, the (Gaussian) copula method has been widely used to model the spatial and temporal characteristics of wind \citep{pinson2009probabilistic} and PV generation \citep{zhang2019coordinated}. For instance, the problem of generating probabilistic forecasts for the aggregated power of a set of renewable power plants harvesting different energy sources is addressed by \citet{camal2019scenario}.

Overall, these approaches usually make statistical assumptions increasing the difficulty to model the underlying stochastic process. The generated scenarios approximate the future uncertainty but cannot correctly describe all the salient features in the power output from renewable energy sources. \textit{Deep learning} is one of the newest trends in artificial intelligence and machine learning to tackle the limitations of statistical methods with promising results across various application domains.

\subsection{Related work}

Recurrent neural networks (RNNs) are among the most famous deep learning techniques adopted in energy forecasting applications. A novel pooling-based deep recurrent neural network is proposed by \citet{shi2017deep} in the field of short-term household load forecasting. It outperforms statistical approaches such as autoregressive integrated moving average and classical RNN. A tailored forecasting tool, named encoder-decoder, is implemented in \citet{dumas2020deep} to compute intraday multi-output PV quantiles forecasts.
Guidelines and best practices are developed by \citet{hewamalage2020recurrent} for forecasting practitioners on an extensive empirical study with an open-source software framework of existing RNN architectures. In the continuity, \citet{toubeau2018deep} implemented a bidirectional long short-term memory (BLSTM) architecture. It is trained using quantile regression and combined with a copula-based approach to generate scenarios. A scenario-based stochastic optimization case study compares this approach to other models regarding forecast quality and value. Finally, \citet{salinas2020deepar} trained an autoregressive recurrent neural network on several real-world datasets. It produces accurate probabilistic forecasts with little or no hyper-parameter tuning.


Deep generative modeling is a class of techniques that trains deep neural networks to model the distribution of the observations. In recent years, there has been a growing interest in this field made possible by the appearance of large open-access datasets and breakthroughs in both general deep learning architectures and generative models. Several approaches exist such as energy-based models, variational autoencoders, generative adversarial networks, autoregressive models, normalizing flows, and numerous hybrid strategies. They all make trade-offs in terms of computation time, diversity, and architectural restrictions. We recommend two papers to get a broader knowledge of this field. (1) The comprehensive overview of generative modeling trends conducted by \citet{bond2021deep}. It presents generative models to forecasting practitioners under a single cohesive statistical framework. (2) The thorough comparison of normalizing flows, variational autoencoders, and generative adversarial networks provided by \citet{ruthotto2021introduction}. It describes the advantages and disadvantages of each approach using numerical experiments in the field of computer vision. In the following, we focus on the applications of generative models in power systems.

In contrast to statistical approaches, deep generative models such as \textit{Variational AutoEncoders} (VAEs) \citep{kingma2013auto} and \textit{Generative Adversarial Networks} (GANs) \citep{goodfellow2014generative} directly learn a generative process of the data. They have demonstrated their effectiveness in many applications to compute accurate probabilistic forecasts including power system applications. 
They both make probabilistic forecasts in the form of Monte Carlo samples that can be used to compute consistent quantile estimates for all sub-ranges in the prediction horizon. Thus, they cannot suffer from the issue raised by \citet{ordiano2020probabilistic} on the non-differentiable quantile loss function. Note that generative models such as GANs and VAEs allow generating scenarios of the variable of interest directly. In contrast with methods that first compute weather scenarios to generate probabilistic forecasts such as implemented by \citet{sun2020probabilistic} and \citet{khoshrou2019short}.
A VAE composed of a succession of convolutional and feed-forward layers is proposed by \citet{zhanga2018optimized} to capture the spatial-temporal complementary and fluctuant characteristics of wind and PV power with high model accuracy and low computational complexity. Both single and multi-output PV forecasts using a VAE are compared by \citet{dairi2020short} to several deep learning methods such as LSTM, BLSTM, convolutional LSTM networks and stacked autoencoders, where the VAE consistently outperformed the other methods.
A GAN is used by \citet{chen2018model} to produce a set of wind power scenarios that represent possible future behaviors based only on historical observations and point forecasts. This method has a better performance compared to Gaussian Copula. A Bayesian GAN is introduced by \citet{chen2018bayesian} to generate wind and solar scenarios, and a progressive growing of GANs is designed by \citet{yuan2021multi} to propose a novel scenario forecasting method.
In a different application, a GAN is implemented for building occupancy modeling without any prior assumptions \citep{chen2018building}. Finally, a conditional version of the GAN using several labels representing some characteristics of the demand is introduced by \citet{lan2018demand} to output power load data considering demand response programs.

Improved versions of GANs and VAEs have also been studied in the context of energy forecasting. The Wasserstein GAN enforces the Lipschitz continuity through a gradient penalty term (WGAN-GP), as the original GANs are challenging to train and suffer from mode collapse and over-fitting. Several studies applied this improved version in power systems: (1) a method using unsupervised labeling and conditional WGAN-GP models the uncertainties and variation in wind power \citep{zhang2020typical}; (2) a WGAN-GP models both the uncertainties and the variations of the load \citep{wang2020modeling}; (3) \citet{jiang2021day} implemented scenario generation tasks both for a single site and for multiple correlated sites without any changes to the model structure.
Concerning VAEs, they suffer from inherent shortcomings, such as the difficulties of tuning the hyper-parameters or generalizing a specific generative model structure to other databases. An improved VAE is proposed by~\citet{qi2020optimal} with the implementation of a $\beta$ hyper-parameter into the VAE objective function to balance the two parts of the loss. This improved VAE is used to generate PV and power scenarios from historical values.

However, most of these studies did not benefit from conditional information such as weather forecasts to generate improved PV, wind power, and load scenarios. In addition, to the best of our knowledge,
only~\citet{ge2020modeling} compared NFs to these techniques for the generation of daily load profiles. Nevertheless, the comparison only considers quality metrics, and the models do not incorporate weather forecasts.

\subsection{Research gaps and scientific contributions}

This study investigates the implementation of \textit{Normalizing Flows} \citep[NFs]{rezende2015variational} in power system applications. NFs define a new class of probabilistic generative models. They have gained increasing interest from the deep learning community in recent years. A NF learns a sequence of transformations, a \textit{flow}, from a density known analytically, \textit{e.g.}, a \textit{Normal} distribution, to a complex target distribution. In contrast to other deep generative models, NFs can directly be trained by maximum likelihood estimation. They have proven to be an effective way to model complex data distributions with neural networks in many domains. First, speech synthesis \citep{oord2018parallel}. Second, fundamental physics to increase the speed of gravitational wave inference by several orders of magnitude \citep{green2021complete} or for sampling Boltzmann distributions of lattice field theories \citep{albergo2021introduction}. Finally, in the capacity firming framework by \citet{dumas2021probabilistic}.

This present work goes several steps further than~\citet{ge2020modeling} that demonstrated the competitiveness of NFs regarding GANs and VAEs for generating daily load profiles. First, we study the conditional version of these models to demonstrate that they can handle additional contextual information such as weather forecasts or geographical locations. Second, we extensively compare the model's performances both in terms of forecast value and quality. The forecast quality corresponds to the ability of the forecasts to genuinely inform of future events by mimicking the characteristics of the processes involved. The forecast value relates to the benefits of using forecasts in decision-making, such as participation in the electricity market. Third, we consider PV and wind generations in addition to load profiles. Finally, in contrast to the affine NFs used in their work, we rely on monotonic transformations, which are universal density approximators~\citep{huang2018neural}.\\

Given that Normalizing Flows are rarely used in the power systems community despite their potential, our main aim is to present this recent deep learning technique and demonstrate its interest and competitiveness with state-of-the-art generative models such as GANs and VAEs on a simple and easily reproducible case study. The research gaps motivating this paper are three-fold:
\begin{enumerate}
	\item To the best of our knowledge, only \citet{ge2020modeling} compared NFs to GANs and VAEs for the generation of daily load profiles. Nevertheless, the comparison is only based on quality metrics, and the models do not take into account weather forecasts; 
	\item Most of the studies that propose or compare forecasting techniques only consider the forecast quality such as \citet{ge2020modeling}, \citet{sun2020probabilistic}, and \citet{mashlakov2021assessing};
	\item The conditional versions of the models are not always addressed, such as in \citet{ge2020modeling}. However, weather forecasts are essential for computing accurate probabilistic forecasts.
\end{enumerate}

\noindent With these research gaps in mind, the main contributions of this paper are three-fold:
\begin{enumerate}
	\item We provide a fair comparison both in terms of quality and value with the state-of-the-art deep learning generative models, GANs and VAEs, using the open data of the Global Energy Forecasting Competition 2014 (GEFcom 2014) \citep{hong2016bprobabilistic}. To the best of our knowledge, it is the first study that extensively compares the NFs, GANs, and VAEs on several datasets, PV generation, wind generation, and load with a proper assessment of the quality and value based on complementary metrics, and an easily reproducible case study; 
	\item We implement conditional generative models to compute improved weather-based PV, wind power, and load scenarios. In contrast to most of the previous studies that focused mainly on past observations;
	\item Overall, we demonstrate that NFs are more accurate in quality and value, providing further evidence for deep learning practitioners to implement this approach in more advanced power system applications.
\end{enumerate}

\begin{figure}[htbp]
	\centering
	\includegraphics[width=\linewidth]{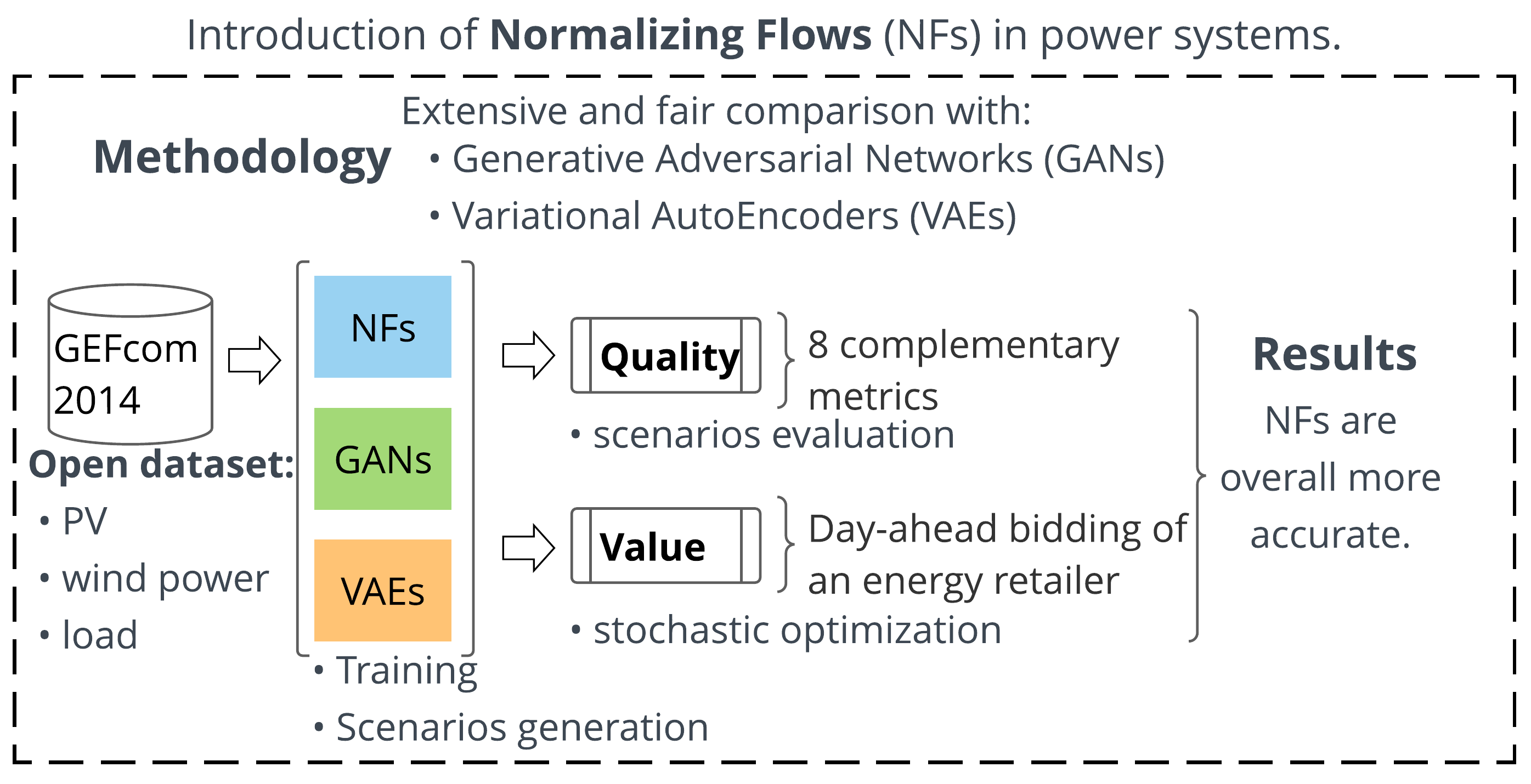}
	\caption{The framework of the paper. \\
	The paper's primary purpose is to present and demonstrate the potential of NFs in power systems. A fair comparison is conducted both in terms of quality and value with the state-of-the-art deep learning generative models, GANs and VAEs, using the open data of the Global Energy Forecasting Competition 2014 \citep{hong2016bprobabilistic}. The PV, wind power, and load datasets are used to assess the models. The quality evaluation is conducted by using eight complementary metrics, and the value assessment by considering the day-ahead bidding of an energy retailer using stochastic optimization. Overall, NFs tend to be more accurate both in terms of quality and value and are competitive with GANs and VAEs.}
	\label{fig:paper-framework}
\end{figure}

In addition to these contributions, this study also provides open-access to the Python code\footnote{\url{https://github.com/jonathandumas/generative-models}} to help the community to reproduce the experiments.
Figure \ref{fig:paper-framework} provides the framework of the proposed method and Table \ref{tab:contributions} presents a comparison of the present study to three state-of-the-art papers using deep learning generative models to generate scenarios.
\begin{table}[htbp]
\renewcommand{\arraystretch}{1.25}
	\begin{center}
		\begin{tabular}{lcccc}
			\hline \hline
Criteria			     & \citep{wang2020modeling}  &  \citep{qi2020optimal}  & \citep{ge2020modeling} & study \\ \hline
GAN                      & \checkmark & $\times$   & \checkmark & \checkmark \\
VAE                      & $\times$   & \checkmark & \checkmark & \checkmark   \\
NF                       & $\times$   & $\times$   & \checkmark & \checkmark   \\
Number of models         & 4          & 1          &  3         & 3  \\
PV 	                     & $\times$   & \checkmark & $\times$   & \checkmark  \\
Wind power 	             & $\times$   & \checkmark & $\times$   & \checkmark \\
Load 	                 & \checkmark & $\sim$     & \checkmark & \checkmark    \\
Weather-based            & \checkmark & $\times$   & $\times$   & \checkmark    \\
Quality assessment 	     & \checkmark & \checkmark & \checkmark & \checkmark   \\
Quality metrics  	     & 5          & 3          &   5        & 8 \\
Value assessment     	 & $\times$   & \checkmark & $\times$   & \checkmark \\
Open dataset             & $\sim$     & $\times$   & \checkmark & \checkmark \\
Value replicability      & -          & $\sim$     &  -         & \checkmark \\
Open-access code         & $\times$   & $\times$   &  $\times$  & \checkmark \\
\hline \hline
\end{tabular}
\caption{Comparison of the paper's contributions to three state-of-the-art studies using deep generative models. \\
\checkmark: criteria fully satisfied, $\sim$: criteria partially satisfied, $\times$: criteria not satisfied, ?: no information, -: not applicable. 
GAN: a GAN model is implemented; VAE: a VAE model is implemented; NF: a NF model is implemented; PV: PV scenarios are generated; Wind power: wind power scenarios are generated; Load: load scenarios are generated; Weather-based: the model generates weather-based scenarios; Quality assessment: a quality evaluation is conducted: Quality metrics: number of quality metrics considered; Value assessment: a value evaluation is considered with a case study; Open dataset: the data used for the quality and value evaluations are in open-access; Value replicability: the case study considered for the value evaluation is easily reproducible; Open-access code: the code used to conduct the experiments is in open-access.
Note: the justifications are provided in \ref{annex:table1}.}
\label{tab:contributions}
\end{center}
\end{table}

\subsection{Applicability of the generative models}
Probabilistic forecasting of PV, wind generation, electrical consumption, and electricity prices plays a vital role in renewable integration and power system operations. The deep learning generative models presented in this paper can be integrated into practical engineering applications. We present a non-exhaustive list of five applications in the following. (1) The forecasting module of an energy management system (EMS) \citep{silva2021optimal}. Indeed, EMSs are used by several energy market players to operate various power systems such as a single renewable plant, a grid-connected or off-grid microgrid composed of several generations, consumption, and storage devices. An EMS is composed of several key modules: monitoring, forecasting, planning, control, \textit{etc}. The forecasting module aims to provide the most accurate forecast of the variable of interest to be used as inputs of the planning and control modules. (2) Stochastic unit commitment models that employ scenarios to model the uncertainty of weather-dependent renewables. For instance, the optimal day-ahead scheduling and dispatch of a system composed of renewable plants, generators, and electrical demand are addressed by \citet{camal2019scenario}. (3) Ancillary services market participation. A virtual power plant aggregating wind, PV, and small hydropower plants is studied by \citet{camal2019scenario} to optimally bid on a day-ahead basis the energy and automatic frequency restoration reserve. (4) More generally, generative models can be used to compute scenarios for any variable of interest, \textit{e.g.}, energy prices, renewable generation, loads, water inflow of hydro reservoirs, as long as data are available. (5) Finally, quantiles can be derived from scenarios and used in robust optimization models such as in the capacity firming framework \citep{dumas2021probabilistic}. 

\subsection{Organization}

The remainder of this paper is organized as follows. Section~\ref{sec:background} presents the generative models implemented: NFs, GANs, and VAEs. Section~\ref{sec:quality_assessment} provides the quality and value assessment methodologies. Section~\ref{sec:numerical_results} details empirical results on the GEFcom 2014 dataset, and Section~\ref{sec:conclusion} summarizes the main findings and highlights ideas for further work. 
\ref{annex:arguments} presents the justifications of Table \ref{tab:contributions} and Table \ref{tab:comparison}, \ref{annex:background} provides additional information on the generative models, \ref{annex:quality-assessment} and \ref{annex:value-assessment} detail the quality metrics and the retailer energy case study formulation, and \ref{annex:assessment_results} presents additional quality results.

\section{Background}\label{sec:background}

This section formally introduces the conditional version of NFs, GANs, and VAEs implemented in this study. We assume the reader is familiar with the neural network's basics. However, for further information, \citet{goodfellow2016deep,zhang2020dive} provide a comprehensive introduction to modern deep learning approaches.

\subsection{Multi-output forecast}

Let us consider some dataset $\mathcal{D} = \{ \mathbf{x}^i, \mathbf{c}^i \}_{i=1}^N$ of $N$ independent and identically distributed samples from the joint distribution $p(\mathbf{x},\mathbf{c})$ of two continuous variables $X$ and $C$. $X$ being the wind generation, PV generation, or load, and $C$ the weather forecasts. They are both composed of $T$ periods per day, with $\mathbf{x}^i := [x_1^i, \ldots , x_T^i]^\intercal \in \mathbb{R}^T$ and $\mathbf{c}^i := [c_1^i, \ldots , c_T^i]^\intercal \in \mathbb{R}^T$. The goal of this work is to generate multi-output weather-based scenarios $\mathbf{\hat{x}} \in \mathbb{R}^T$ that are distributed under $p(\mathbf{x}|\mathbf{c})$.

A generative model is a probabilistic model $p_\theta(\cdot)$, with parameters $\theta$, that can be used as a generator of the data. Its purpose is to generate synthetic but realistic data $\mathbf{\hat{x}} \sim p_\theta(\mathbf{x}|\mathbf{c})$ whose distribution is as close as possible to the unknown data distribution $p(\mathbf{x}|\mathbf{c})$. In our application, it computes on a day-ahead basis a set of $M$ scenarios at day $d-1$ for day $d$
\begin{align}
	\label{eq:multi_output_scenario}	
	\mathbf{\hat{x}}_d^i := &  \big[\hat{x}_{d, 1}^i, \cdots,\hat{x}_{d, T}^i\big]^\intercal	\in \mathbb{R}^T \quad i=1, \ldots, M.
\end{align}
For the sake of clarity, we omit the indexes $d$ and $i$ when referring to a scenario $\mathbf{\hat{x}}$ in the following.

\subsection{Deep generative models}

\begin{figure}[htbp]
	\centering
	\includegraphics[width=\linewidth]{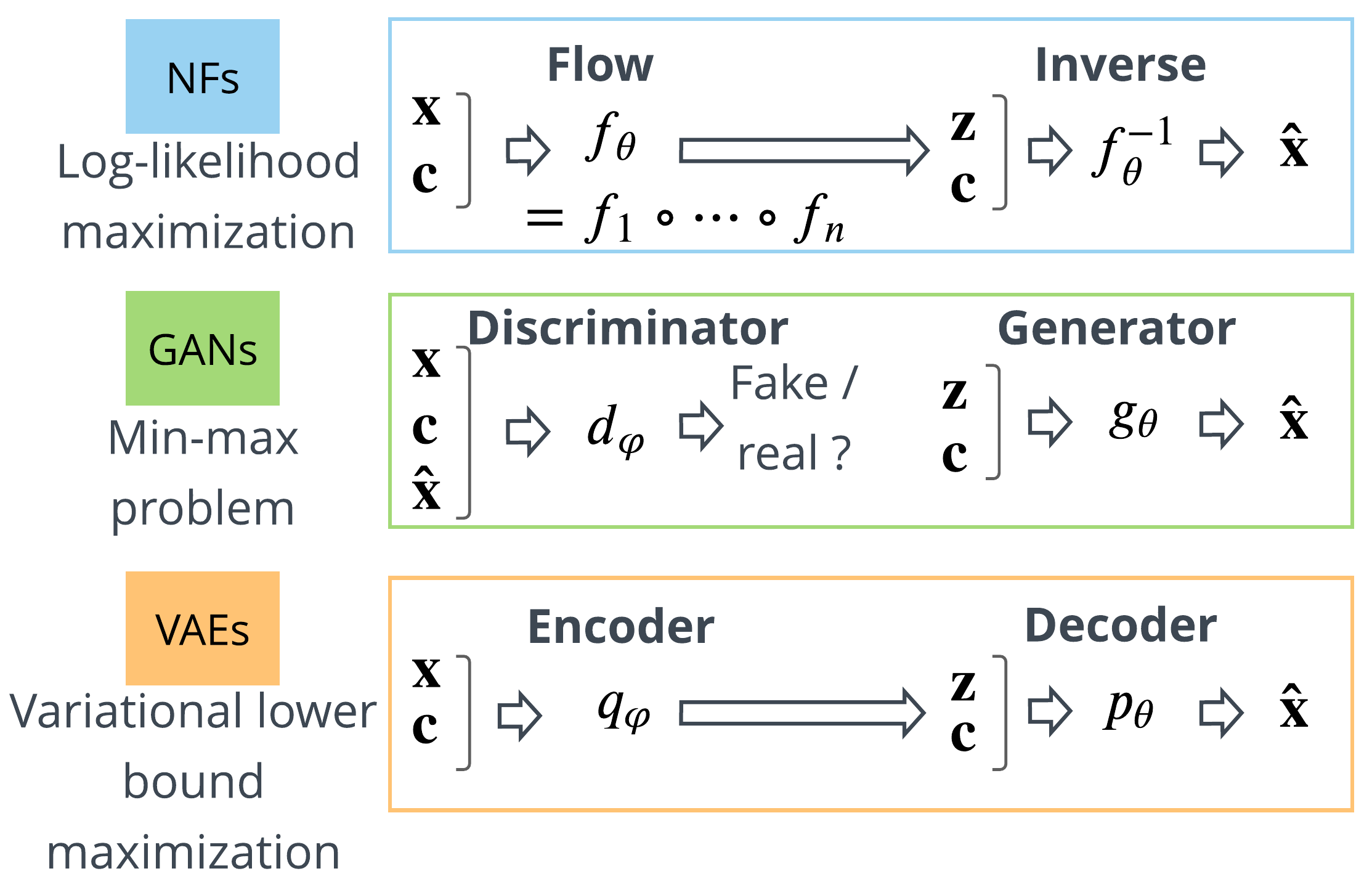}
	\caption{High-level comparison of three categories of generative models considered in this paper: normalizing flows, generative adversarial networks, and variational autoencoders. \\
	All models are conditional as they use the weather forecasts $\mathbf{c}$ to generate scenarios $\mathbf{\hat{x}}$ of the distribution of interest $\mathbf{x}$: PV generation, wind power, load.
	Normalizing flows allow exact likelihood calculation. In contrast to generative adversarial networks and variational autoencoders, they explicitly learn the data distribution and directly access the exact likelihood of the model's parameters. The inverse of the flow is used to generate scenarios.
	The training of generative adversarial networks relies on a min-max problem where the generator and the discriminator parameters are jointly optimized. The generator is used to compute the scenarios.
	Variational autoencoders indirectly optimize the log-likelihood of the data by maximizing the variational lower bound. The decoder computes the scenarios.
	Note: Section \ref{sec:comparison} provides a theoretical comparison of these models. 
	}
	\label{fig:methods-comparison}
\end{figure}

Figure \ref{fig:methods-comparison} provides a high-level comparison of three categories of generative models considered in this paper: Normalizing Flows, Generative Adversarial Networks, and Variational AutoEncoders.

\subsubsection{Normalizing flows}\label{sec:nfs_description}

A normalizing flow is defined as a sequence of invertible transformations $f_k : \mathbb{R}^T  \rightarrow \mathbb{R}^T$, $k = 1, \ldots, K$, composed together to create an expressive invertible mapping $f_\theta := f_1 \circ \ldots  \circ f_K : \mathbb{R}^T  \rightarrow \mathbb{R}^T$. This composed function can be used to perform density estimation, using $f_\theta$ to map a sample $\mathbf{x} \in  \mathbb{R}^T $ onto a latent vector $\mathbf{z} \in  \mathbb{R}^T $ equipped with a known and tractable probability density function $p_z$, \textit{e.g.}, a Normal distribution. The transformation $f_\theta$ implicitly defines a density $p_\theta(\mathbf{x})$ that is given by the change of variables
\begin{align}
\label{eq:change_formula}	
p_\theta(\mathbf{x})  & = p_z(f_\theta(\mathbf{x}))|\det J_{f_\theta}(\mathbf{x})| , 
\end{align}
where $ J_{f_\theta}$ is the Jacobian of $f_\theta$ regarding $\mathbf{x}$. The model is trained by maximizing the log-likelihood $\sum_{i=1}^N \log p_\theta(\mathbf{x}^i, \mathbf{c}^i)$ of the model's parameters $\theta$ given the dataset $\mathcal{D}$. For simplicity let us assume a single-step flow $f_\theta$ to drop the index $k$ for the rest of the discussion.

In general, $f_\theta$ can take any form as long as it defines a bijection. However, a common solution to make the Jacobian computation tractable in (\ref{eq:change_formula}) consists of implementing an \textit{autoregressive} transformation \citep{kingma2016improving}, \textit{i.e.}, such that $f_\theta$ can be rewritten as a vector of scalar bijections $f^i$
\begin{subequations}
\label{eq:f_autoregressive}	
\begin{align}
	f_\theta(\mathbf{x}) & := [f^1(x_1; h^1), \ldots, f^T(x_T; h^T)]^\intercal , \\
	h^i & := h^i (\mathbf{x}_{<i} ; \varphi^i) \quad 2\leq i\leq T, \\
	\mathbf{x}_{<i} & := [x_1, \ldots, x_{i-1}]^\intercal  \quad 2\leq i\leq T, \\
	h^1 & \in  \mathbb{R},
\end{align}
\end{subequations}
where $f^i(\cdot; h^i ) : \mathbb{R} \rightarrow \mathbb{R}$ is partially parameterized by an autoregressive conditioner $h^i(\cdot; \varphi^i): \mathbb{R}^{i-1} \rightarrow \mathbb{R}^{|h^i|}$ with parameters $\varphi^i$, and $\theta$ the union of all parameters $\varphi^i$. 

There is a large choice of transformers $f^i$: affine, non-affine, integration-based, \textit{etc}. This work implements an integration-based transformer by using the class of Unconstrained Monotonic Neural Networks (UMNN) \citep{wehenkel2019unconstrained}. It is a universal density approximator of continuous random variables when combined with autoregressive functions.
The UMNN consists of a neural network architecture that enables learning arbitrary monotonic functions. It is achieved by parameterizing the bijection $f^i$ as follows
\begin{align}
\label{eq:UMNN_bijective_mapping}	
	f^i(x_i; h^i ) & =\int_0^{x_i} \tau^i(x_i, h^i) dt + \beta^i(h^i),
\end{align}
where $\tau^i(\cdot; h^i) : \mathbb{R}^{|h^i|+1}  \rightarrow \mathbb{R}^+$ is the integrand neural network with a strictly positive scalar output, $h^i \in  \mathbb{R}^{|h^i|} $ an embedding made by the conditioner, and $ \beta^i(\cdot) : \mathbb{R}^{|h^i|}  \rightarrow \mathbb{R}$ a neural network with a scalar output. The forward evaluation of $f^i$ requires solving the integral (\ref{eq:UMNN_bijective_mapping}) and is efficiently approximated numerically by using the Clenshaw-Curtis quadrature. The pseudo-code of the forward and backward passes is provided by \citet{wehenkel2019unconstrained}.

%
%
%

\begin{figure}[htbp]
	\centering
	\includegraphics[width=\linewidth]{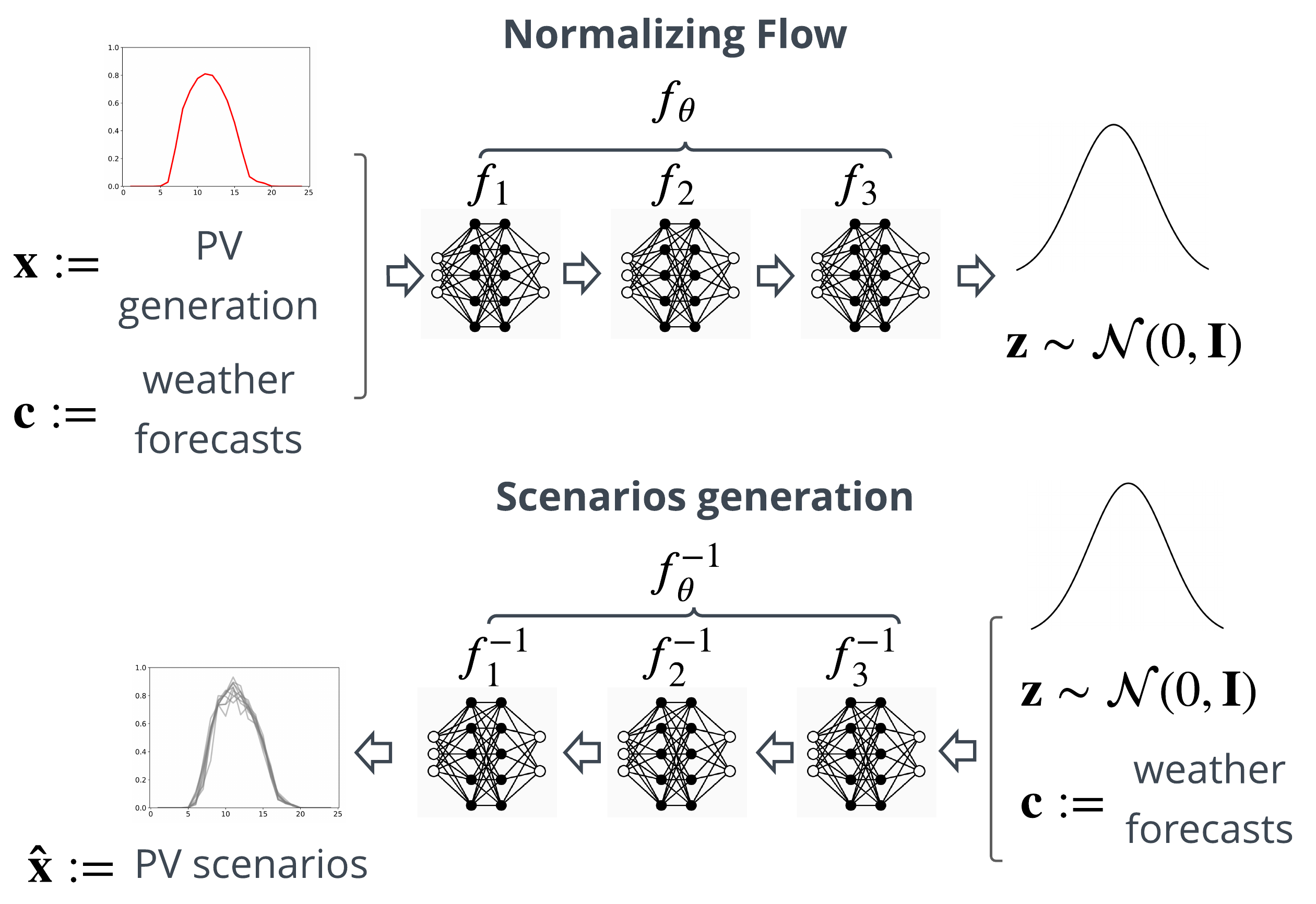}
	\caption{The process of conditional normalizing flows is illustrated with a three-step NF for PV generation. \\
	The model $f_\theta$ is trained by maximizing the log-likelihood of the model's parameters $\theta$ given a dataset composed of PV observations and weather forecasts. Recall $f_\theta$ defines a bijection between the variable of interest $\mathbf{x}$, PV generation, and a Normal distribution $\mathbf{z}$. Then, the PV scenarios $\mathbf{\hat{x}}$ are generated by using the inverse of $f_\theta$ that takes as inputs samples from the Normal distribution $\mathbf{z}$ and the weather forecasts $\mathbf{c}$.}
	\label{fig:UMNN_structure}
\end{figure}

\citet{papamakarios2017masked}'s Masked Autoregressive Network (MAF) is implemented to simultaneously parameterize the $T$ autoregressive embeddings $h^i$ of the flow (\ref{eq:f_autoregressive}). Then, the change of variables formula applied to the UMMN-MAF transformation results in the following log-density when considering weather forecasts
\begin{subequations}
\label{eq:UMNN_likelihood_estimation}
\begin{align}
\log p_\theta(\mathbf{x}, \mathbf{c}) & = \log p_z(f_\theta(\mathbf{x}, \mathbf{c}))|\det J_{f_\theta}(\mathbf{x}, \mathbf{c})| , \\
 & = \log p_z(f_\theta(\mathbf{x}, \mathbf{c})) + \sum_{i=1}^T \log \tau^i\big(x_i, h^i(\mathbf{x}_{<i}), \mathbf{c}\big),
\end{align}
\end{subequations}
that can be computed exactly and efficiently with a single forward pass. The UMNN-MAF approach implemented is referred to as NF in the rest of the paper. Figure~\ref{fig:UMNN_structure} depicts the process of conditional normalizing flows with a three-step NF for PV generation. Note: \ref{annex:nf} provides additional information on NFs.

\subsubsection{Variational autoencoders}\label{sec:vae}

\begin{figure}[htbp]
	\centering
	\includegraphics[width=\linewidth]{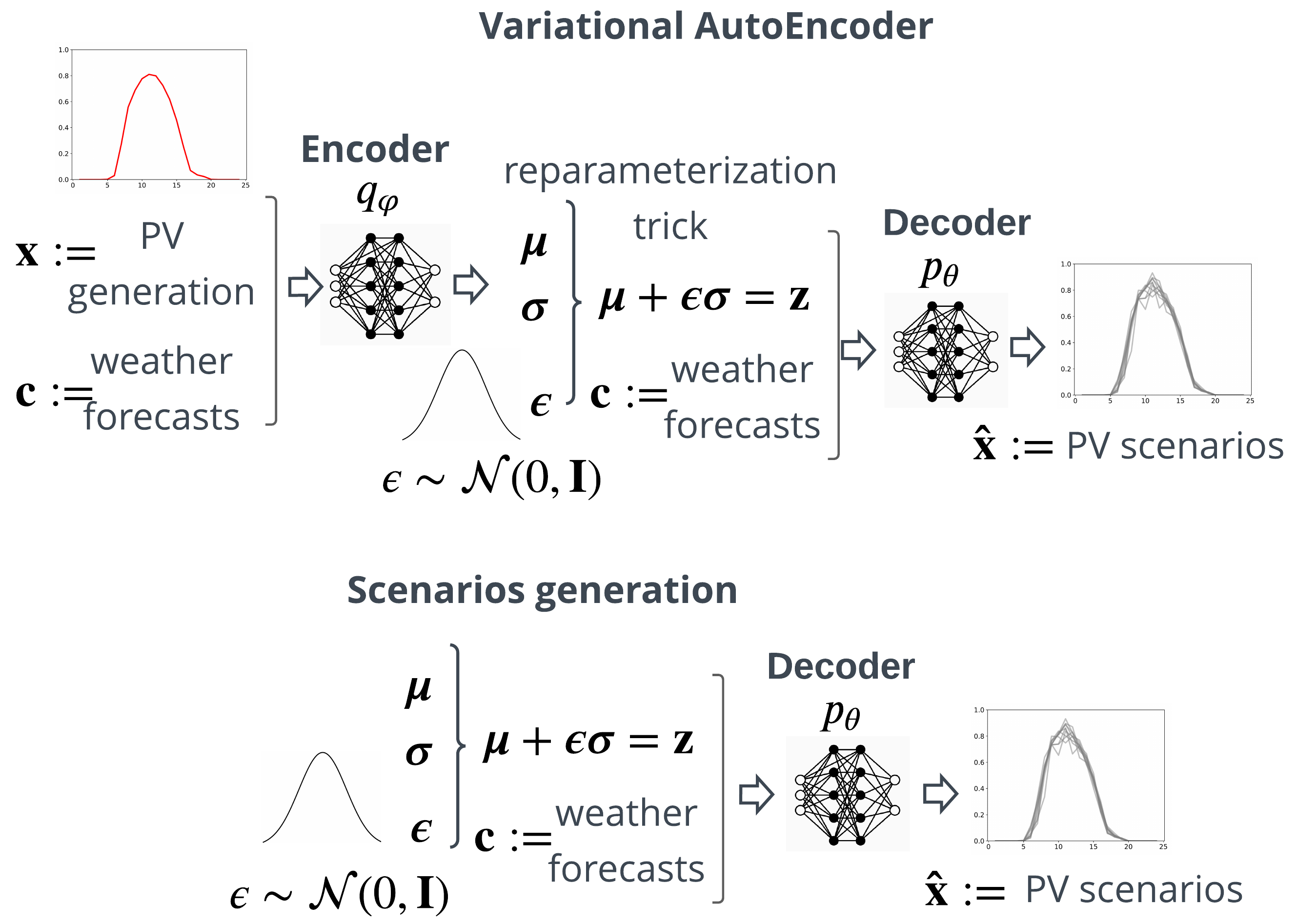}
	\caption{The process of the conditional variational autoencoder is illustrated for PV generation. \\
	The VAE is trained by maximizing the variational lower bound given a dataset composed of PV observations and weather forecasts. The encoder $q_\varphi$ maps the variable of interest $\mathbf{x}$ to a latent space $\mathbf{z}$. The decoder $p_\theta$ generates the PV scenarios $\mathbf{\hat{x}}$ by taking as inputs samples $\mathbf{z}$ from the latent space and the weather forecasts $\mathbf{c}$.}
	\label{fig:VAE_structure}
\end{figure}

A VAE is a deep latent variable model composed of an \textit{encoder} and a \textit{decoder} which are jointly trained to maximize a lower bound on the likelihood. The encoder $q_\varphi(\cdot) : \mathbb{R}^T \times \mathbb{R}^{|\mathbf{c}|}  \rightarrow \mathbb{R}^d$ approximates the intractable posterior $p(\mathbf{z}|\mathbf{x}, \mathbf{c})$, and the decoder $p_\theta(\cdot) : \mathbb{R}^d \times \mathbb{R}^{|\mathbf{c}|} \rightarrow \mathbb{R}^T $ the likelihood $p(\mathbf{x}|\mathbf{z}, \mathbf{c})$ with $\mathbf{z} \in \mathbb{R}^d$.
Maximum likelihood is intractable as it would require marginalizing with respect to all possible realizations of the latent variables $\mathbf{z}$. \citet{kingma2013auto} addressed this issue by maximizing the \textit{variational lower bound} $\mathcal{L}_{\theta, \varphi}(\mathbf{x, c})$ as follows
\begin{subequations}
\label{eq:variational_lower_bound}	
\begin{align}
\log p_\theta(\mathbf{x}|\mathbf{c}) = & KL[ q_\varphi(\mathbf{z}|\mathbf{x}, \mathbf{c}) || p(\mathbf{z}|\mathbf{x},\mathbf{c})] +\mathcal{L}_{\theta, \varphi}(\mathbf{x, c}) , \\
\geq & \mathcal{L}_{\theta, \varphi}(\mathbf{x, c}), \\
\mathcal{L}_{\theta, \varphi}(\mathbf{x, c}) := & \mathbb{E}_{q_\varphi(\mathbf{z}|\mathbf{x}, \mathbf{c})}\left[\log\frac{p(\mathbf{z}) p_\theta(\mathbf{x}|\mathbf{z}, \mathbf{c})}{q_\varphi(\mathbf{z}|\mathbf{x}, \mathbf{c})}\right],
\end{align}
\end{subequations}
as the Kullback-Leibler (KL) divergence \citep{perez2008kullback} is non-negative. \ref{annex:vae} details how to compute the gradients of $\mathcal{L}_{\theta, \varphi}(\mathbf{x, c})$, and its exact expression for the implemented VAE composed of fully connected neural networks for both the encoder and decoder. Figure~\ref{fig:VAE_structure} depicts the process of a conditional variational autoencoder for PV generation.

\subsubsection{Generative adversarial networks}\label{sec:gan}

\begin{figure}[htbp]
	\centering
	\includegraphics[width=\linewidth]{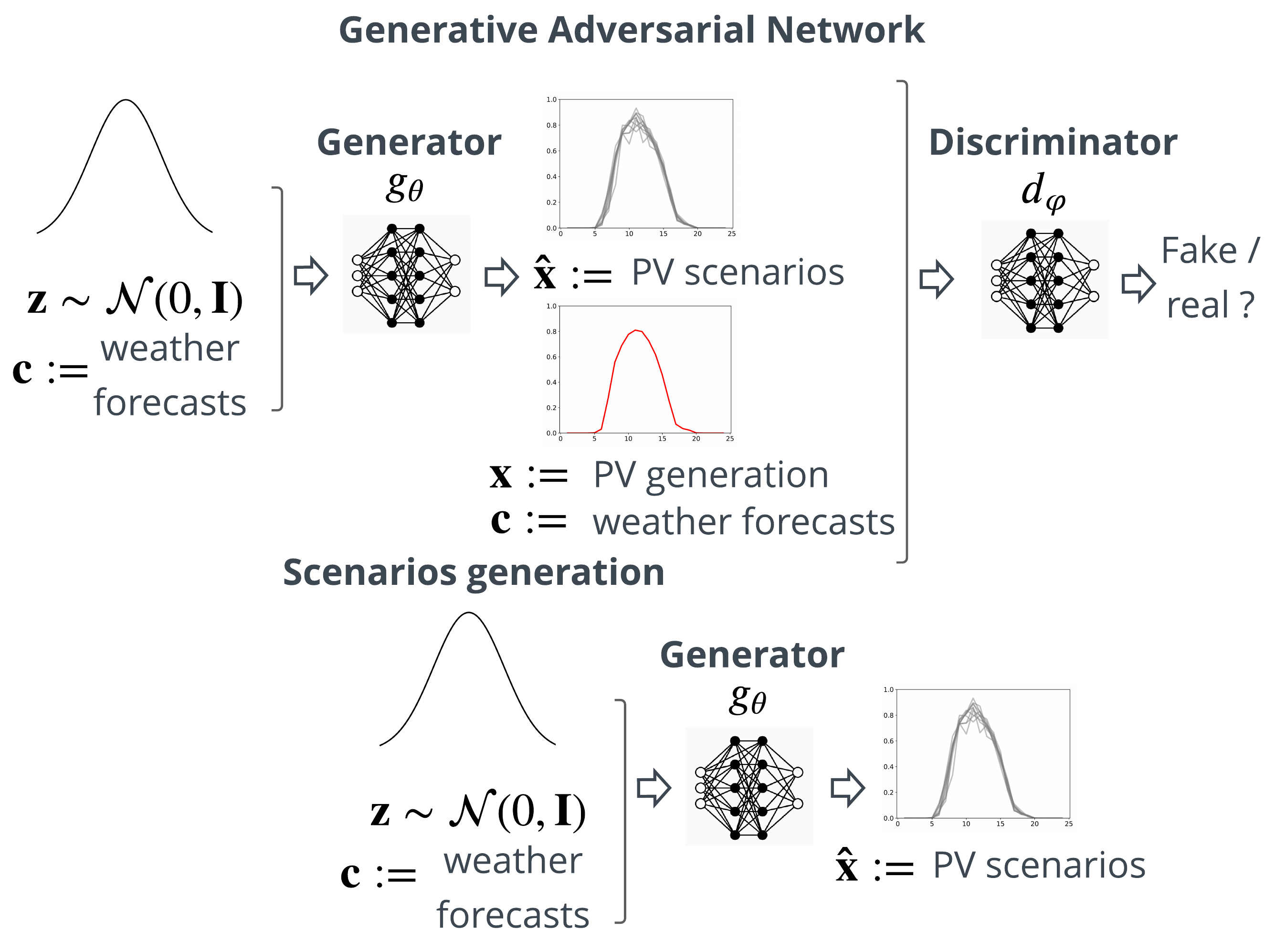}
	\caption{The process of the conditional generative adversarial network is illustrated for PV generation. \\
	The GAN is trained by solving a min-max problem given a dataset composed of PV observations $\mathbf{x}$ and weather forecasts. The generator $g_\theta$ computes PV scenarios $\mathbf{\hat{x}}$ by taking as inputs samples from the Normal distribution $\mathbf{z}$ and the weather forecasts $\mathbf{c}$, and the decoder $d_\phi$ tries to distinguishes true data from scenarios.}
	\label{fig:GAN_structure}
\end{figure}

GANs are a class of deep generative models proposed by \citet{goodfellow2014generative} where the key idea is the adversarial training of two neural networks, the \textit{generator} and the \textit{discriminator}, during which the generator learns iteratively to produce realistic scenarios until they cannot be distinguished anymore by the discriminator from real data. The generator $g_\theta(\cdot) : \mathbb{R}^d \times \mathbb{R}^{|\mathbf{c}|} \rightarrow \mathbb{R}^T $ maps a latent vector $\mathbf{z} \in \mathbb{R}^d $ equipped with a known and tractable prior probability density function $p(\mathbf{z})$, \textit{e.g.}, a Normal distribution, onto a sample $\mathbf{x} \in  \mathbb{R}^T $,  and is trained to fool the discriminator. The discriminator $d_\phi(\cdot) :\mathbb{R}^T\times \mathbb{R}^{|\mathbf{c}|} \rightarrow [0, 1]$ is a classifier trained to distinguish between true samples $\mathbf{x}$ and generated samples $\mathbf{\hat{x}}$. \citet{goodfellow2014generative} demonstrated that solving the following min-max problem
\begin{align}
\label{eq:WGAN_GP}	
 \theta^\star = & \arg \min_\theta \max_\phi V(\phi, \theta),
\end{align}
where $V(\phi, \theta)$ is the value function, recovers the data generating distribution if $g_\theta(\cdot)$ and $d_\phi(\cdot)$ are given enough capacity. The state-of-the-art conditional Wasserstein GAN with gradient penalty (WGAN-GP) proposed by \citet{gulrajani2017improved} is implemented with $V(\phi, \theta)$ defined as
\begin{subequations}
\label{eq:WGAN-GP}	
\begin{align}
V(\phi, \theta) = &   - \bigg(\underset{\mathbf{\hat{x}}}{\mathbb{E}} [ d_\phi (\mathbf{\hat{x}}|\mathbf{c}) ] -\underset{\mathbf{x} }{\mathbb{E}} [ d_\phi (\mathbf{x}|\mathbf{c}) ]  + \lambda \text{GP} \bigg) , \\
\text{GP} = & \underset{\mathbf{\tilde{x}}}{\mathbb{E}}  \bigg[ \bigg( \Vert \nabla_{\mathbf{\tilde{x}}}  d_\phi (\mathbf{\tilde{x}}|\mathbf{c})  \Vert_2 - 1 \bigg)^2 \bigg ],
\end{align}
\end{subequations}
where $\mathbf{\tilde{x}}$ is implicitly defined by sampling convex combinations between the data and the generator distributions $\mathbf{\tilde{x}} = \rho \mathbf{\hat{x}} + (1-\rho) \mathbf{x}$ with $\rho \sim \mathbb{U}(0,1)$. The WGAN-GP constrains the gradient norm of the discriminator's output with respect to its input to enforce the 1-Lipschitz conditions. This strategy differs from the weight clipping of WGAN that sometimes generates only poor samples or fails to converge.
\ref{annex:gan} details the successive improvements from the original GAN to the WGAN and the final WGAN-GP implemented, referred to as GAN in the rest of the paper.
Figure~\ref{fig:GAN_structure} depicts the process of a conditional generative adversarial network for PV generation.

\subsection{Theoretical comparison}\label{sec:comparison}

Normalizing flows are a generative model that allows exact likelihood calculation. They are efficiently parallelizable and offer a valuable latent space for downstream tasks. In contrast to GANs and VAEs, NFs explicitly learn the data distribution and provide direct access to the exact likelihood of the model’s parameters, hence offering a sound and direct way to optimize the network parameters \citep{wehenkel2020graphical}. However, NFs suffer from drawbacks \citep{bond2021deep}. One disadvantage of requiring transformations to be invertible is that the input dimension must be equal to the output dimension, making the model difficult to train or inefficient. Each transformation must be sufficiently expressive while being easily invertible to compute the Jacobian determinant efficiently. The first issue is also raised by \citet{ruthotto2021introduction} where ensuring sufficient similarity of the distribution of interest and the latent distribution is of high importance to obtain meaningful and relevant samples. However, in our numerical simulations, we did not encounter this problem. Concerning the second issue, the UMNN-MAF transformation provides an expressive and effective way of computing the Jacobian.

VAEs indirectly optimize the log-likelihood of the data by maximizing the variational lower bound. The advantage of VAEs over NFs is their ability to handle non-invertible generators and the arbitrary dimension of the latent space. However, it has been observed that when applied to complex datasets such as natural images, VAEs samples tend to be unrealistic. There is evidence that the limited approximation of the true posterior, with a common choice being a normally distributed prior with diagonal covariance, is the root cause \citep{zhao2017towards}. This statement comes from the field of computer vision. However, it may explain the shape of the scenarios observed in our numerical experiments in Section \ref{sec:numerical_results}.

The training of GANs relies on a min-max problem where the generator and the discriminator are jointly optimized. Therefore, it does not rely on estimates of the likelihood or latent variable. The adversarial nature of GANs makes them notoriously difficult to train due to the saddle point problem \citep{arjovsky2017towards}. Another drawback is the mode collapsing, where one network stays in bad local minima, and only a small subset of the data distribution is learned. Several improvements have been designed to address these issues, such as the Wasserstein GAN with gradient penalty. Thus, GANs models are widely used in computer vision and power systems. However, most GAN approaches require cumbersome hyperparameter tuning to achieve similar results to VAEs or NFs. In our numerical simulations, the GAN is highly sensitive to hyperparameter variations, which is consistent with \citep{ruthotto2021introduction}.

Each method has its advantages and drawbacks and makes trade-offs in terms of computing time, hyper-parameter tuning, architecture complexity, \textit{etc}. Therefore, the choice of a particular method is dependent on the user criteria and the dataset considered. In addition, the challenges of power systems are different from computer vision. Therefore, the limitations established in the computer vision literature such as \citet{bond2021deep} and \citet{ruthotto2021introduction} must be addressed with caution. Therefore, we encourage the energy forecasting practitioners to test and compare these methods in power systems applications.

\section{Value and quality assessment}\label{sec:quality_assessment}

For predictions in any form, one must differentiate between their quality and their value \citep{morales2013integrating}. Forecast quality corresponds to the ability of the forecasts to genuinely inform of future events by mimicking the characteristics of the processes involved. Forecast value relates, instead, to the benefits from using forecasts in a decision-making process, such as participation in the electricity market. 
%

\subsection{Forecast quality}\label{sec:quality_methodology}

Evaluating and comparing generative models remains a challenging task. Several measures have been introduced with the emergence of new models, particularly in the field of computer vision. However, there is no consensus or guidelines as to which metric best captures the strengths and limitations of models. Generative models need to be evaluated directly to the application they are intended for \citep{theis2015note}. Indeed, good performance to one criterion does not imply good performance to the other criteria. Several studies propose metrics and make attempts to determine the pros and cons. We selected two that provide helpful information. (1) 24 quantitative and five qualitative measures for evaluating generative models are reviewed and compared by \citet{borji2019pros} with a particular emphasis on GAN-derived models. (2) several representative sample-based evaluation metrics for GANs are investigated by \citet{xu2018empirical} where the kernel Maximum Mean Discrepancy (MMD) and the 1-Nearest-Neighbour (1-NN) two-sample test seem to satisfy most of the desirable properties. The key message is to combine several complementary metrics to assess the generative models. Some of the metrics proposed are related to image generation and cannot directly be transposed to energy forecasting.

Therefore, we used eight complementary quality metrics to conduct a relevant quality analysis inspired by the energy forecasting and computer vision fields.
They can be divided into four groups: (1) the \textit{univariate} metrics composed of the continuous ranked probability score, the quantile score, and the reliability diagram. They can only assess the quality of the scenarios to their marginals; (2) the \textit{multivariate} metrics are composed of the energy and the variogram scores. They can directly assess multivariate scenarios; (3) the \textit{specific} metrics composed of a classifier-based metric and the correlation matrix between scenarios for a given context; (4) the Diebold and Mariano statistical test.
The basics of these metrics are provided in the following, and \ref{annex:quality-assessment} presents the mathematical definitions and the details of implementation.
%

\subsubsection*{Univariate metrics}

The continuous ranked probability score (CRPS) \citep{gneiting2007strictly} is a univariate scoring rule that penalizes the lack of resolution of the predictive distributions as well as biased forecasts. It is negatively oriented, \textit{i.e.}, the lower, the better, and for deterministic forecasts, it turns out to be the mean absolute error (MAE). The CRPS is used to compare the skill of predictive marginals for each component of the random variable of interest. In our case, for the twenty-four time periods of the day. 

The quantile score (QS), also known as the pinball loss score, complements the CRPS. It permits obtaining detailed information about the forecast quality at specific probability levels, \textit{i.e.}, over-forecasting or under-forecasting, and particularly those related to the tails of the predictive distribution \citep{lauret2019verification}. It is negatively oriented and assigns asymmetric weights to negative and positive errors for each quantile. 

Finally, the reliability diagram is a visual verification used to evaluate the reliability of the quantiles derived from the scenarios. Quantile forecasts are reliable if their nominal proportions are equal to the proportions of the observed value.

\subsubsection*{Multivariate metrics}

The energy score (ES) is the most commonly used scoring rule when a finite number of trajectories represents distributions. It is a multivariate generalization of the CRPS and has been formulated and introduced by \citet{gneiting2007strictly}. The ES is proper and negatively oriented, \textit{i.e.}, a lower score represents a better forecast. 
The ES is used as a multivariate scoring rule by \citet{golestaneh2016generation} to investigate and analyze the spatio-temporal dependency of PV generations. They emphasize the ES pros and cons. It is capable of evaluating forecasts relying on marginals with correct variances but biased means. Unfortunately, its ability to detect incorrectly specified correlations between the components of the multivariate quantity is somewhat limited.
The ES is selected as a multivariate scoring rule in this study to quantitatively assess the performance of the generative methods similar to the mean absolute error when considering point forecasts.

An alternative class of proper scoring rules based on the geostatistical concept of variograms is proposed by \citet{scheuerer2015variogram}. They study the sensitivity of these variogram-based scoring rules to inaccurate predicted means, variances, and correlations. The results indicate that these scores are distinctly more discriminative to the correlation structure. Thus, in contrast to the Energy score, the Variogram score captures correlations between multivariate components.

\subsubsection*{Specific metrics}

A conditional classifier-based scoring rule is designed by implementing an Extra-Trees classifier \citep{geurts2006extremely} to discriminate true from generated samples. The receiver operating characteristic (ROC) curves are computed for each generative model on the testing set. The best generative model should achieve an area under the ROC curve (AUC) of 0.5, \textit{i.e.}, each sample is equally likely to be predicted as true or false, meaning the classifier is unable to discriminate generated scenarios from the actual observations.
Note: ROC curve is the relationship between True Positive Rate and False Positive Rate given by different thresholds. AUC ROC is the area under the ROC curve, and it is the metric used to measure how well the model can distinguish two classes. We recommend the article of \citet{fawcett2004roc} that is designed both as a tutorial introduction to ROC graphs and as a practical guide for using them in research.

The second specific metric consists of computing the correlation matrix between the scenarios generated for given weather forecasts. Formally, let $ \{    \mathbf{\hat{x}}^i  \}_{i=1}^M$ be the set of $M$ scenarios generated for a given day of the testing set. It is a matrix ($M \times 24$) where each row is a scenario. Then, the Pearson’s correlation coefficients are computed into a correlation matrix ($24 \times 24$). This metric indicates the variety of scenario shapes.

\subsubsection*{Statistical testing}
Using relevant metrics to assess the forecast quality is essential. However, it is also necessary to analyze whether any difference in accuracy is statistically significant. Indeed, when different models have almost identical values in the selected error measures, it is not easy to draw statistically significant conclusions on the outperformance of the forecasts of one model by those of another.
The Diebold-Mariano (DM) test \citep{diebold2002comparing} is probably the most commonly used statistical testing tool to evaluate the significance of differences in forecasting accuracy. It is model-free, \textit{i.e.}, it compares the forecasts of models, and not models themselves.
The DM test is used in this study to assess the CRPS, QS, ES, and VS metrics. The CRPS and QS are univariate scores, and a value of CRPS and QS is computed per marginal (time period of the day). Therefore, the multivariate variant of the DM test is implemented following \citet{ziel2018day}, where only one statistic for each pair of models is computed based on the 24-dimensional vector of errors for each day.

\subsection{Forecast value}\label{sec:forecast_value}

A model that yields lower errors in terms of forecast quality may not always point to a more effective model for forecast practitioners \citep{hong2020energy}. To this end, similarly to \citet{toubeau2018deep}, the forecast value is assessed by considering the day-ahead market scheduling of electricity aggregators, such as energy retailers or generation companies. The energy retailer aims to balance its portfolio on an hourly basis to avoid financial penalties in case of imbalance by exchanging the surplus or deficit of energy in the day-ahead electricity market. The energy retailer may have a battery energy storage system (BESS) to manage its portfolio and minimize imports from the main grid when day-ahead prices are prohibitive.

Let $e_t$ [MWh] be the net energy retailer position on the day-ahead market during the $t$-th hour of the day, modeled as a first stage variable. Let $y_t$ [MWh] be the realized net energy retailer position during the $t$-th hour of the day, which is modeled as a second stage variable due to the stochastic processes of the PV generation, wind generation, and load. Let $\pi_t$ [\euro / MWh] the clearing price in the spot day-ahead market for the $t$-th hour of the day, $q_t$ ex-post settlement price for negative imbalance $y_t < e_t$, and $\lambda_t$ ex-post settlement price for positive imbalance $y_t > e_t$. 
The energy retailer is assumed to be a price taker in the day-ahead market. It is motivated by the individual energy retailer capacity being negligible relative to the whole market. The forward settlement price $\pi_t$ is assumed to be fixed and known. As imbalance prices tend to exhibit volatility and are difficult to forecast, they are modeled as random variables, with expectations denoted by $\bar{q}_t = \mathop{\mathbb{E}}  [ q_t  ]$ and $\bar{\lambda}_t = \mathop{\mathbb{E}}  [ \lambda_t  ]$. They are assumed to be independent random variables from the energy retailer portfolio.

A stochastic planner with a linear programming formulation and linear constraints is implemented using a scenario-based approach. The planner computes the day-ahead bids $e_t$ that cannot be modified in the future when the uncertainty is resolved. The second stage corresponds to the dispatch decisions $y_{t,\omega}$ in scenario $\omega$ that aims at avoiding portfolio imbalances modeled by a cost function $f^c$. The second-stage decisions are therefore scenario-dependent and can be adjusted according to the realization of the stochastic parameters. 
The stochastic planner objective to maximize is
\begin{align}\label{eq:S_objective_1}	
J_S &  = \mathop{\mathbb{E}} \bigg[ \sum_{t\in \mathcal{T}}  \pi_t e_t +  f^c(e_t, y_{t,\omega}) \bigg] ,
\end{align}
where the expectation is taken to the random variables, the PV generation, wind generation, and load. Using a scenario-based approach, (\ref{eq:S_objective_1}) is approximated by
\begin{align}\label{eq:S_objective_2}	
J_S &  \approx  \sum_{\omega \in \Omega} \alpha_\omega  \sum_{t\in \mathcal{T}} \bigg[ \pi_t e_t  +  f^c(e_t, y_{t,\omega})  \bigg],
\end{align}
with $\alpha_\omega$ the probability of scenario $\omega \in \Omega$, and $\sum_{\omega \in \Omega} \alpha_\omega = 1$. The optimization problem is detailed in~\ref{annex:value-assessment}. 

\section{Numerical Results}\label{sec:numerical_results}

\begin{figure}[htbp]
	\centering
	\includegraphics[width=\linewidth]{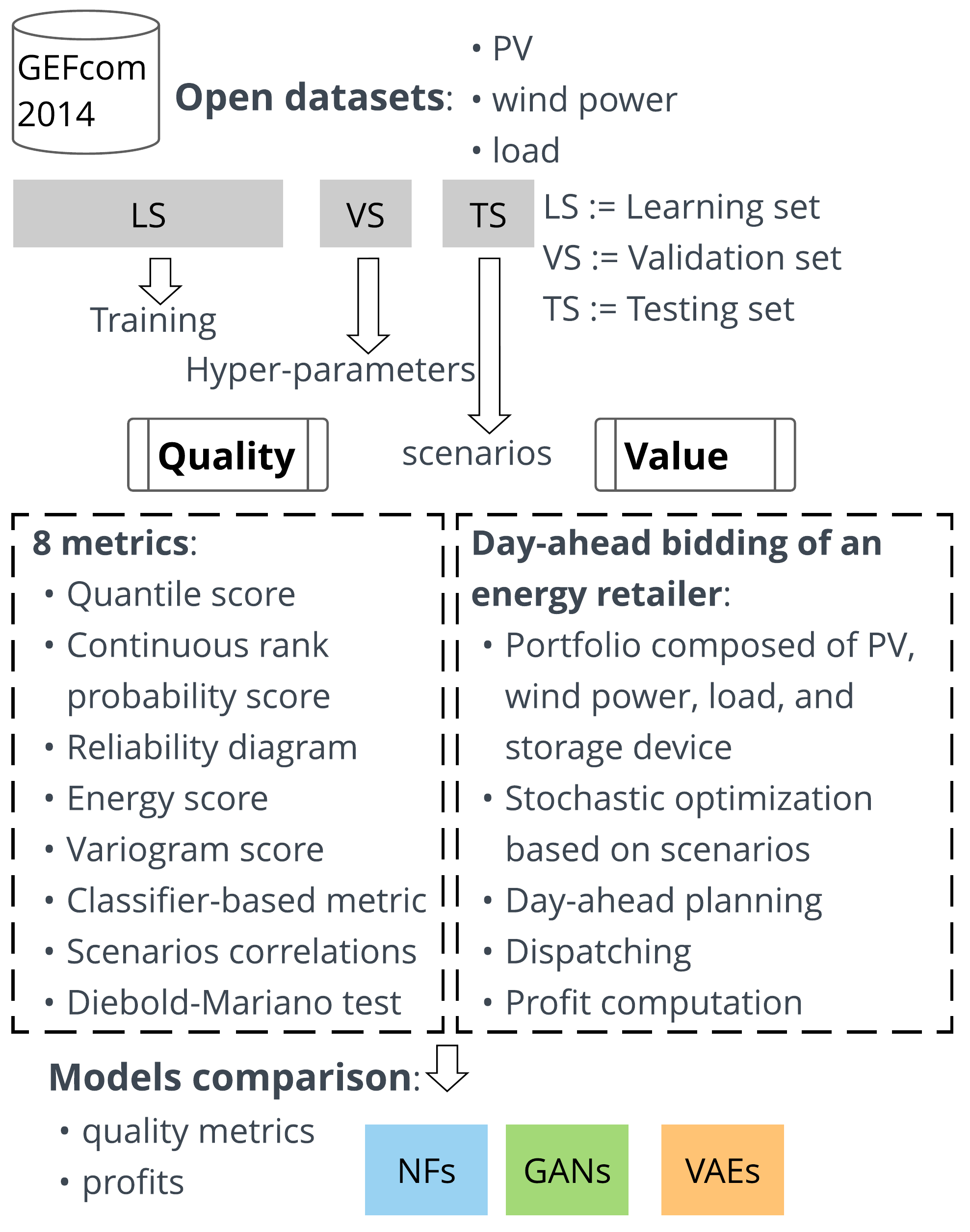}
	\caption{Methodology to assess both the quality and value of the GAN, VAE, and NF models implemented in this study. \\
    The PV, wind power, and load datasets of the open-access Global Energy Forecasting Competition 2014 are divided into three parts: learning, validation, and testing sets. The learning set is used to train the models, the validation set to select the optimal hyper-parameters, and the testing set to conduct the numerical experiments. The quality and value of the models are assessed by using the scenarios generated on the testing set. The quality evaluation consists of eight complementary metrics. The value assessment uses the simple and easily reproducible case study of the day-ahead bidding of an energy retailer. The energy retailer portfolio is composed of PV, wind power generation, load, and a storage system device. The retailer bids on the day-ahead market by computing a planning based on stochastic optimization. The dispatch is computed by using the observations of the PV generation, wind power, and load. Then, the profits are evaluated and compared.
	}
	\label{fig:numerical-experiments-methodology}
\end{figure}

The quality and value evaluations of the models are conducted on the load, wind, and PV tracks of the open-access GEFCom 2014 dataset \citep{hong2016bprobabilistic}, composed of one, ten, and three zones, respectively.
Figure \ref{fig:numerical-experiments-methodology} depicts the methodology to assess both the quality and value of the GAN, VAE and NF models implemented in this study.
%

\subsection{Implementation details}\label{sec:data_description}

By appropriate normalization, we standardize the weather forecasts to have a zero mean and unit variance. Table~\ref{tab:features} provides a summary of the implementation details described in what follows. For the sake of proper model training and evaluation, the dataset is divided into three parts per track considered: learning, validation, and testing sets. The learning set (LS) is used to fit the models, the validation set (VS) to select the optimal hyper-parameters, and the testing set (TS) to assess the forecast quality and value. The number of samples~($\#$), expressed in days, of the VS and TS, is $50 \cdot n_z$, with $n_z$ the number of zones of the track considered. The 50 days are selected randomly from the dataset, and the learning set is composed of the remaining part with $D \cdot n_z$ samples, where $D$ is provided for each track in Table~\ref{tab:features}.
The NF, VAE, and GAN use the weather forecasts as inputs to generate on a day-ahead basis $M$ scenarios $\mathbf{\hat{x}} \in \mathbb{R}^T$. The hyper-parameters values used for the experiments are provided in~\ref{annex:hp}.


\subsubsection*{Wind track}

The zonal $\mathbf{u}^\text{10}$, $\mathbf{u}^\text{100}$ and meridional $\mathbf{v}^\text{10}$, $\mathbf{v}^\text{100}$ wind components at 10 and 100 meters are selected, and six features are derived following the formulas provided by \citet{landry2016probabilistic} to compute the wind speed $\mathbf{ws}^\text{10}$, $\mathbf{ws}^\text{100}$, energy $\mathbf{we}^\text{10}$, $\mathbf{we}^\text{100}$ and direction $\mathbf{wd}^\text{10}$, $\mathbf{wd}^\text{100}$ at 10 and 100 meters
\begin{subequations}
\label{eq:wind_derived_features}	
\begin{align}
\mathbf{ws} = & \sqrt{\mathbf{u} + \mathbf{v}} ,\\
\mathbf{we} = & \frac{1}{2} \mathbf{ws}^3,\\
\mathbf{wd} = & \frac{180}{\pi} \arctan(\mathbf{u}, \mathbf{v}).
\end{align}
\end{subequations}
For each generative model, the wind zone is taken into account with one hot-encoding variable $Z_1, \ldots, Z_{10}$, and the wind feature input vector for a given day $d$ is
\begin{align}
\label{eq:wind_input}	
\mathbf{c}_d^\text{wind} = & [\mathbf{u}^\text{10}_d, \mathbf{u}^\text{100}_d, \mathbf{v}^\text{10}_d, \mathbf{v}^\text{100}_d, \mathbf{ws}^\text{10}_d, \mathbf{ws}^\text{100}_d, \notag \\
& \mathbf{we}^\text{10}_d, \mathbf{we}^\text{100}_d, \mathbf{wd}^\text{10}_d, \mathbf{wd}^\text{100}_d, Z_1, \ldots, Z_{10}],
\end{align}
of dimension $n_f \cdot T + n_z = 10 \cdot 24 + 10$. 

\subsubsection*{PV track}

The solar irradiation $\mathbf{I}$, the air temperature $\mathbf{T}$, and the relative humidity $\mathbf{rh}$ are selected, and two features are derived by computing $\mathbf{I}^2$ and $\mathbf{IT}$. For each generative model, the PV zone is taken into account with one hot-encoding variable $Z_1, Z_2, Z_3$, and the PV feature input vector for a given day $d$ is
\begin{align}
\label{eq:PV_input}	
\mathbf{c}_d^\text{PV} = & [\mathbf{I}_d, \mathbf{T}_d, \mathbf{rh}_d, \mathbf{I}^2_d, \mathbf{IT}_d, Z_1, Z_2, Z_3],
\end{align}
of dimension $n_f \cdot T + n_z$. For practical reasons, the periods where the PV generation is always 0, across all zones and days, are removed, and the final dimension of the input feature vector is $n_f \cdot T + n_z = 5 \cdot 16 + 3$.

\subsubsection*{Load track}

The 25 weather station temperature $\mathbf{w}_1, \ldots, \mathbf{w}_{25}$ forecasts are used. There is only one zone, and the load feature input vector for a given day $d$ is
\begin{align}
\label{eq:load_input}	
\mathbf{c}_d^\text{load} = & [\mathbf{w}_1, \ldots, \mathbf{w}_{25}],
\end{align}
of dimension $n_f \cdot T = 25 \cdot 24$.
\begin{table}[htbp]
\renewcommand{\arraystretch}{1.25}
	\begin{center}
		\begin{tabular}{lrrr}
			\hline \hline
			            			     & Wind &  PV   & Load  \\ \hline
		$T$ periods 	                 & 24 &  16     & 24  \\
		$n_z$ zones     			     & 10 &  3      & ---  \\
		$n_f$ features  			     & 10 &  5      & 25  \\
		$\mathbf{c}_d$ dimension           & $n_f \cdot T + n_z$ &  $n_f \cdot T + n_z$   & $n_f \cdot T$  \\
		$\#$ LS (days)   		         & $631\cdot n_z$ &  $720 \cdot n_z$        	  & $1999$  \\
		$\#$ VS/TS (days)   	         & $50\cdot n_z$ &  $50 \cdot n_z$         	  & $50$\\
         \hline \hline
		\end{tabular}
		\caption{Dataset and implementation details. \\
		Each dataset is divided into three parts: learning, validation, and testing sets. The number of samples~($\#$) is expressed in days and is set to 50 days for the validation and testing sets. $T$ is the number of periods per day considered, $n_z$ the number of zones of the dataset, $n_f$ the number of weather variables used, and $\mathbf{c}_d$ is the dimension of the conditional vector for a given day that includes the weather forecasts and the one hot-encoding variables when there are several zones. Note: the days of the learning, validation, and testing sets are selected randomly.
		}
		\label{tab:features}
	\end{center}
\end{table}

The number of samples~($\#$), expressed in days, of the VS and TS, is $50 \cdot n_z$, with $n_z$ the number of zones of the track considered. The 50 days are selected randomly from the dataset, and the learning set is composed of the remaining part with $D \cdot n_z$ samples, where $D$ is provided for each track.

\subsection{Quality results}\label{sec:quality_res}

A thorough comparison of the models is conducted on the wind track, and~\ref{annex:assessment_results} provides the Figures of the other tracks for the sake of clarity. Note: the model ranking slightly differs depending on the track.

\subsubsection*{Wind track}

\begin{figure}[htbp]
	\centering
	\begin{subfigure}{.4\textwidth}
		\centering
		\includegraphics[width=\linewidth]{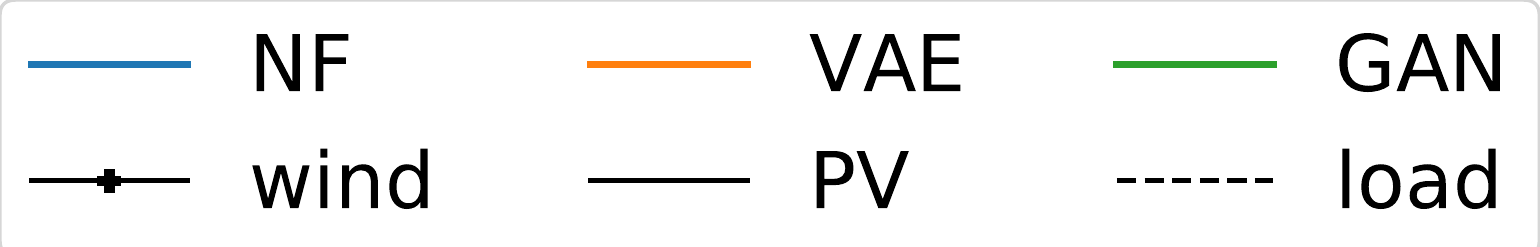}
	\end{subfigure}
	\begin{subfigure}{.45\textwidth}
		\centering
		\includegraphics[width=\linewidth]{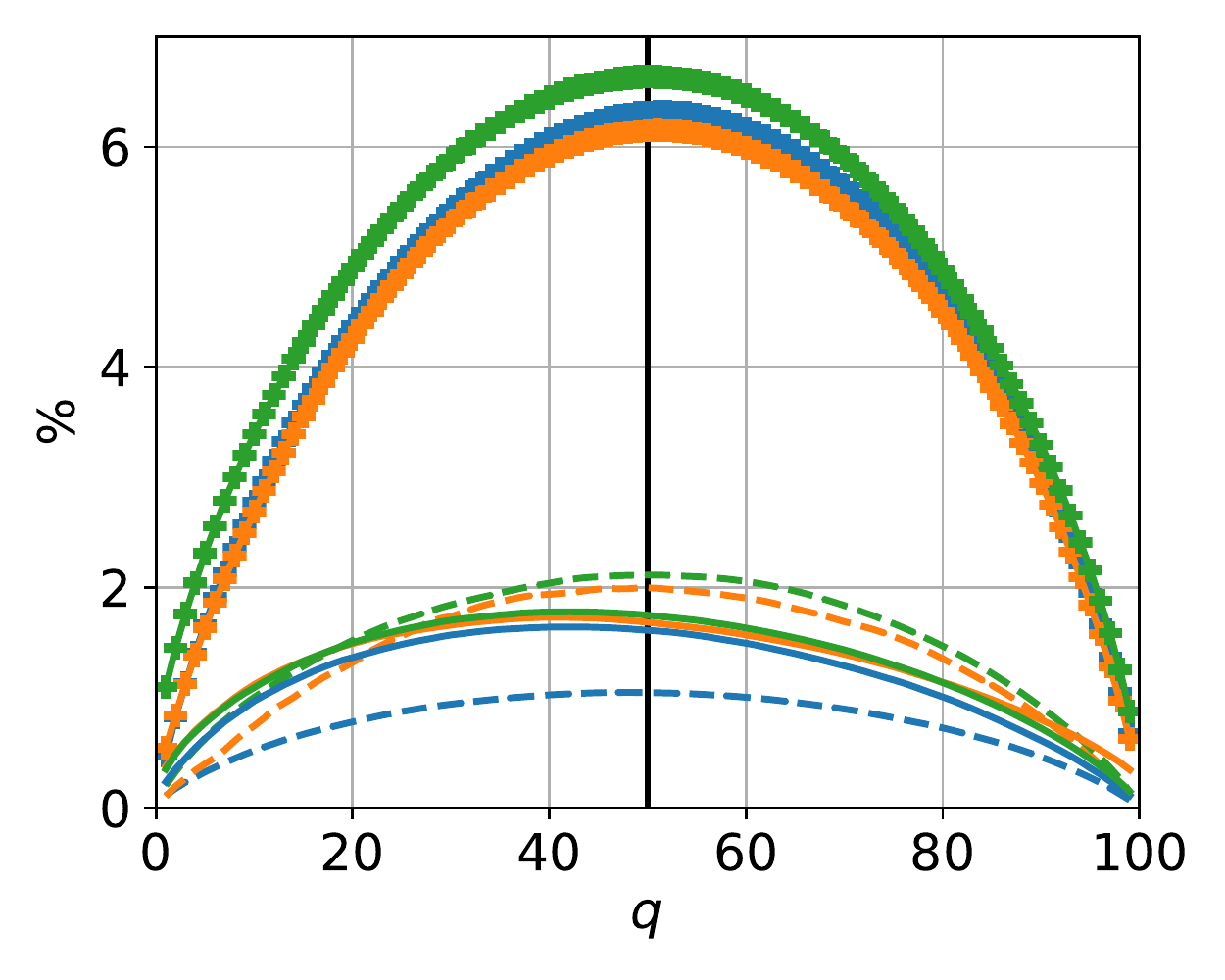}
		\caption{Quantile score.}
	\end{subfigure}
	\begin{subfigure}{.45\textwidth}
		\centering
		\includegraphics[width=\linewidth]{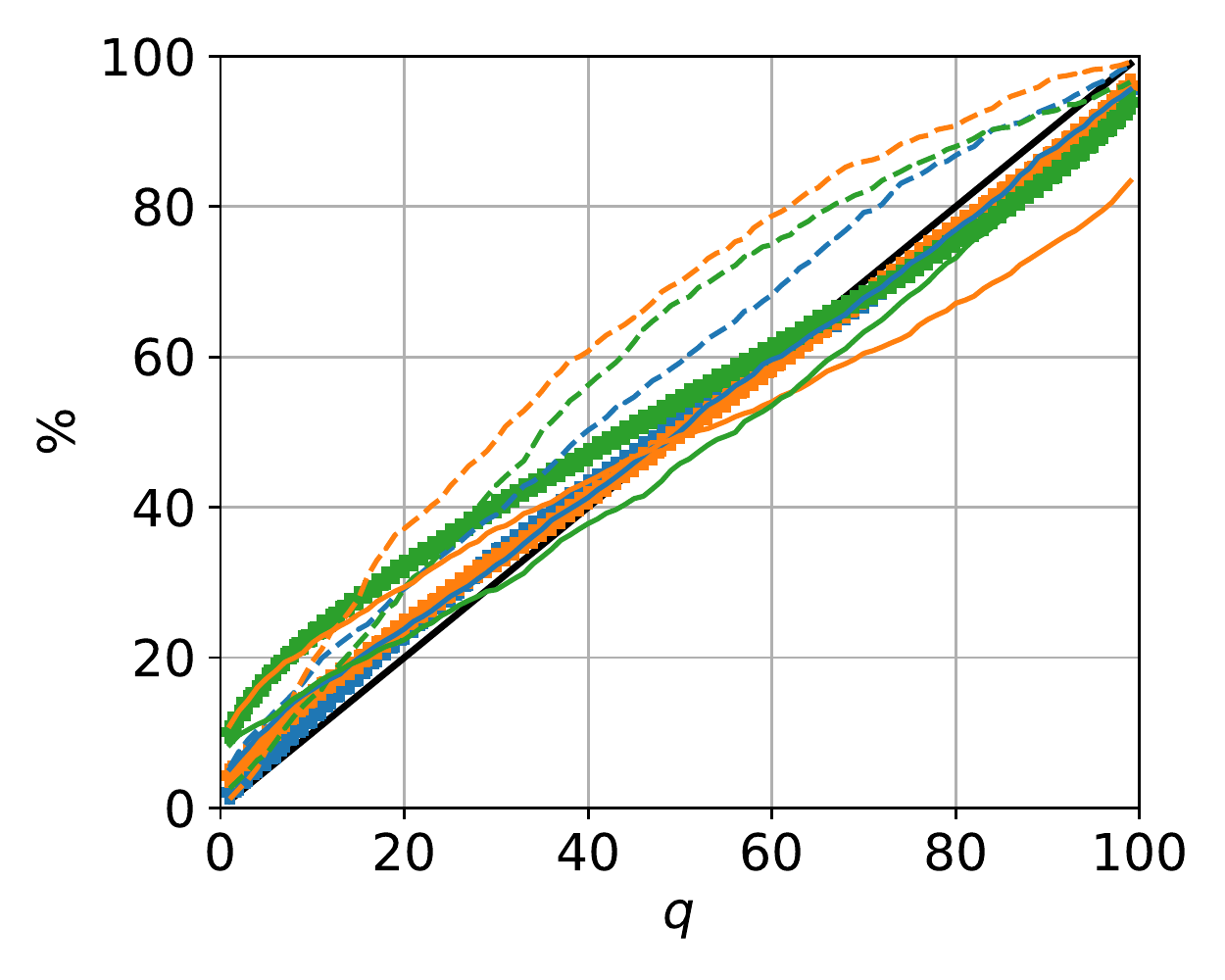}
		\caption{Reliability diagram.}
	\end{subfigure}
	\begin{subfigure}{.45\textwidth}
		\centering
		\includegraphics[width=\linewidth]{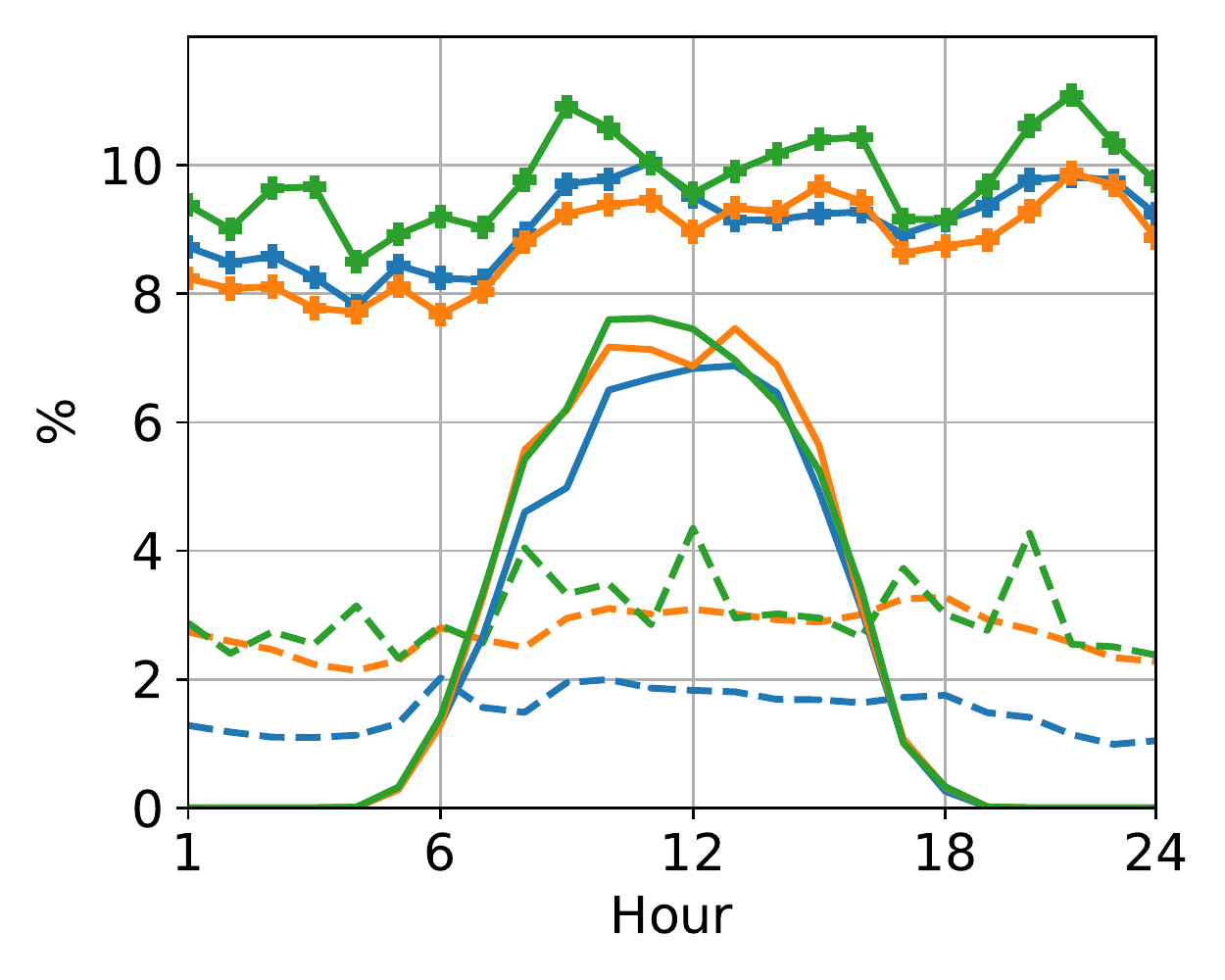}
		\caption{CRPS per marginal.}
	\end{subfigure}
	\caption{Quality standard metrics comparison on the wind (markers), PV (plain), and load (dashed) tracks. \\
	Quantile score (a): the lower and the more symmetrical, the better. Note: the quantile score has been averaged over the marginals (the 24 time periods of the day). Reliability diagram (b): the closer to the diagonal, the better. Continuous ranked probability score per marginal (c): the lower, the better. 
	NF outperforms the VAE and GAN for both the PV and load tracks and is slightly outperformed by the VAE on the wind track. Note: all models tend to have more difficulties forecasting the wind power that seems less predictable than the PV generation or the load.
	}
	\label{fig:quality_comparison}
\end{figure}

In addition to the generative models, a naive approach is designed (RAND), where the scenarios of the learning, validation, and testing sets are sampled randomly from the learning, validation, and testing sets, respectively. Intuitively, it assumes that past observations are repeated, and these scenarios are realistic but may not be compatible with the context.
Each model generates a set of 100 scenarios for each day of the testing set, and the scores are computed following the mathematical definitions provided in \ref{annex:quality-assessment}.
Figure~\ref{fig:quality_comparison} compares the QS, reliability diagram, and CRPS of the wind (markers), PV (plain), and load (dashed) tracks. Overall, for the wind track in terms of CRPS, QS, and reliability diagrams, the VAE achieves slightly better scores, followed by the NF and the GAN. 
The ES and VS multivariate scores confirm this trend with 54.82 and 18.87 for the VAE \textit{vs} 56.71 and 18.54 for the NF, respectively.

\begin{figure}[tb]
	\centering
		\begin{subfigure}{.25\textwidth}
		\centering
		\includegraphics[width=\linewidth]{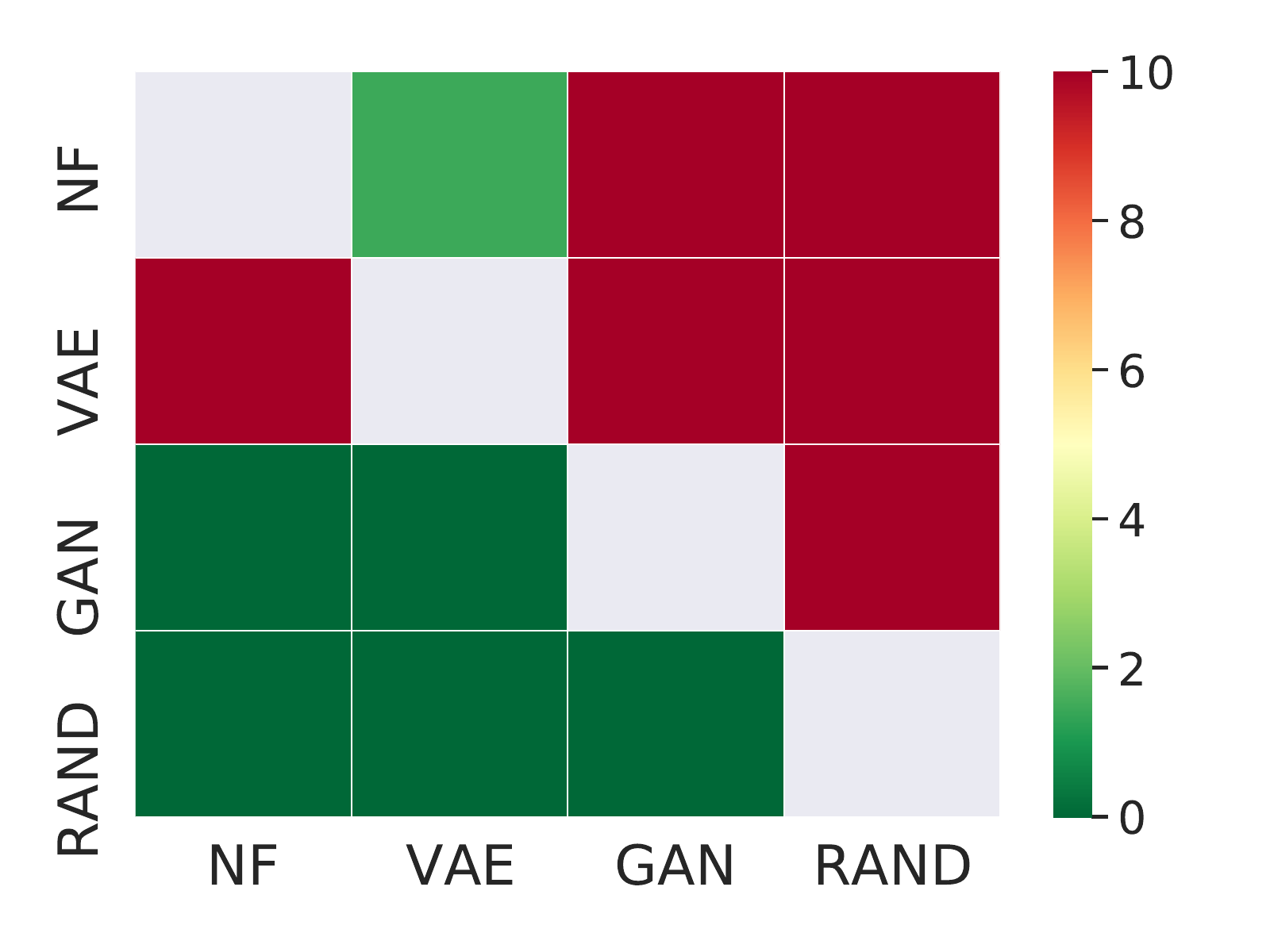}
		\caption{CRPS DM test.}
		\label{fig:DM_wind_CRPS}
	\end{subfigure}%
	\begin{subfigure}{.25\textwidth}
		\centering
		\includegraphics[width=\linewidth]{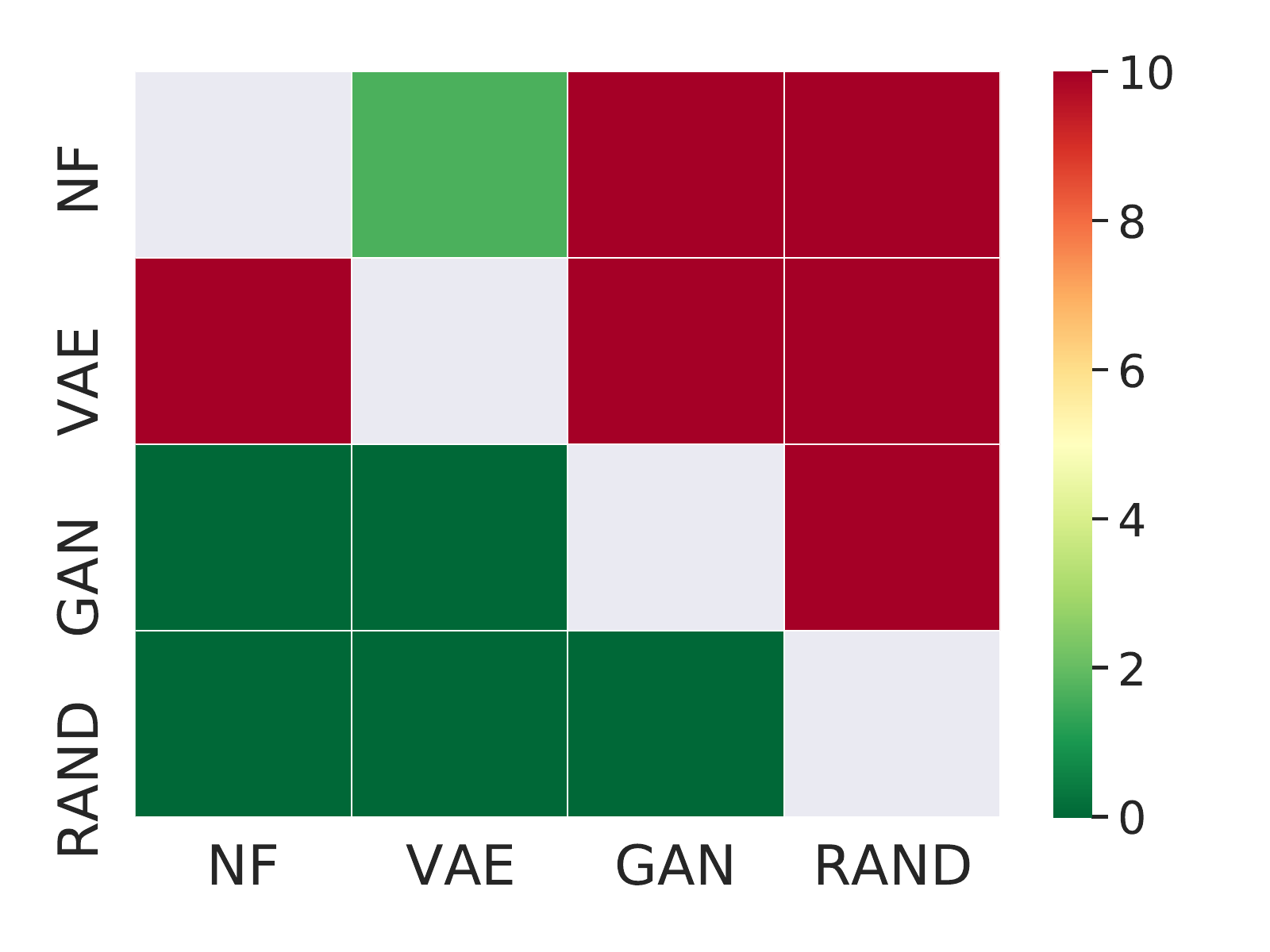}
		\caption{QS DM test.}
		\label{fig:DM_wind_QS}
	\end{subfigure}
	\begin{subfigure}{.25\textwidth}
		\centering
		\includegraphics[width=\linewidth]{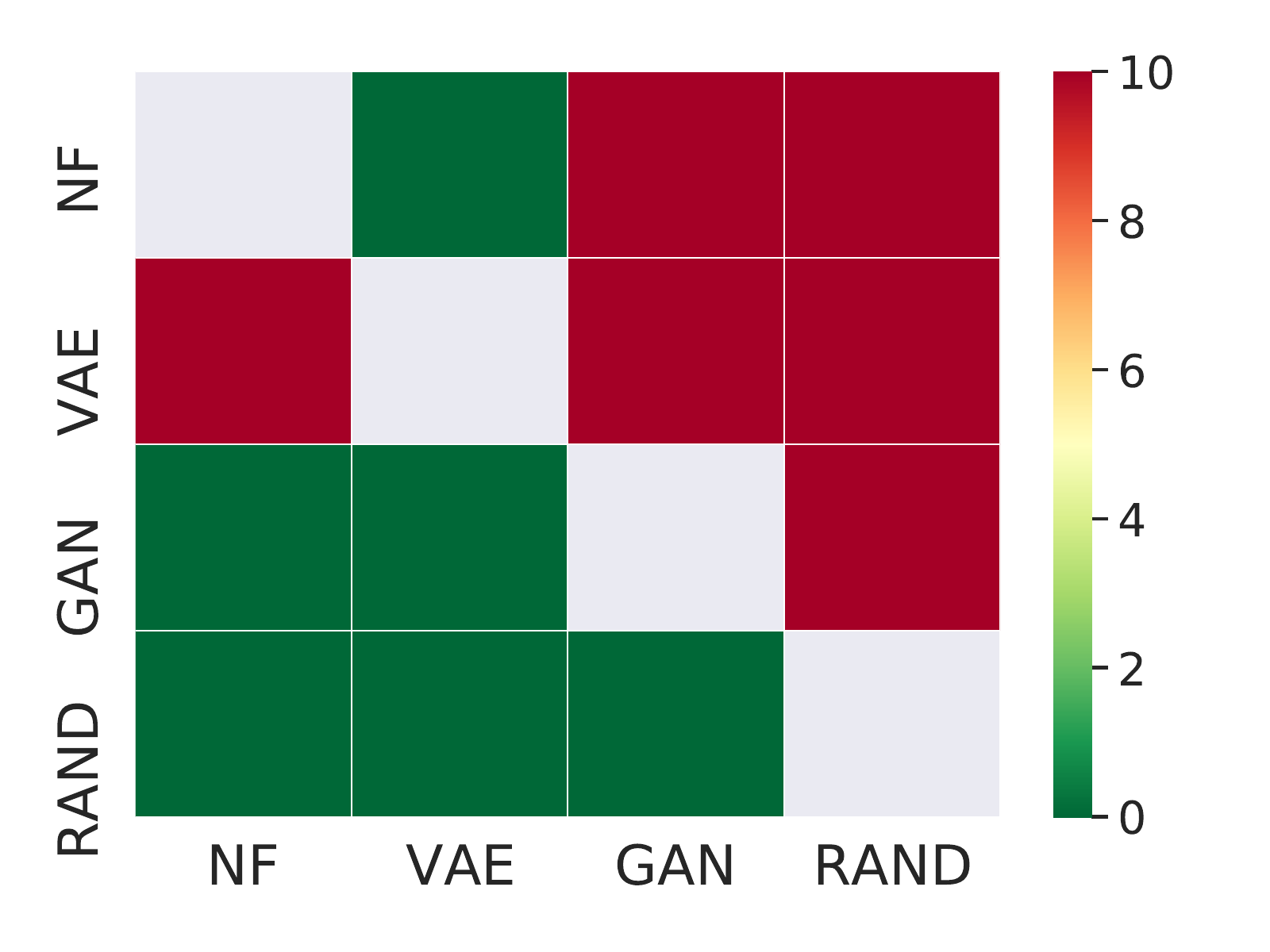}
		\caption{ES DM test.}
	\end{subfigure}%
	\begin{subfigure}{.25\textwidth}
		\centering
		\includegraphics[width=\linewidth]{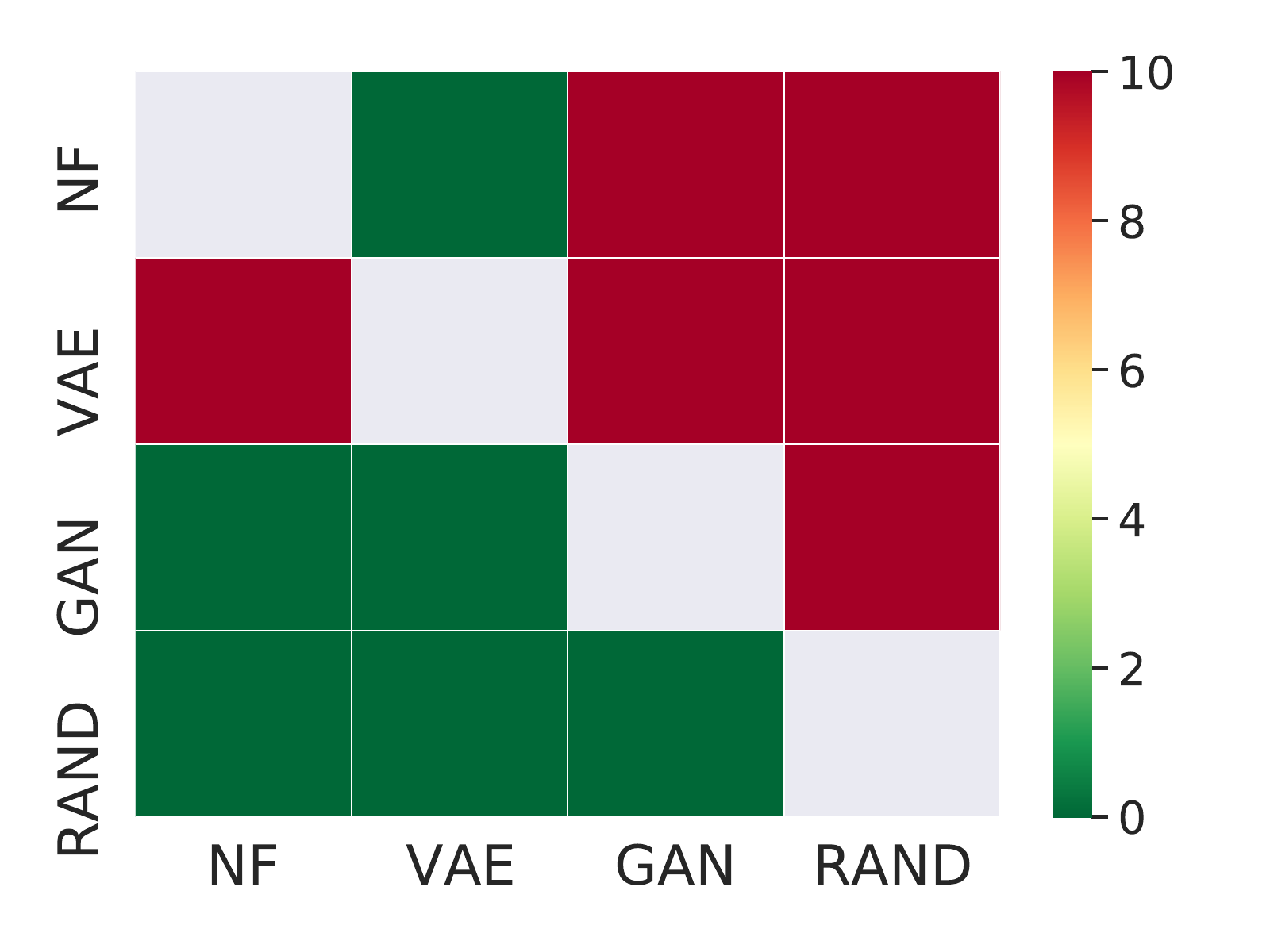}
		\caption{VS DM test.}
	\end{subfigure}
	\caption{Wind track Diebold-Mariano tests of the CRPS, QS, ES, and VS metrics. \\
	The Diebold-Mariano tests of the continuous ranked probability, quantile, energy, and variogram scores confirm that the VAE outperforms the NF on the wind track for these metrics. 
	The heat map indicates the range of the $p$-values. The closer they are to zero, dark green, the more significant the difference between the scores of two models for a given metric. The statistical threshold is five \%, but the scale color is capped at ten \% for a better exposition of the relevant results.
	}
\label{fig:wind-DM}
\end{figure}

\begin{figure}[tb]
\centering
\includegraphics[width=\linewidth]{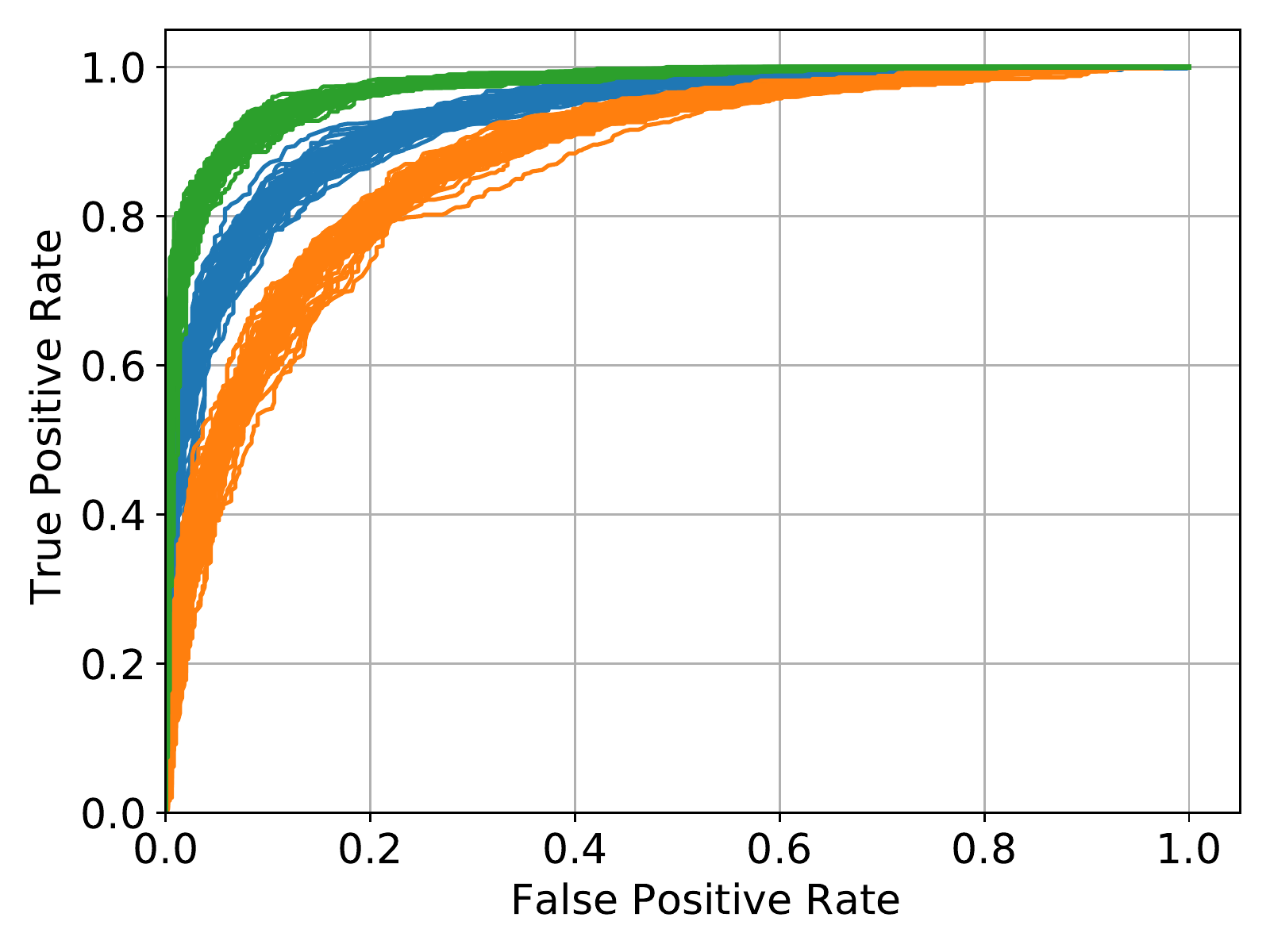}
\caption{Wind track classifier-based metric. \\
The VAE (orange) is the best to mislead the classifier, followed by the NF (blue) and GAN (green). Note: there are 50 ROC curves depicted for each model, each corresponding to a scenario generated used as input of the classifier. It allows taking into account the variability of the scenarios to avoid having results dependent on a particular scenario.
}	
\label{fig:ROC_wind}
\end{figure}

Figure~\ref{fig:wind-DM} provides the results of the DM tests for these metrics. The heat map indicates the range of the $p$-values. The closer they are to zero, dark green, the more significant the difference between the scores of two models for a given metric. The statistical threshold is five \%, but the scale color is capped at ten \% for a better exposition of the relevant results.
For instance, when considering the DM test for the RAND CRPS, all the columns of the RAND row are in dark green, indicating that the RAND scenarios are always significantly outperformed by the other models.
These DM tests confirm that the VAE outperforms the NF for the wind track considering these metrics. Then, the NF is only outperformed by the VAE and the GAN by both the VAE and NF.
%
These results are consistent with the classifier-based metric depicted in Figure~\ref{fig:ROC_wind}, where the VAE is the best to mislead the classifier, followed by the NF and GAN. 

\begin{figure}[tb]
	\centering
	\begin{subfigure}{.45\textwidth}
		\centering
		\includegraphics[width=\linewidth]{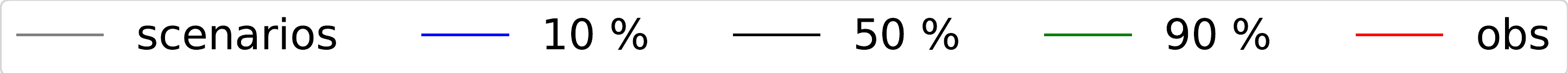}
	\end{subfigure}
	\begin{subfigure}{.25\textwidth}
		\centering
		\includegraphics[width=\linewidth]{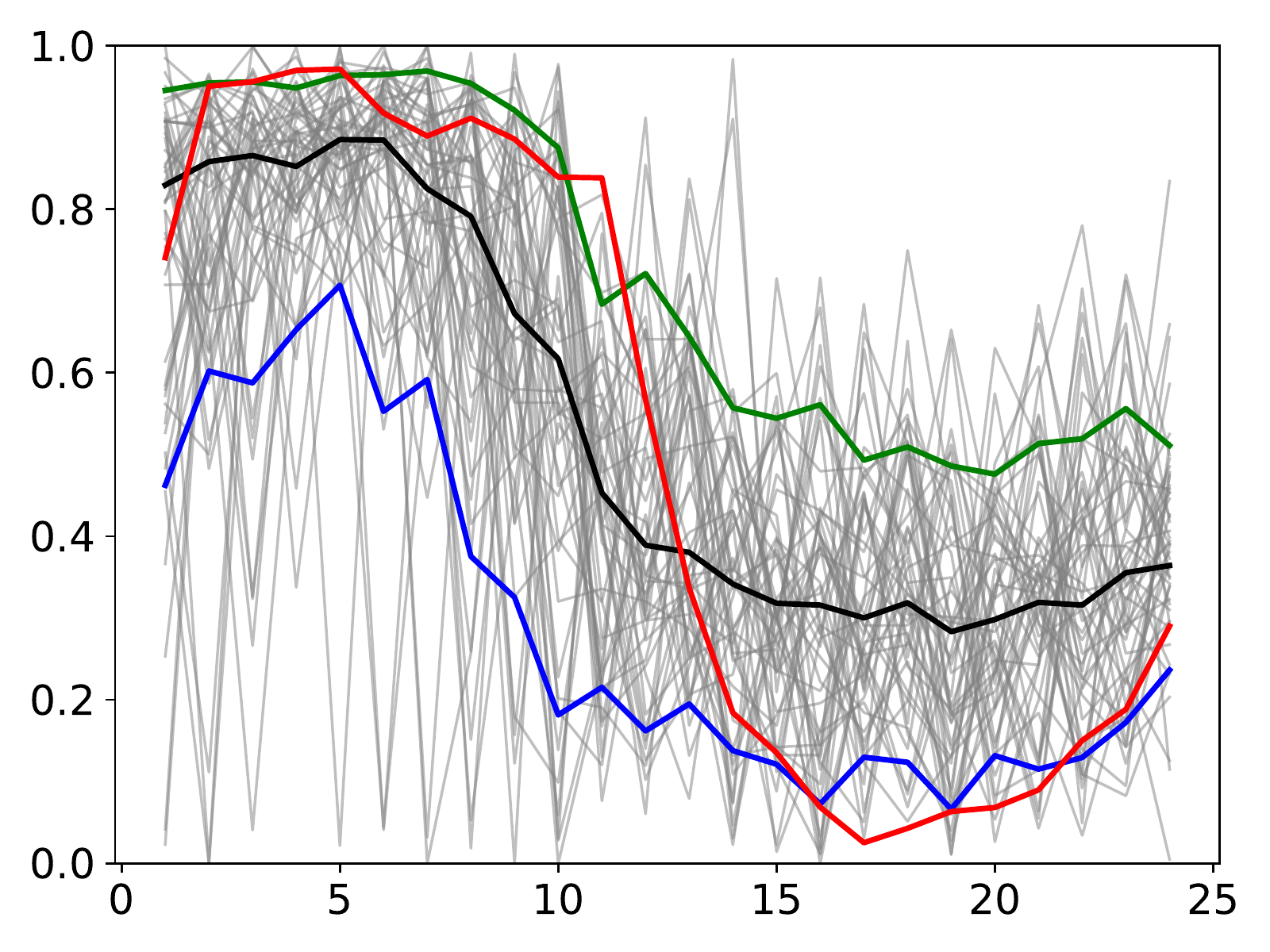}
		\caption{NF.}
	\end{subfigure}%
	\begin{subfigure}{.25\textwidth}
		\centering
		\includegraphics[width=\linewidth]{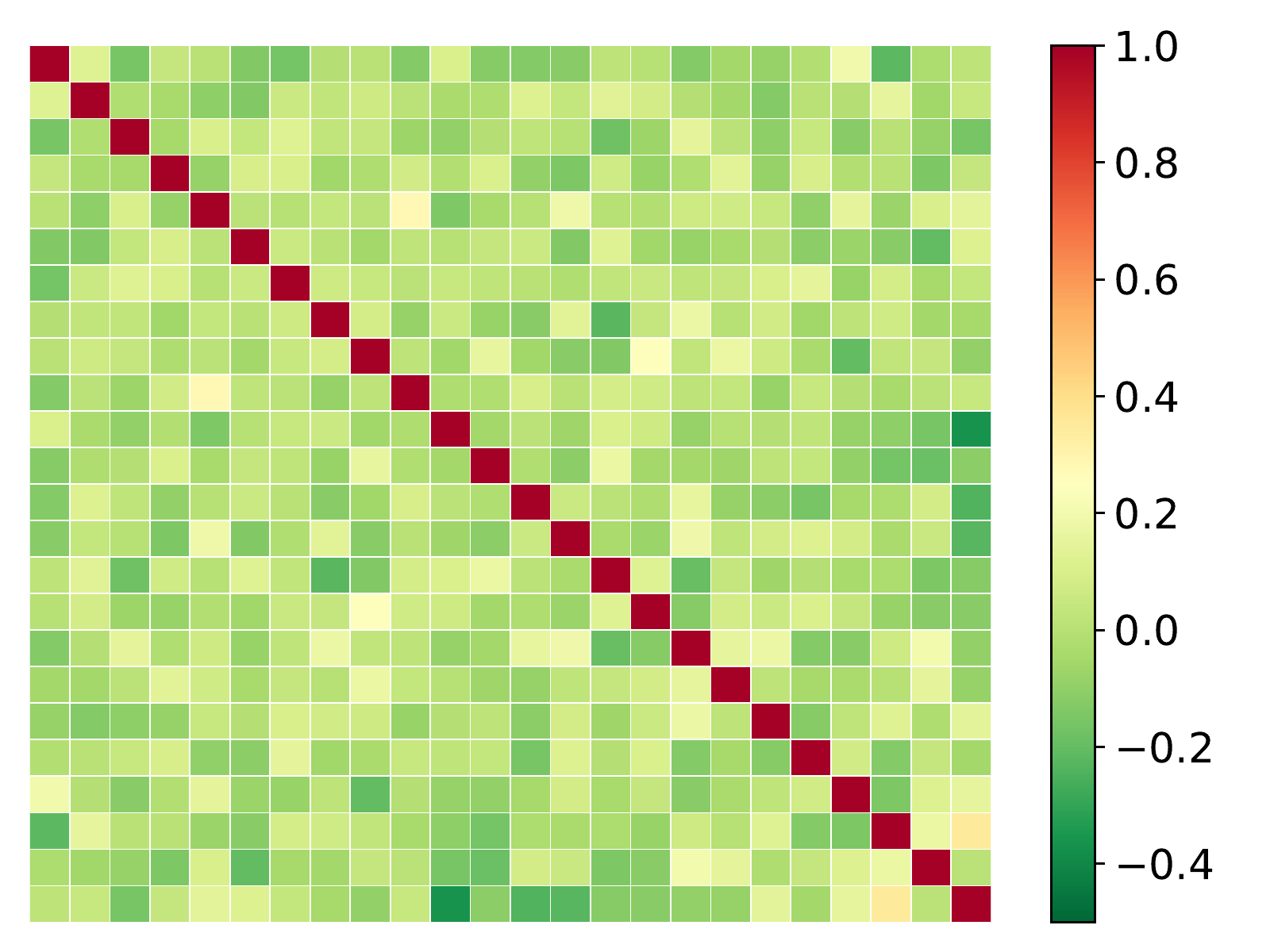}
		\caption{NF.}
	\end{subfigure}
	\begin{subfigure}{.25\textwidth}
		\centering
		\includegraphics[width=\linewidth]{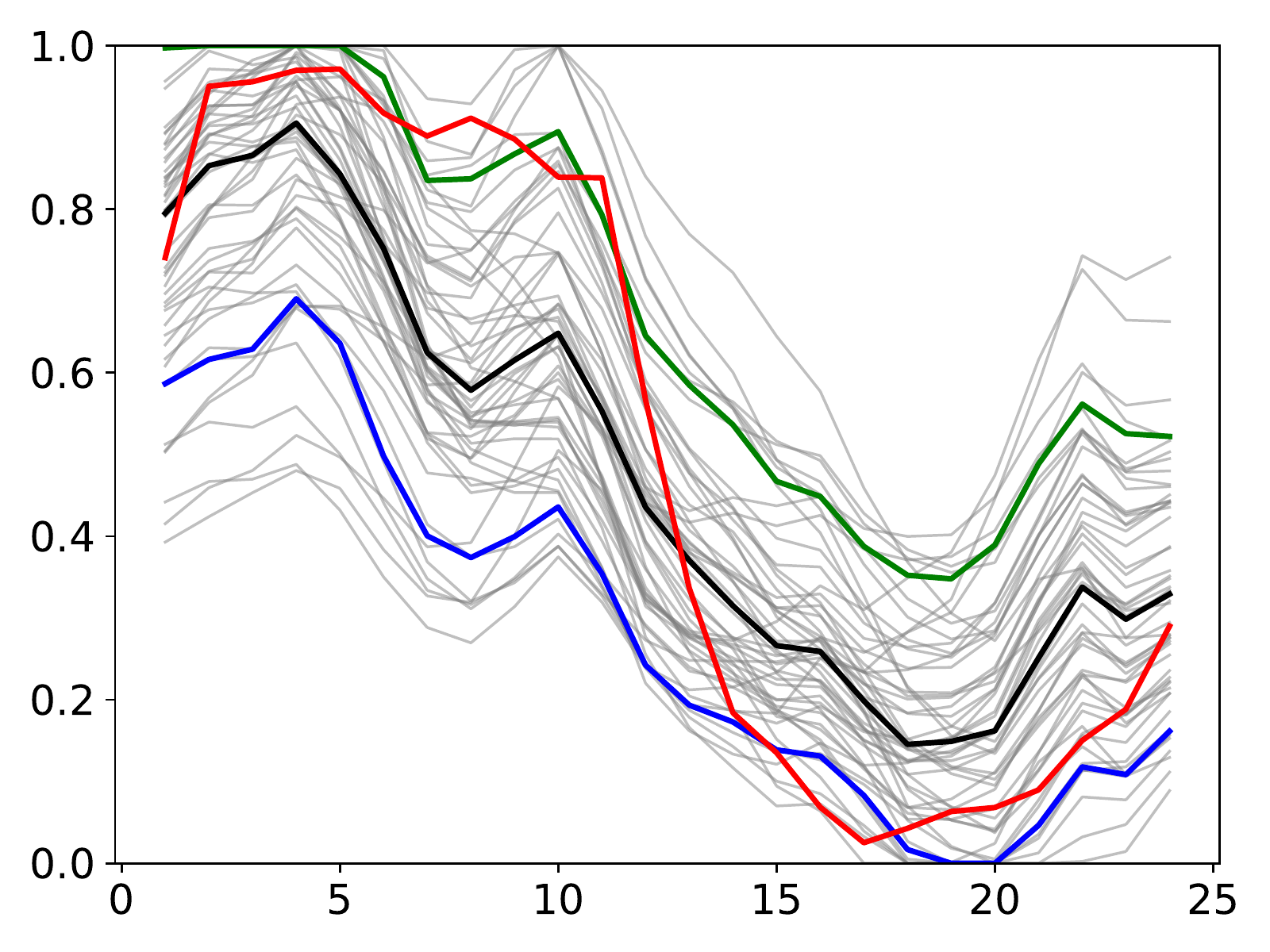}
		\caption{GAN.}
	\end{subfigure}%
	\begin{subfigure}{.25\textwidth}
		\centering
		\includegraphics[width=\linewidth]{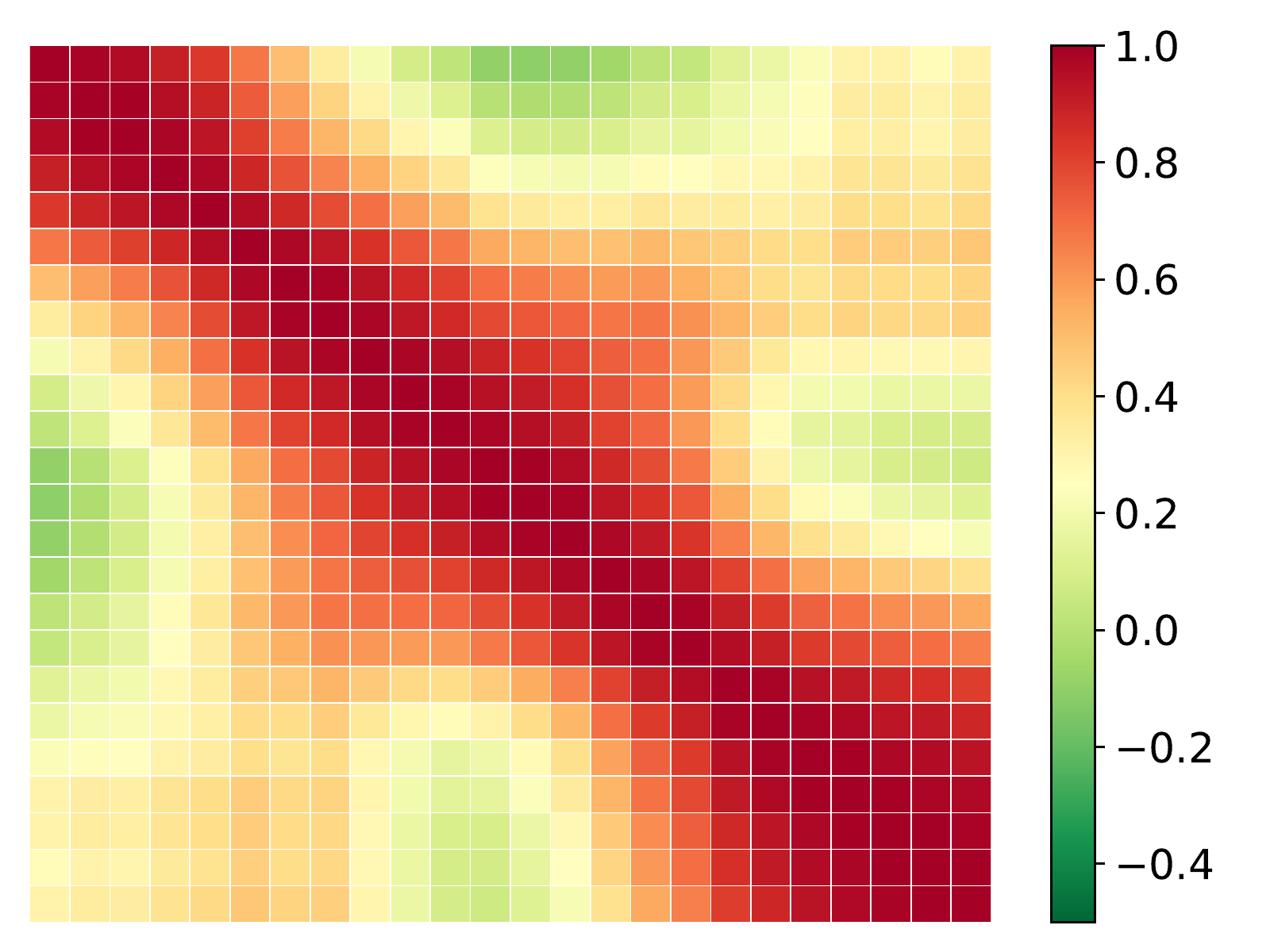}
		\caption{GAN.}
	\end{subfigure}
	\begin{subfigure}{.25\textwidth}
		\centering
		\includegraphics[width=\linewidth]{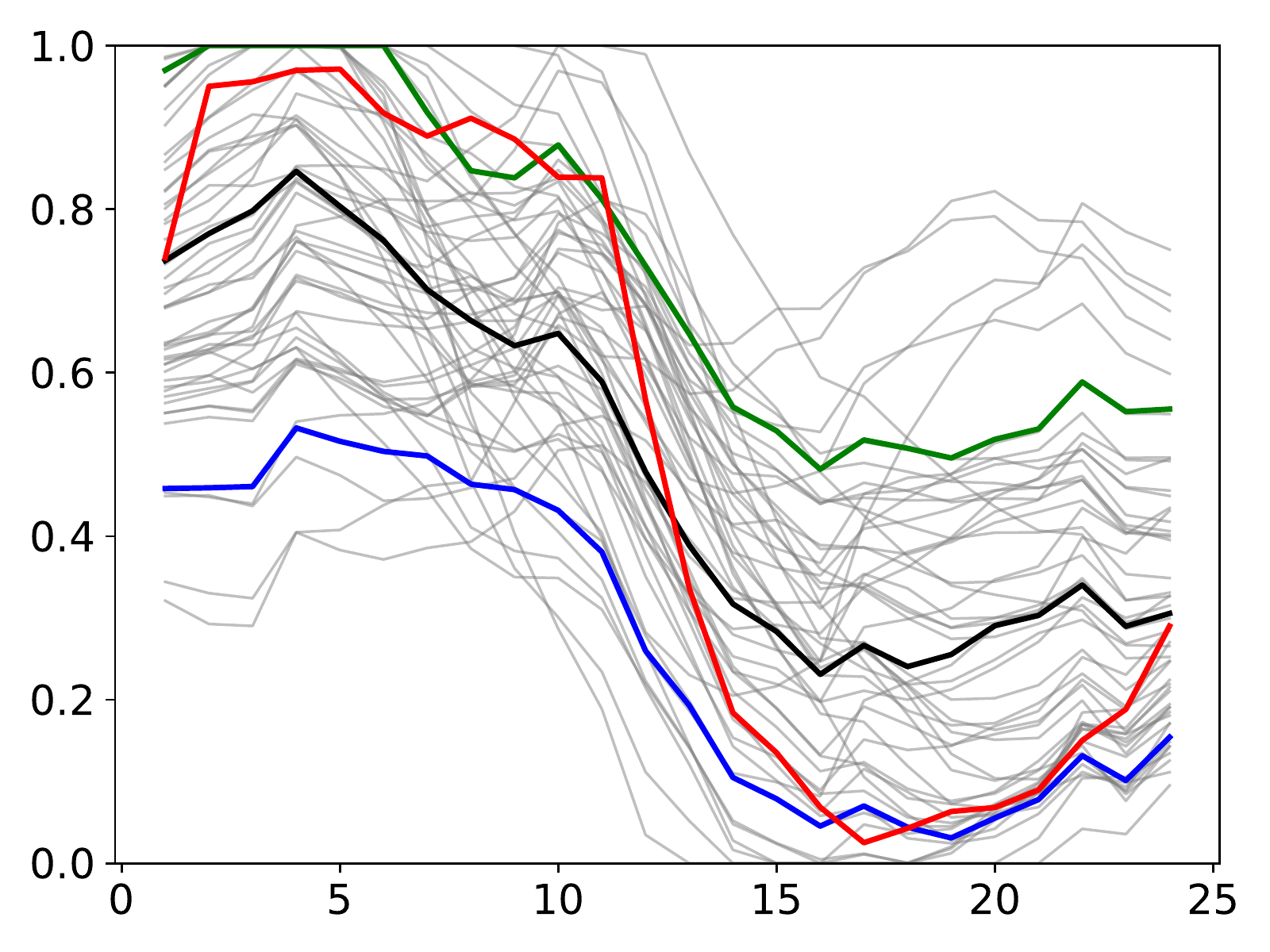}
		\caption{VAE.}
	\end{subfigure}%
	\begin{subfigure}{.25\textwidth}
		\centering
		\includegraphics[width=\linewidth]{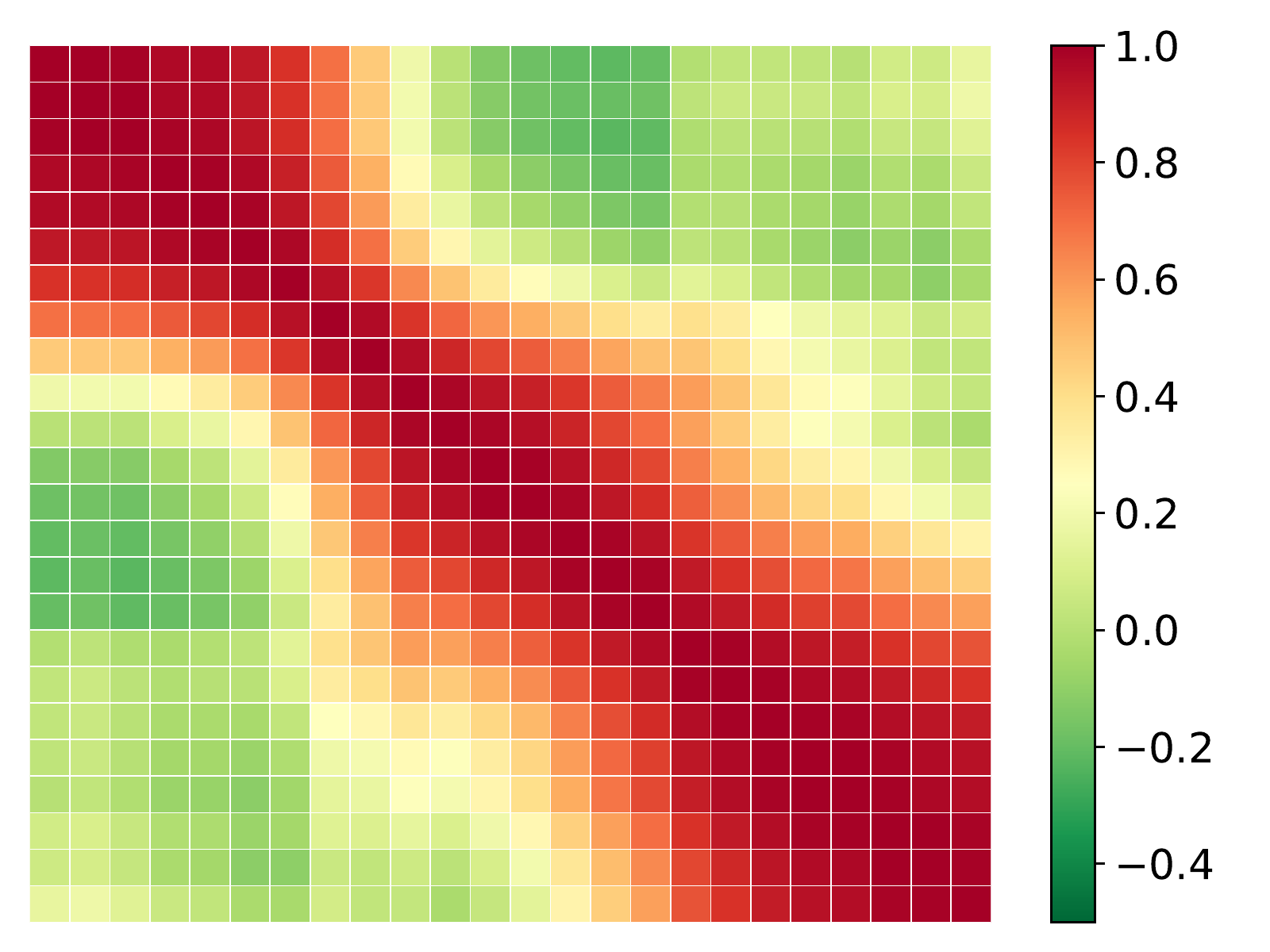}
		\caption{VAE.}
	\end{subfigure}
	\caption{Wind power scenarios shape comparison and analysis. \\
	Left part (a) NF, (c) GAN, and (e) VAE: 50 wind power scenarios (grey) of a randomly selected day of the testing set along with the ten \% (blue), 50 \% (black), and 90 \% (green) quantiles, and the observations (red). Right part (b) NF, (d) GAN, and (f) VAE: the corresponding Pearson time correlation matrices of these scenarios with the periods as rows and columns. The NF tends to exhibit no time correlation between scenarios. In contrast, the VAE and GAN tend to be partially time-correlated over a few periods.}
	\label{fig:wind_scenarios}
\end{figure}

The left part of Figure~\ref{fig:wind_scenarios} provides 50 scenarios, (a) NF, (c) GAN, and (e) VAE, generated for a given day selected randomly from the testing set. Notice how the shape of the NF's scenarios differs significantly from the GAN and VAE as they tend to be more variable with no identifiable trend. In contrast, the VAE and GAN scenarios seem to differ mainly in nominal power but have similar shapes. This behavior is even more pronounced for the GAN, where the scenarios rarely crossed over periods. For instance, there is a gap in generation around periods 17 and 18 where all the GAN's scenarios follow this trend.
These observations are confirmed by computing the corresponding time correlation matrices, depicted by the right part of Figure~\ref{fig:wind_scenarios} demonstrating there is no correlation between NF's scenarios. On the contrary, the VAE and GAN correlation matrices tend to be similar with a time correlation of the scenarios over a few periods, with more correlated periods when considering the GAN. This difference in the scenario's shape is striking and not necessarily captured by metrics such as the CRPS, QS, or even the classifier-based metric and is also observed on the PV and load tracks, as explained in the next paragraph.
%
%
%
%
%

\subsubsection*{All tracks}\label{sec:all_tracks_res}

Table~\ref{tab:quality_average_scores} provides the averaged quality scores. The CRPS is averaged over the 24 time periods $\overline{\text{CRPS}}$. The QS over the 99 percentiles $\overline{\text{QS}}$. The MAE-r is the mean absolute error between the reliability curve and the diagonal, and $\overline{\text{AUC}}$ is the mean of the 50 AUC.
\begin{table}[htbp]
\renewcommand{\arraystretch}{1.25}
\begin{center}
\begin{tabular}{m{0.5cm}lrrrr}
			\hline  \hline
& &  NF &  VAE & GAN & RAND \\ \hline
\multirow{6}{*}{Wind}
& $\overline{\text{CRPS}}$  & 9.07  & \textbf{8.80} & 9.79 & 16.92 \\ 
& $\overline{\text{QS}}$    & 4.58  & \textbf{4.45} & 4.95 & 8.55 \\ 
& MAE-r &  2.83    &  \textbf{2.67} &  6.82  & 1.01  \\ 
& $\overline{\text{AUC}}$  & 0.935   &  \textbf{0.877} &  0.972 & 0.918 \\
& ES & 56.71 & \textbf{54.82} & 60.52 & 96.15 \\
& VS & 18.54 & \textbf{17.87} & 19.87 & 23.21 \\  \hline 
\multirow{6}{*}{PV}
& $\overline{\text{CRPS}}$  & \textbf{2.35}  & 2.60 & 2.61 & 4.92 \\ 
& $\overline{\text{QS}}$    & \textbf{1.19}  & 1.31 & 1.32 & 2.48 \\ 
& MAE-r &  \textbf{2.66}    & 9.04  &  4.94  & 3.94  \\ 
& $\overline{\text{AUC}}$  & \textbf{0.950}  & 0.969 &  0.997 & 0.947 \\
& ES & \textbf{23.08} & 24.65 & 24.15 & 41.53 \\
& VS & \textbf{4.68} & 5.02 & 4.88 & 13.40 \\  \hline 
\multirow{6}{*}{Load}
& $\overline{\text{CRPS}}$  & \textbf{1.51}  & 2.74 & 3.01 & 6.74 \\ 
& $\overline{\text{QS}}$    & \textbf{0.76}  & 1.39 & 1.52 & 3.40 \\ 
& MAE-r &  \textbf{7.70}    & 13.97  &  9.99  & 0.88  \\ 
& $\overline{\text{AUC}}$  & \textbf{0.823}  & 0.847 &  0.999 & 0.944 \\
& ES & \textbf{9.17} & 15.11 & 17.96 & 38.08 \\
& VS & \textbf{1.63} & 1.66 & 3.81 & 7.28 \\  \hline  \hline
\end{tabular}
\caption{Averaged quality scores per dataset. \\
The best performing deep learning generative model for each track is written in bold. 
The CRPS, QS, MAE-r, and ES are expressed in \%. Overall, for both the PV and load tracks, the NF outperforms the VAE and GAN and slightly outperforms the VAE on the wind track.
}
\label{tab:quality_average_scores}
	\end{center}
\end{table}
%
Overall, for the PV and load tracks in CRPS, QS, reliability diagrams, AUC, ES, and VS, the NF outperforms the VAE and GAN and is slightly outperformed by the VAE on the wind track. On the load track, the VAE outperforms the GAN. However, the VAE and GAN achieved similar results on the PV track, and the GAN performed better in terms of ES and VS.
These results are confirmed by the DM tests depicted in Figure~\ref{fig:DM-test-pv-load}.
%
The classifier-based metric results for both the load and PV tracks, provided by Figure~\ref{fig:classifier-pv-load}, confirm this trend where the NF is the best to trick the classifier followed by the VAE and GAN. 

Similar to the wind track, the shape of the scenarios differs significantly between the NF and the other models for both the load and PV tracks as indicated by the left part of Figures~\ref{fig:pv_scenarios} and~\ref{fig:load_scenarios}, and the corresponding correlation matrices provided by the right part of Figures~\ref{fig:pv_scenarios} and~\ref{fig:load_scenarios}. Note: the load track scenarios are highly correlated for both the VAE and GAN. Finally, Figure \ref{fig:average-correlation} provides the average of the correlation matrices over all days of the testing set for each dataset. The trend depicted above is confirmed.
This difference between the NF and the other generative model may be explicated by the design of the methods. The NF explicitly learns the probability density function (PDF) of the multi-dimensional random variable considered. Thus, the NF scenarios are generated according to the learned PDF producing multiple shapes of scenarios. In contrast, the generator of the GAN is trained to fool the discriminator, and it may find a shape particularly efficient leading to a set of similar scenarios. Concerning the VAE, it is less obvious. However, by design, the decoder is trained to generate scenarios from the latent space assumed to follow a Gaussian distribution that may lead to less variability.

\subsection{Value results}\label{sec:value_res}

The energy retailer portfolio comprises wind power, PV generation, load, and a battery energy storage device. The 50 days of the testing set are used and combined with the 30 possible PV and wind generation zones (three PV zones and ten wind farms), resulting in 1 500 independent simulated days. 
%
A two-step approach is employed to evaluate the forecast value:
\begin{itemize}
    \item First, for each generative model and the 1 500 days simulated, the two-stage stochastic planner computes the day-ahead bids of the energy retailer portfolio using the PV, wind power, and load scenarios. After solving the optimization, the day-ahead decisions are recorded.
    \item Then, a real-time dispatch is carried out using the PV, wind power, and load observations, with the day-ahead decisions as parameters.
\end{itemize}
This two-step methodology is applied to evaluate the three generative models, namely the NF, GAN, and VAE.
\begin{figure}[tb]
	\centering
    \includegraphics[width=\linewidth]{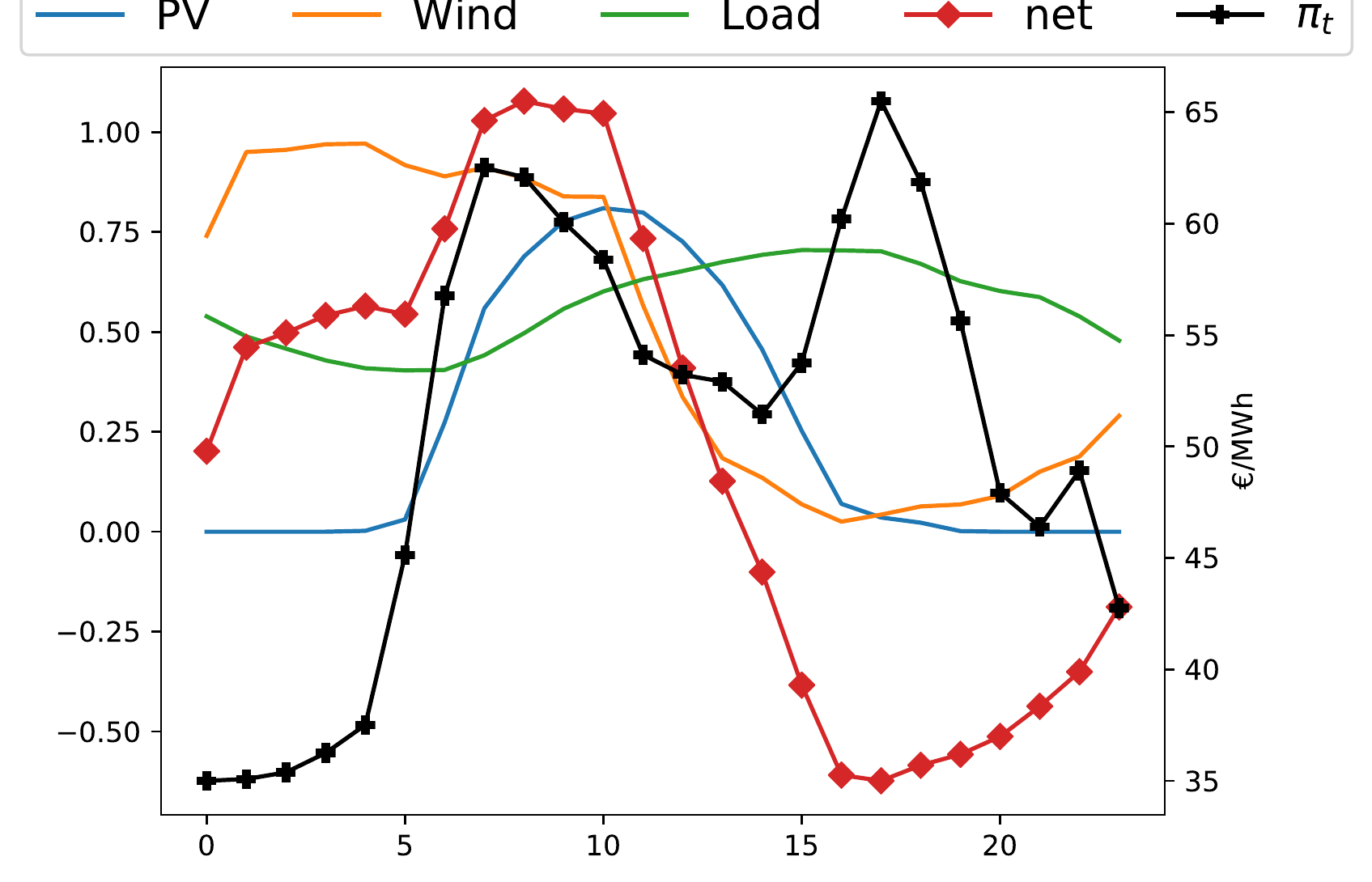}
	\caption{Energy retailer case study: illustration of the observations on a random day of the testing set. \\
	The energy retailer portfolio comprises PV generation, wind power, load, and storage device. The PV, wind power, and load scenarios from the testing set are used as inputs of the stochastic day-ahead planner to compute the optimal bids. The net is the power balance of the energy retailer portfolio. The day-ahead prices $\pi_t$ are obtained from the Belgian day-ahead market on $\daddate$.}
	\label{fig:value_dataset}
\end{figure}
%
Figure~\ref{fig:value_dataset} illustrates an arbitrary random day of the testing set with the first zone for both the PV and wind. $\pi_t$~[\euro / MWh] is the day-ahead prices on $\daddate$ of the Belgian day-ahead market used for the 1 500 days simulated. The negative $\bar{q}_t$ and positive $\bar{\lambda}_t$ imbalance prices are set to $ 2 \times \pi_t$, $\forallt$.
The retailer aims to balance the net power, the red curve in Figure~\ref{fig:value_dataset}, by importing/exporting from/to the main grid. Usually, the net is positive (negative) at noon (evening) when the PV generation is maximal (minimal), and the load is minimal (maximal). As the day-ahead spot price is often maximal during the evening load peak, the retailer seeks to save power during the day by charging the battery to decrease the import during the evening. Therefore, the more accurate the PV, wind generation, and load scenarios are, the better is the day-ahead planning.

The battery minimum $s^\text{min}$ and maximum $s^\text{max}$ capacities are 0 and 1, respectively. It is assumed to be capable of fully (dis)charging in two hours with $y^\text{dis}_\text{max} = y^\text{cha}_\text{max} = s^\text{max} / 2$, and the (dis)charging efficiencies are $\eta^\text{dis} =\eta^\text{cha} = 95$ \%. Each simulation day is independent with a fully discharged battery at the first and last period of each day $s^\text{ini} = s^\text{end} = 0$. 
The 1 500 stochastic optimization problems are solved with 50 PV, wind generation, and load scenarios. The python Gurobi library is used to implement the algorithms in Python 3.7, and Gurobi\footnote{\url{https://www.gurobi.com/}} 9.0.2 is used to solve the optimization problems. Numerical experiments are performed on an Intel Core i7-8700 3.20 GHz based computer with 12 threads and 32 GB of RAM running on Ubuntu 18.04 LTS. 

The net profit, that is, the profit minus penalty, is computed for the 1 500 days of the simulation and aggregated in the first row of Table~\ref{tab:value-res}. The ranking of each model is computed for the 1 500 days. The cumulative ranking is expressed in terms of percentage in Table~\ref{tab:value-res}. NF outperformed both the GAN and VAE with a total net profit of 107 k\euro. There is still room for improvement as the oracle, which has perfect future knowledge, achieved 300 k\euro. NF ranked first 39.0 \% during the 1 500 simulation days and achieved the first and second ranks 69.6 \%.
\begin{table}[htbp]
\renewcommand{\arraystretch}{1.25}
	\begin{center}
		\begin{tabular}{lrrrr}
			\hline \hline
			        &  NF & VAE & GAN \\ \hline
			Net profit (k\euro) &  \textbf{107}  &  97    &  93\\   \hline
			1 (\%)  &  \textbf{39.0} &  31.8  &  29.2\\  
			1 \& 2 (\%)  &  \textbf{69.6}  &  68.3  &  62.1\\  
			1 \& 2 \& 3 (\%)  &  100  &  100   &  100\\  \hline \hline
		\end{tabular}
		\caption{Total net profit (k\euro) and cumulative ranking (\%). \\
Using the NF PV, wind power, and load scenarios, the stochastic planner achieved the highest net profit with 107 k\euro, ranked first 39.0 \%, second 30.6 \%, and third 30.4 \% over 1 500 days of simulation. The second-best model, the VAE, achieved a net profit of 97 k\euro, ranked first 31.8 \%, second 36,5 \%, and third 31.7 \%.}
		\label{tab:value-res}
	\end{center}
\end{table}
Overall, in terms of forecast value, the NF outperforms the VAE and GAN. However, this case study is "simple," and stochastic optimization relies mainly on the quality of the average of the scenarios. Therefore, one may consider taking advantage of the particularities of a specific method by considering more advanced case studies. In particular, the specificity of the NFs to provide direct access to the probability density function may be of great interest in specific applications. It is left for future investigations as more advanced case studies would prevent a fair comparison between models.

\subsection{Results summary}\label{sec:discussion}

Table \ref{tab:comparison} summarizes the main results of this study by comparing the VAE, GAN, and NF implemented through easily comparable star ratings. The rating for each criterion is determined using the following rules - 1 star: third rank, 2 stars: second rank, and 3 stars: first rank. Specifically, training speed is assessed based on reported total training times for each dataset: PV generation, wind power, and load; sample speed is based on reported total generating times for each dataset; quality is evaluated with the metrics considered; value is based on the case study of the day-ahead bidding of the energy retailer; the hyper-parameters search is assessed by the number of configurations tested before reaching satisfactory and stable results over the validation set; the hyper-parameters sensitivity is evaluated by the impact on the quality metric of deviations from the optimal the hyper-parameter values found during the hyper-parameter search; the implementation-friendly criterion is appraised regarding the complexity of the technique and the amount of knowledge required to implement it.
\begin{table}[htbp]
\renewcommand{\arraystretch}{1.25}
\begin{center}
\begin{tabular}{lccc}
\hline \hline
Criteria			     & VAE          &  GAN         & NF            \\ \hline
Train speed              & $\nstars[3]$ & $\nstars[2]$ & $\nstars[1]$    \\
Sample speed             & $\nstars[3]$ & $\nstars[2]$ &  $\nstars[1]$   \\
Quality                  & $\nstars[2]$ & $\nstars[1]$ & $\nstars[3]$    \\
Value                    & $\nstars[2]$ & $\nstars[1]$ & $\nstars[3]$    \\
Hp search                & $\nstars[2]$ & $\nstars[1]$ & $\nstars[3]$  \\
Hp sensibility 	         & $\nstars[2]$ & $\nstars[1]$ & $\nstars[3]$  \\
Implementation           & $\nstars[3]$ & $\nstars[2]$ & $\nstars[1]$  \\
\hline \hline
\end{tabular}
\caption{Comparison between the deep generative models. \\
The rating for each criterion is determined using the following rules - 1 star: third rank, 2 stars: second rank, and 3 stars: first rank.
Train speed: training computation time; Sample speed: scenario generation computation time; Quality: forecast quality based on the eight complementary metrics considered; Value: forecast value based on the day-ahead energy retailer case study; Hp search: assess the difficulty to identify relevant hyper-parameters; Hp sensibility: assess the sensitivity of the model to a given set of hyper-parameters (the more stars, the more robust to hyper-parameter modifications); Implementation: assess the difficulty to implement the model (the more stars, the more implementation-friendly). Note: the justifications are provided in \ref{annex:table2}.
}
\label{tab:comparison}
\end{center}
\end{table}

\section{Conclusion}\label{sec:conclusion}

This paper proposed a fair and thorough comparison of normalizing flows with the state-of-the-art deep learning generative models, generative adversarial networks, and variational autoencoders, both in quality and value. The experiments adopt the open data of the Global Energy Forecasting Competition 2014, where the generative models use the conditional information to compute improved weather-based PV power, wind power, and load scenarios. The results demonstrate that normalizing flows can challenge generative adversarial networks and variational autoencoders. Overall, they are more accurate in quality and value and can be used effectively by non-expert deep learning practitioners. 
In addition, normalizing flows have several advantages over more traditional deep learning approaches that should motivate their introduction into power system applications:
\begin{enumerate}[label=(\roman*)]
	\item Normalizing flows directly learn the stochastic multivariate distribution of the underlying process by maximizing the likelihood. In contrast to variational autoencoders and generative adversarial networks, they provide access to the exact likelihood of the model's parameters. Hence, they offer a sound and direct way to optimize the network parameters. It may open a new range of advanced applications benefiting from this advantage. For instance, to transfer scenarios from one location to another based on the knowledge of the probability density function. A second application is the importance sampling for stochastic optimization based on a scenario approach. Indeed, normalizing flows provide the likelihood of each generated scenario, making it possible to filter relevant scenarios used in stochastic optimization. 
	\item In our opinion, normalizing flows are easier to use by non-expert deep learning practitioners once the libraries are available. Indeed, they are more reliable and robust in terms of hyper-parameters selection. Generative adversarial networks and variational autoencoders are particularly sensitive to the latent space dimension, the structure of the neural networks, the learning rate, \textit{etc}. Generative adversarial networks convergence, by design, is unstable, and for a given set of hyper-parameters, the scenario's quality may differ completely. In contrast, it was easier to retrieve relevant normalizing flows hyper-parameters by manually testing a few sets of values that led to satisfying training convergence and quality results.  
\end{enumerate}
Nevertheless, their usage as a base component of the machine learning toolbox is still limited compared to generative adversarial networks or variational autoencoders.

\section{Acknowledgments}\label{sec:acknowledgments}

The authors would like to acknowledge the authors and contributors of the Scikit-Learn \citep{sklearn_api}, Pytorch \citep{NEURIPS2019_9015}, and Weights \& Biases \citep{wandb} Python libraries. In addition, the authors would like to thank the editor and the reviewers for the comments that helped improve the paper.
Antoine Wehenkel is a recipient of an F.R.S.- FNRS fellowship and acknowledges the financial support of the FNRS (Belgium). Antonio Sutera is supported via the Energy Transition Funds project EPOC 2030-2050 organized by the FPS economy, S.M.E.s, Self-employed and Energy.

\bibliographystyle{elsarticle-num-names}  
\bibliography{biblio}

\appendix

\section{Additional arguments}\label{annex:arguments}

\subsection{Table \ref{tab:contributions} justifications}\label{annex:table1}

\citet{wang2020modeling} use a Wasserstein GAN with gradient penalty to model both the uncertainties and the variations of the load. First, point forecasting is conducted, and the corresponding residuals are derived. Then, the GAN generates residual scenarios conditional on the day type, temperatures, and historical loads. 
The GAN model is compared with the same version without gradient penalty and two quantile regression models: random forest and gradient boosting regression tree. 
The quality evaluation is conducted on open load datasets from the Independent System Operator-New England\footnote{\url{https://www.iso-ne.com/}} with five metrics: (1) the continuous ranked probability score; (2) the quantile score; (3) the Winkler score; (4) reliability diagrams; (5) Q-Q plots. Note: the forecast value is not assessed. \\

\citet{qi2020optimal} propose a concentrating solar power (CSP) configuration method to determine the CSP capacity in multi-energy power systems. The configuration model considers the uncertainty by scenario analysis. A $\beta$ VAE generates the scenarios. It is an improved version of the original VAE
However, it does not consider weather forecasts, and the model is trained only by using historical observations.
The quality evaluation is conducted on two wind farms and six PV plants using three metrics. (1) The leave-one-out accuracy of the 1-nearest neighbor classifier. (2) The comparison of the frequency distributions of the actual data and the generated scenarios. (3) Comparing the spatial and temporal correlations of the actual data and the scenarios by computing Pearson correlation coefficients. 
The value is assessed by considering the case study of the CSP configuration model, where the $\beta$ VAE is used to generate PV, wind power, and load scenarios. However, the VAE is not compared to another generative model for both the quality and value evaluations. 
Note: the dataset does not seem to be in open-access. 
Finally, the value evaluation case study is not trivial due to the mathematical formulation that requires a certain level of knowledge of the system. Thus, the replicability criterion is partially satisfied. \\

\citet{ge2020modeling} compared NFs to VAEs and GANs for the generation of daily load profiles. The models do not take into account weather forecasts but only historical observations. However, an example is given to illustrate the principle of generating conditional daily load profiles by using three groups: light load, medium load, and heavy load.
The quality evaluation uses five indicators. Four to assess the temporal correlation: (1) probability density function; (2) autocorrelation function;  (3) load duration curve; (4) a wave rate is defined to evaluate the volatility of the daily load profile. Furthermore, one additional for the spatial correlation: (5) Pearson correlation coefficient is used to measure the spatial correlation among multiple daily load profiles.
The simulations use the open-access London smart meter and Spanish transmission service operator datasets of Kaggle.
The forecast value is not assessed.

\subsection{Table \ref{tab:comparison} justifications}\label{annex:table2}

The VAE is the fastest to train, with a recorded computation time of 7 seconds on average per dataset. The training time of the GAN is approximately three times longer, with an average computation time of 20 seconds per dataset. Finally, the NF is the slowest, with an average training time of 4 minutes. This ranking is preserved with the VAE the fastest concerning the sample speed, followed by the GAN and NF models. The VAE and the GAN generate the samples over the testing sets, 5 000 in total, in less than a second. However, the NF considered takes a few minutes. In contrast, the affine autoregressive version of the NF is much faster to train and generate samples. Note: even a training time of a few hours is compatible with day-ahead planning applications. In addition, once the model is trained, it is not necessarily required to retrain it every day.

The quality and value assessments have already been discussed in section \ref{sec:numerical_results}. Overall, the NF outperforms both the VAE and GAN models.

Concerning the hyper-parameters search and sensibility, the NF tends to be the most accessible model to calibrate. Compared with the VAE and GAN, we found relevant hyper-parameter values by testing only a few combinations. In addition, the NF is robust to hyper-parameter modifications. In contrast, the GAN is the most sensitive. Variations of the hyper-parameters may result in very poor scenarios both in terms of quality and shape. Even for a fixed set of hyper-parameters values, two separate training may not converge towards the same results illustrating the GAN training instabilities. The VAE is more straightforward to train than the GAN but is also sensitive to hyper-parameters values. However, it is less evident than the GAN.

Finally, we discuss the implementation-friendly criterion of the models. Note: this discussion is only valid for the models implemented in this study. There exist various architectures of GANs, VAEs, and NFs with simple and complex versions.
In our opinion, the VAE is the effortless model to implement as the encoder and decoder are both simple feed-forward neural networks. The only difficulty lies in the reparameterization trick that should be carefully addressed. The GAN is a bit more difficult to deploy due to the gradient penalty to handle but is similar to the VAE with both the discriminator and the generator that are feed-forward neural networks. The NF is the most challenging model to implement from scratch because the UMNN-MAF approach requires an additional integrand network. An affine autoregressive NF is easier to implement. However, it may be less capable of modeling the stochasticity of the variable of interest. 
However, forecasting practitioners do not necessarily have to implement generative models from scratch and can use numerous existing Python libraries. 

\section{Background}\label{annex:background}

\subsection{Normalizing flows}\label{annex:nf}

\subsubsection*{Normalizing flow computation}

\noindent Evaluating the likelihood of a distribution modeled by a normalizing flow requires computing (\ref{eq:change_formula}), \textit{i.e.}, the normalizing direction, as well as its log-determinant. Increasing the number of sub-flows by $K$ of the transformation results in only $\mathcal{O}(K)$ growth in the computational complexity as the log-determinant of $ J_{f_\theta} $ can be expressed as
\begin{subequations}
\label{eq:jacobian_composite}	
\begin{align}
\log |\det J_{f_\theta}(\mathbf{x}) | & = \log \bigg| \prod_{k=1}^K \det J_{f_{k,\theta}}(\mathbf{x}) \bigg|, \\
                             & = \sum_{k=1}^K \log| \det J_{f_{k,\theta}}(\mathbf{x})|.
\end{align}
\end{subequations}
However, with no further assumption on $f_\theta$, the computational complexity of the log-determinant is $\mathcal{O}(K \cdot T^3)$, which can be intractable for large $T$. Therefore, the efficiency of these operations is essential during training, where the likelihood is repeatedly computed. There are many possible implementations of NFs detailed by \citet{papamakarios2019normalizing,kobyzev2020normalizing} to address this issue. 

\subsubsection*{Autoregressive flow}

\noindent The Jacobian of the autoregressive transformation $f_\theta$ defined by (\ref{eq:f_autoregressive}) is lower triangular, and its log-absolute-determinant is
\begin{subequations}
\label{eq:jacobian_autoregressive}	
\begin{align}
	\log | \det J_{f_\theta}(\mathbf{x})  |& = \log  \prod_{i=1}^T \bigg|\dfrac{\partial f^i}{\partial x_i} (x_i; h^i) \bigg|, 	\\
	& = \sum_{i=1}^T \log \bigg|\dfrac{\partial f^i}{\partial x_i} (x_i; h^i) \bigg|,
\end{align}
\end{subequations}
that is calculated in $\mathcal{O}(T)$ instead of $\mathcal{O}(T^3)$. 

\subsubsection*{Affine autoregressive flow}

\noindent A simple choice of transformer is the class of affine functions
\begin{align}
\label{eq:f_affin_autoregressive}	
	f^i(x_i; h^i) & = \alpha_i x_i + \beta_i ,
\end{align}
where $f^i(\cdot; h^i ) : \mathbb{R} \rightarrow \mathbb{R}$ is parameterized by $h^i = \{ \alpha_i,  \beta_i\}$, $\alpha_i$ controls the scale, and $\beta_i$ controls the location of the transformation. Invertibility is guaranteed if $\alpha_i \ne 0$, and this can be easily achieved by \textit{e.g.} taking $\alpha_i = \exp{(\tilde{\alpha}_i)}$, where $\tilde{\alpha}_i$ is an unconstrained parameter in which case $h^i = \{ \tilde{\alpha}_i,  \beta_i\}$. The derivative of the transformer with respect to $x_i$ is equal to $\alpha_i$. Hence the log-absolute-determinant of the Jacobian becomes
\begin{align}
\label{eq:jacobian_affine_autoregressive}	
	\log | \det J_{f_\theta}(\mathbf{x})  |& = \sum_{i=1}^T \log |\alpha_i | =  \sum_{i=1}^T \tilde{\alpha}_i.
\end{align}
Affine autoregressive flows are simple and computation efficient. However, they are limited in expressiveness requiring many stacked flows to represent complex distributions. It is unknown whether affine autoregressive flows with multiple layers are universal approximators \citep{papamakarios2019normalizing}.

\subsection{Variational autoencoders}\label{annex:vae}

\subsubsection*{Gradients computation}
\noindent By using (\ref{eq:variational_lower_bound}) $\mathcal{L}_{\theta, \varphi}$ is decomposed in two parts
\begin{align}
\label{eq:lower_bound}	
\mathcal{L}_{\theta, \varphi}(\mathbf{x},\mathbf{c}) = & \underset{q_\varphi(\mathbf{z}|\mathbf{x},\mathbf{c})}{\mathbb{E}} [\log p_\theta(\mathbf{x}|\mathbf{z},\mathbf{c}) ] -\text{KL}[ q_\varphi(\mathbf{z}|\mathbf{x},\mathbf{c}) || p(\mathbf{z})] .
\end{align}
$\nabla_\theta  \mathcal{L}_{\theta, \varphi}$ is estimated with the usual Monte Carlo gradient estimator. However, the estimation of $\nabla_\varphi  \mathcal{L}_{\theta, \varphi}$ requires the reparameterization trick proposed by \citet{kingma2013auto}, where the random variable $\mathbf{z}$ is re-expressed as a deterministic variable 
\begin{align}
	\label{eq:reparameterization_trick1}	
	\mathbf{z} = & g_\varphi(\epsilon, \mathbf{x}),
\end{align}
with $\epsilon$ an auxiliary variable with independent marginal $p_\epsilon$, and $g_\varphi(\cdot)$ some vector-valued function parameterized by $\varphi$.
%
Then, the first right hand side of (\ref{eq:lower_bound}) becomes
\begin{align}
\label{eq:RHS_2}	
\underset{q_\varphi(\mathbf{z}|\mathbf{x},\mathbf{c})}{\mathbb{E}} [\log p_\theta(\mathbf{x}|\mathbf{z},\mathbf{c})]  = &  \underset{p(\epsilon)}{\mathbb{E}} [\log p_\theta(\mathbf{x}|g_\varphi(\epsilon, \mathbf{x}),\mathbf{c})].
\end{align}
$\nabla_\varphi  \mathcal{L}_{\theta, \varphi}$ is now estimated with Monte Carlo integration.

\subsubsection*{Conditional variational autoencoders implemented}

\noindent Following \citet{kingma2013auto}, we implemented Gaussian multi-layered perceptrons (MLPs) for both the encoder $\text{NN}_\varphi$ and decoder $\text{NN}_\theta$. In this case, $p(\mathbf{z})$ is a centered isotropic multivariate Gaussian, $p_\theta(\mathbf{x}|\mathbf{z}, \mathbf{c})$ and $q_\varphi(\mathbf{x}|\mathbf{z},\mathbf{c})$ are both multivariate Gaussian with a diagonal covariance and parameters $\boldsymbol{\mu}_\theta,\boldsymbol{\sigma}_\theta$ and $\boldsymbol{\mu}_\varphi, \boldsymbol{\sigma}_\varphi$, respectively. Note: there is no restriction on the encoder and decoder architectures, and they could as well be arbitrarily complex convolutional networks. Under these assumptions, the conditional VAE implemented is
\begin{subequations}
	\label{eq:VAE_NN_assumptions}	
	\begin{align}
	p(\mathbf{z}) & = \mathcal{N} (\mathbf{z}; \mathbf{0}, \mathbf{I}), \\
	p_\theta(\mathbf{x}|\mathbf{z},\mathbf{c}) & = \mathcal{N}(\mathbf{x};\boldsymbol{\mu}_\theta,\boldsymbol{\sigma}_\theta^2 \mathbf{I}), \\
	q_\varphi(\mathbf{z}|\mathbf{x},\mathbf{c}) & = \mathcal{N}(\mathbf{z};\boldsymbol{\mu}_\varphi,\boldsymbol{\sigma}_\varphi^2\mathbf{I} ), \\
	\boldsymbol{\mu}_\theta, \log \boldsymbol{\sigma}_\theta^2 & =\text{NN}_\theta (\mathbf{x},\mathbf{c}), \\
	\boldsymbol{\mu}_\varphi, \log \boldsymbol{\sigma}_\varphi^2 & =\text{NN}_\varphi (\mathbf{z},\mathbf{c}).
	\end{align}
\end{subequations}
%
Then, by using the valid reparameterization trick proposed by \citet{kingma2013auto}
\begin{subequations}
	\label{eq:reparameterization_trick2}	
	\begin{align}
	\boldsymbol{\epsilon} \sim & \mathcal{N}(\mathbf{0},\mathbf{I}) ,\\
	\mathbf{z} : = & \boldsymbol{\mu}_\varphi + \boldsymbol{\sigma}_\varphi \boldsymbol{\epsilon},
	\end{align}
\end{subequations}
$\mathcal{L}_{\theta, \varphi}$ is computed and differentiated without estimation using the expressions
\begin{subequations}
\label{eq:lower_bound_exact}	
\begin{align}
\text{KL} [ q_\varphi(\mathbf{z}|\mathbf{x},\mathbf{c}) || p(\mathbf{z})] = & -\frac{1}{2} \sum_{j=1}^d (1 + \log \boldsymbol{\sigma}_{\varphi,j}^2 - \boldsymbol{\mu}_{\varphi,j}^2 -\boldsymbol{\sigma}_{\varphi,j}^2) , \\
\underset{p(\boldsymbol{\epsilon})}{\mathbb{E}} [\log p_\theta(\mathbf{x}|\mathbf{z},\mathbf{c})] \approx & -\frac{1}{2} \left \Vert \frac{\mathbf{x} - \boldsymbol{\mu}_\theta}{\boldsymbol{\sigma}_\theta} \right \Vert^2 ,
\end{align}
\end{subequations}
with $d$ the dimensionality of $\mathbf{z}$.

\subsection{Generative adversarial network}\label{annex:gan}

\subsubsection*{Original generative adversarial network}

\noindent The original GAN value function $V(\phi, \theta)$ proposed by \citet{goodfellow2014generative} is
\begin{subequations}
\label{eq:GAN_value_function}	
\begin{align}
V(\phi, \theta) = &   \underbrace{\underset{\mathbf{x}}{\mathbb{E}}  [\log d_\phi (\mathbf{x}|\mathbf{c})]  + \underset{\mathbf{\hat{x}}}{\mathbb{E}} [\log(1-d_\phi(\mathbf{\hat{x}}|\mathbf{c}))]}_{:=- L_d} , \\
L_g  := &  -\underset{\mathbf{\hat{x}}}{\mathbb{E}} [\log(1-d_\phi(\mathbf{\hat{x}}|\mathbf{c}))],
\end{align}
\end{subequations}
where $L_d$ is the cross-entropy, and $L_g$ the probability the discriminator wrongly classifies the samples.

\subsubsection*{Wasserstein generative adversarial network}

\noindent 
The divergences which GANs typically minimize are responsible for their training instabilities for reasons investigated theoretically by \citet{arjovsky2017towards}. 
\citet{arjovsky2017wasserstein} proposed instead using the \textit{Earth mover} distance, also known as the Wasserstein-1 distance
\begin{align}
\label{eq:Wasserstein_distance}	
W_1(p,q) = & \inf_{\gamma \in \Pi(p,q)} \mathbb{E}_{(x,y)\sim \gamma}  [\left \Vert x - y \right \Vert ],
\end{align}
where $\Pi(p,q)$ denotes the set of all joint distributions $\gamma(x,y)$ whose marginals are
respectively $p$ and $q$, $\gamma(x,y)$ indicates how much mass must be transported from $x$ to $y$ in order to transform the distribution $p$ into $q$, $\left \Vert \cdot \right \Vert$ is the L1 norm, and $\left \Vert x - y \right \Vert$ represents the cost of moving a unit of mass from $x$ to $y$. 
%
However, the infimum in (\ref{eq:Wasserstein_distance}) is intractable. Therefore, \citet{arjovsky2017wasserstein} used the Kantorovich-Rubinstein duality \citep{villani2008optimal} to propose the Wasserstein GAN (WGAN) by solving the min-max problem
\begin{align}
\label{eq:GAN_wasserstein}	
 \theta^\star = \arg \min_\theta \max_{\phi \in \mathcal{W}} & \ \underset{\mathbf{x}}{\mathbb{E}}  [ d_\phi (\mathbf{x}|\mathbf{c})]  - \underset{\mathbf{\hat{x}}}{\mathbb{E}} [ d_\phi (\mathbf{\hat{x}}|\mathbf{c}) ],
\end{align}
where $\mathcal{W} = \{\phi : \Vert d_\phi(\cdot) \Vert_L \leq 1 \}$ is the 1-Lipschitz space, and the classifier $d_\phi(\cdot) : \mathbb{R}^T \times \mathbb{R}^{|\mathbf{c}|} \rightarrow [0, 1]$ is replaced by a critic function $d_\phi(\cdot) : \mathbb{R}^T \times \mathbb{R}^{|\mathbf{c}|}  \rightarrow \mathbb{R}$.
%
%
%
%
However, the weight clipping used to enforce $d_\phi$ 1-Lipschitzness can lead sometimes the WGAN to generate only poor samples or failure to converge \citep{gulrajani2017improved}. Therefore, we implemented the WGAN-GP to tackle this issue. 

\subsection{Hyper-parameters}\label{annex:hp}

\noindent Table~\ref{tab:model_hp} provides the hyper-parameters of the NF, VAE, and GAN implemented. The Adam optimizer \citep{kingma2014adam} is used to train the generative models with a batch size of 10 \% of the learning set.
\begin{table}[htbp]
\renewcommand{\arraystretch}{1.25}
	\begin{center}
		\begin{tabular}{m{0.1cm}lrrr}
			\hline  \hline
			                        & & Wind &  PV  & Load  \\ \hline
			\multirow{5}{*}{(a)} & 
			Embedding Net   	    & $4 \times 300$ & $4 \times 300$    & $4 \times 300$ \\ 
		&	Embedding size          & 40 & 40  & 40\\ 
		&	Integrand Net   	    & $3 \times 40$  & $3 \times 40$	 & $3 \times 40$ \\ 
		&	Weight decay            & 5.10$^{-4}$    & 5.10$^{-4}$       & 5.10$^{-4}$   \\
        &   Learning rate           & 10$^{-4}$      & 5.10$^{-4}$       & 10$^{-4}$  \\ \hline 
			\multirow{4}{*}{(b)} & 
			Latent dimension        & 20 & 40  & 5  \\
		&	E/D Net                 & $1 \times 200$ & $2 \times 200$   & $1 \times 500 $ \\ 
		&	Weight decay            & 10$^{-3.4}$    & 10$^{-3.5}$      & 10$^{-4}$  \\
		&	Learning rate           & 10$^{-3.4}$    & 10$^{-3.3}$      & 10$^{-3.9}$  \\ \hline 
			 \multirow{4}{*}{(c)}& 
			Latent dimension        & 64 & 64 & 256  \\
		&	G/D Net                 & $2 \times 256$ & $3 \times 256$   & $2 \times 1024 $ \\ 
		&	Weight decay            & 10$^{-4}$      & 10$^{-4}$        & 10$^{-4}$  \\
		&	Learning rate           & 2.10$^{-4}$    & 2.10$^{-4}$      & 2.10$^{-4}$  \\ \hline \hline
		\end{tabular}
		\caption{(a) NF, (b) VAE, and (c) GAN hyper-parameters. \\
        The hyper-parameters selection is performed on the validation set using the Python library Weights \& Biases \citep{wandb}. This library is an experiment tracking tool for machine learning, making it easier to track experiments. The GAN model was the most time-consuming during this process, followed by the VAE and NF. Indeed, the GAN is highly sensitive to hyper-parameter modifications making it challenging to identify a relevant set of values. In contrast, the NF achieved satisfactory results, both in terms of scenarios shapes and quality, by testing only a few sets of hyper-parameter values.
        }
		\label{tab:model_hp}
	\end{center}
\end{table} 
%
The NF implemented is a one-step monotonic normalizer using the UMNN-MAF\footnote{\url{https://github.com/AWehenkel/Normalizing-Flows}}. The embedding size $|h^i|$ is set to 40, and the embedding neural network is composed of $l$ layers of $n$ neurons ($l \times n$). The same integrand neural network $\tau^i(\cdot)$ $\forall i=1, \ldots, T$ is used and composed of 3 layers of $ |h^i|$ neurons ($3 \times 40$).
Both the encoder and decoder of the VAE are feed-forward neural networks ($l \times n$), ReLU activation functions for the hidden layers, and no activation function for the output layer.
Both the generator and discriminator of the GAN are feed-forward neural networks ($l \times n$). The activation functions of the hidden layers of the generator (discriminator) are ReLU (Leaky ReLU). The activation function of the discriminator output layer is ReLU, and there is no activation function for the generator output layer. The generator is trained once after the discriminator is trained five times to stabilize the training process, and the gradient penalty coefficient $\lambda$ in (\ref{eq:WGAN_GP}) is set to 10 as suggested by \citet{gulrajani2017improved}.

\begin{figure}[tb]
	\centering
    \includegraphics[width=\linewidth]{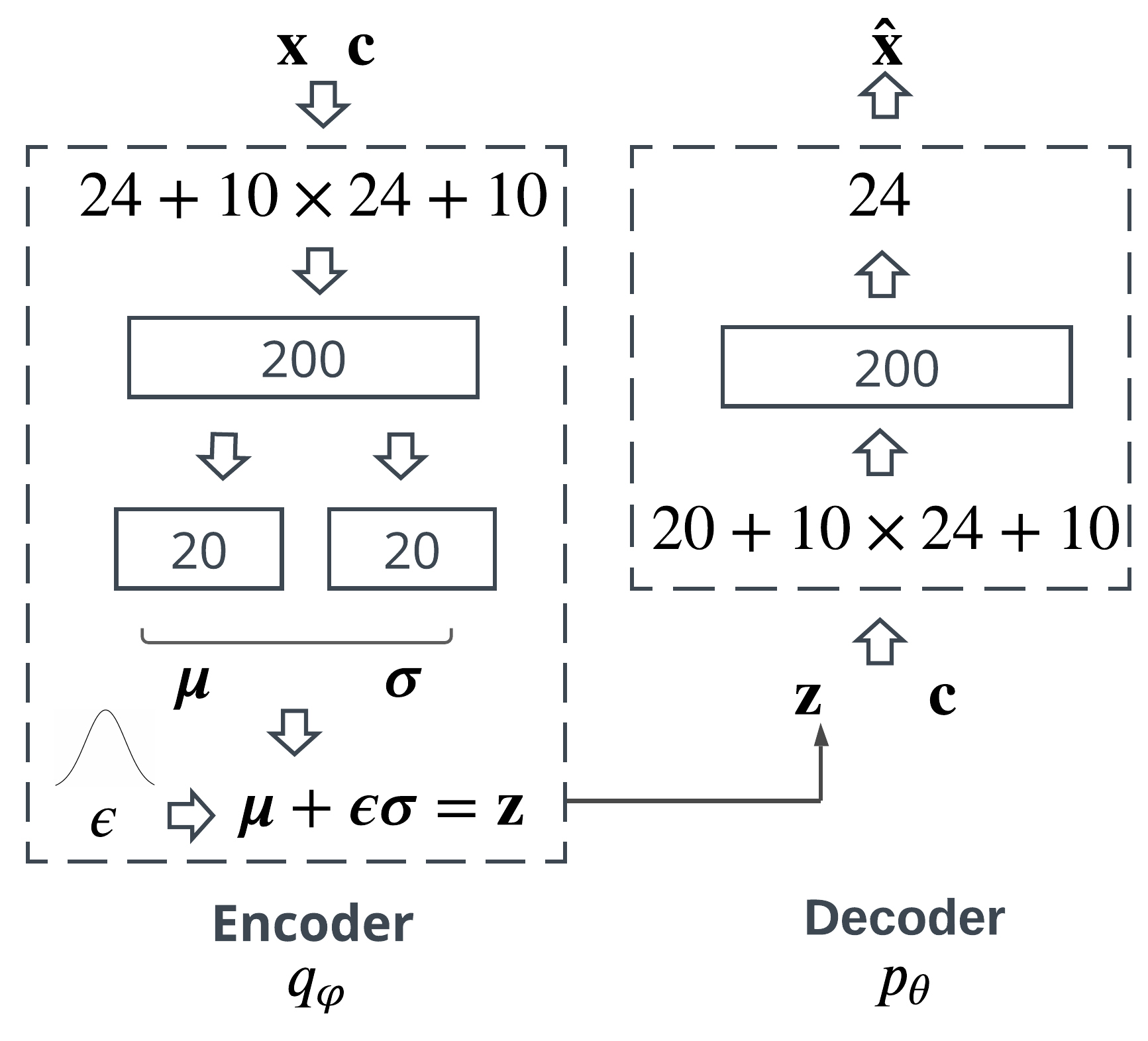}
	\caption{Variational autoencoder structure implemented for the wind dataset. \\
	Both the encoder and decoder are feed-forward neural networks composed of one hidden layer with 200 neurons. Increasing the number of layers did not improve the results for this dataset. The latent space dimension is 20.}
	\label{fig:vae-details}
\end{figure}

\begin{figure}[tb]
	\centering
    \includegraphics[width=\linewidth]{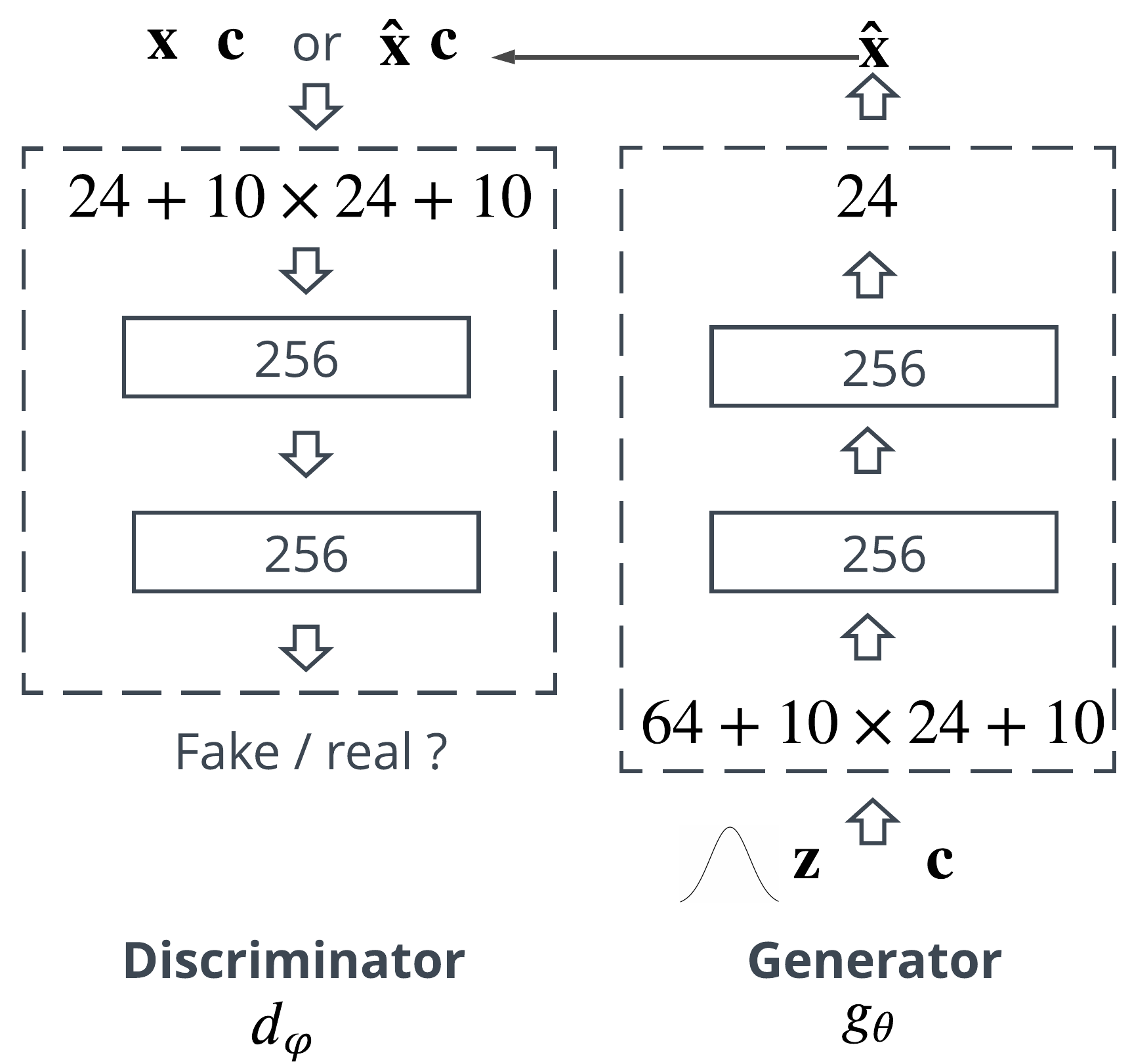}
	\caption{Generative adversarial network structure implemented for the wind dataset. \\
	Both the discriminator and generator are feed-forward neural networks composed of two hidden layers with 256 neurons. The latent space dimension is 64.}
	\label{fig:gan-details}
\end{figure}

\begin{figure}[tb]
	\centering
    \includegraphics[width=\linewidth]{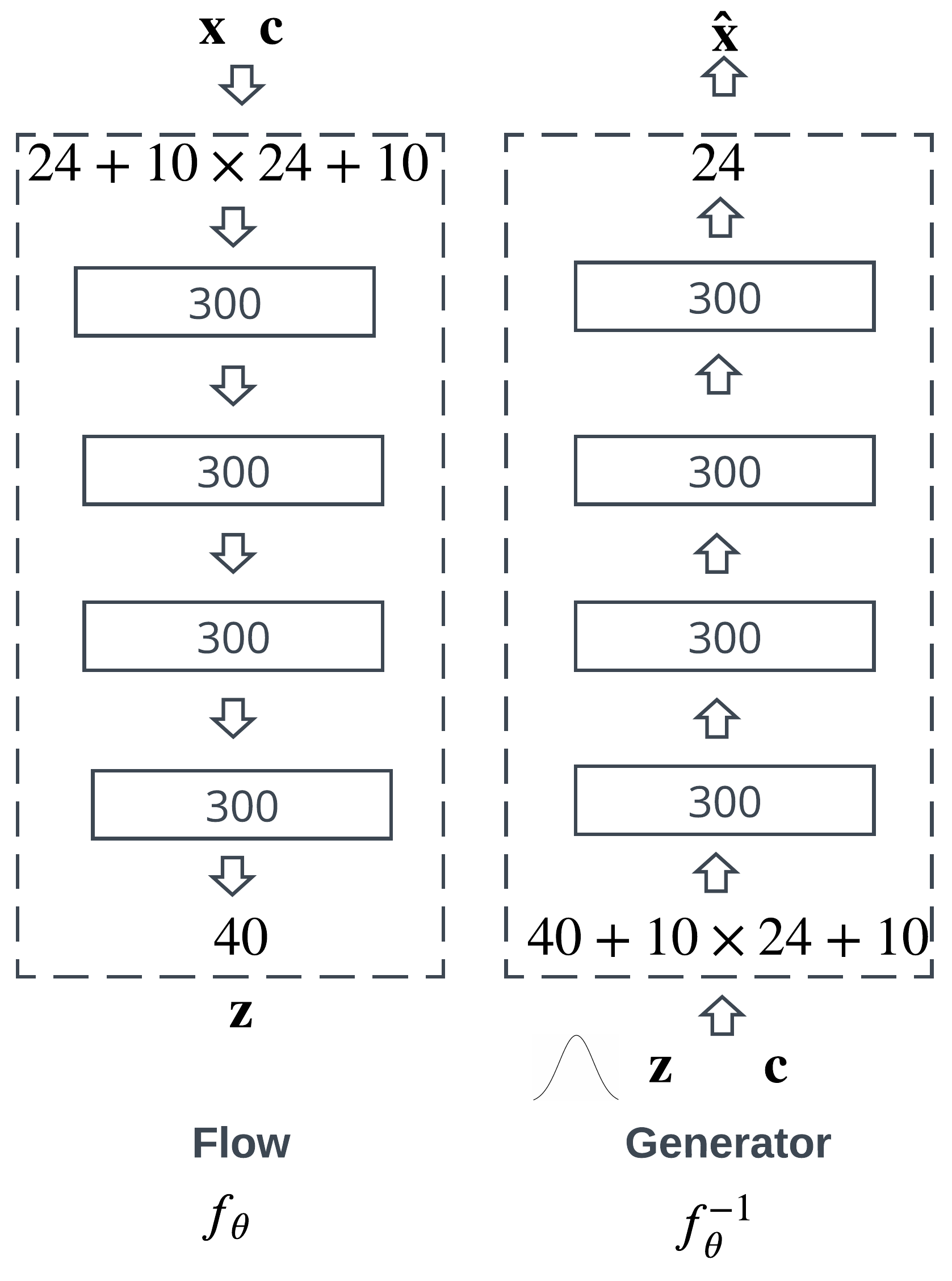}
	\caption{Normalizing flow structure implemented for the wind dataset. \\
A single-step monotonic normalizing flow is implemented with a feed-forward neural network composed of four hidden layers with 300 neurons. The latent space dimension is 40. Note: For clarity, the integrand network is not included but is a feed-forward neural network composed of three hidden layers with 40 neurons.
Increasing the number of steps of the normalizing flow did not improve the results. The monotonic transformation is complex enough to capture the stochasticity of the variable of interest. However, when considering affine autoregressive normalizing flows, the number of steps should be higher. Numerical experiments indicated that a five-step autoregressive flow was required to achieve similar results for this dataset. Note: the results are not reported in this study for the sake of clarity. }
\label{fig:nf-details}
\end{figure}

Figures \ref{fig:vae-details}, \ref{fig:gan-details}, and \ref{fig:nf-details} illustrate the VAE, GAN, and NF structures implemented for the wind dataset where the number of weather variables selected and the number of zones is 10, and 10, respectively. Recall, $\mathbf{c}:=$ weather forecasts,  $\mathbf{\hat{x}}:=$ scenarios $\mathbf{x}:=$ wind power observations, $\mathbf{z}:=$ latent space variable, $\mathbf{\epsilon}:=$ Normal variable (only for the VAE).

\section{Quality metrics}\label{annex:quality-assessment}

Recall that the PV generation, wind power, and load are assumed to be multivariate random variables of dimension $T$, $\mathbf{x} \in \mathbb{R}^T$, with $T$ the number of periods per day. 
Let $\cup_{d \in \text{TS}} \{  \mathbf{\hat{x}}_d^i \}_{i=1}^M$ be the set of $\#\text{TS} \times M$ scenarios generated with $M$ scenarios per day of the testing set, where $\mathbf{\hat{x}}^i_d \in \mathbb{R}^T \forall i,d$. $\hat{x}_{d,k}^i$ is the component $k$ of scenario $i$ on day $d$ of the testing set as specified by (\ref{eq:multi_output_scenario}).

\subsection{Continuous ranked probability score}

\citet{gneiting2007strictly} propose a formulation called the energy form of the CRPS since it is just the one-dimensional case of the energy score, defined in negative orientation as follows
\begin{align}\label{eq:CRPS_NRG}
\text{CRPS}(P, x_k) = & \mathbb{E}_P [|X-x_k|] - \frac{1}{2} \mathbb{E}_P [|X-X'|],
\end{align}
where and X and X' are independent random variables with distribution P and finite first moment, and $\mathbb{E}_P$ is the expectation according to the probabilistic distribution P. 
%
The CRPS is computed over the marginals of $\mathbf{\hat{x}}$ by using the estimator of (\ref{eq:CRPS_NRG}) provided by \citet{zamo2018estimation}. For a given day $d$ of the testing set, the CRPS per marginal $k=1, \cdots, T$ is
\begin{align}\label{eq:CRPS_eNRG}
\text{CRPS}_{d,k} & =  \frac{1}{M}\sum_{i = 1}^M|\hat{x}^i_{d,k}-x_{d,k}| - \frac{1}{2 M^2} \sum_{i,j = 1}^M |\hat{x}^i_{d,k} - \hat{x}^j_{d,k}|.
\end{align}
Then, it is averaged over the entire testing set 
\begin{align}\label{eq:CRPS-TEST}
\text{CRPS}_k = \frac{1}{\#TS} \sum_{d \in TS} \text{CRPS}_{d,k}.
\end{align}
In Table \ref{tab:quality_average_scores}, $\text{CRPS}_k$ is averaged over all time periods
\begin{align}\label{eq:CRPS-TEST-av}
\overline{\text{CRPS}} = \frac{1}{T} \sum_{k=1}^T \text{CRPS}_k.
\end{align}

\subsection{Energy score}

\citet{gneiting2007strictly} introduced a generalization of the continuous ranked probability score defined in negative orientation as follows
\begin{align}\label{eq:ES-def}
\text{ES}(P, \mathbf{x}) = & \mathbb{E}_P  \Vert X-\mathbf{x} \Vert - \frac{1}{2} \mathbb{E}_P \Vert X-X'\Vert,
\end{align}
where and X and X' are independent random variables with distribution P and finite first moment, $\mathbb{E}_P$ is the expectation according to the probabilistic distribution P, and $ \Vert \cdot \Vert $ the Euclidean norm. 
%
For a given day $d$ of the testing set, the ES is computed following \citet{gneiting2008assessing}
\begin{align}\label{eq:ES-estimator}
\text{ES}_d & = \frac{1}{M}\sum_{i = 1}^M \Vert \mathbf{\hat{x}}_d^i-\mathbf{x}_d\Vert - \frac{1}{2 M^2} \sum_{i,j = 1}^M \Vert\mathbf{\hat{x}}_d^i - \mathbf{\hat{x}}_d^j \Vert.
\end{align}
Then, it is averaged over the testing set 
\begin{align}\label{eq:ES-TEST}
\text{ES} & = \frac{1}{\#TS} \sum_{d \in TS} \text{ES}_d.
\end{align}
Note: when we consider the marginals of $\mathbf{x}$, it is easy to recognize that (\ref{eq:ES-estimator}) is the CRPS.

\subsection{Variogram score}

For a given day $d$ of the testing set and a $T$-variate observation $\mathbf{x}_d \in \mathbb{R}^T$, the Variogram score metric of order $\gamma$ is formally defined as
\begin{align}\label{eq:VS-def}
\text{VS}_d & = \sum_{k,k'}^T w_{kk'} \bigg ( |x_{d,k} -x_{d,k'} |^\gamma - \mathbb{E}_P |\hat{x}_{d,k} -\hat{x}_{d,k'} |^\gamma \bigg)^2,
\end{align}
where $\hat{x}_{d,k}$ and $\hat{x}_{d,k'}$ are the $k$-th and $k'$-th components of the random vector $\mathbf{\hat{x}}_d$ distributed according to P for which the $\gamma$-th absolute moment exists, and $w_{kk'}$ are non-negative weights. Given a set of $M$ scenarios $\{  \mathbf{\hat{x}}_d^i \}_{i=1}^M$ for this given day $d$, the forecast variogram $\mathbb{E}_P |\hat{x}_{d,k} -\hat{x}_{d,k'} |^\gamma$ can be approximated $\forall k,k' =1, \cdots, T$ by 
\begin{align}\label{eq:VS-approximation}
\mathbb{E}_P |\hat{x}_{d,k} -\hat{x}_{d,k'} |^\gamma & \approx \frac{1}{M} \sum_{i=1}^M |\hat{x}_{d,k}^i -\hat{x}_{d,k'}^i |^\gamma.
\end{align}
Then, it is averaged over the testing set 
\begin{align}\label{eq:VS-TEST}
\text{VS} & = \frac{1}{\#TS} \sum_{d \in TS} \text{VS}_d.
\end{align}
In this study, we evaluate the Variogram score with equal weights across all hours of the day $w_{kk'}=1$ and using a $\gamma$ of 0.5, which for most cases provides a good discriminating ability as reported in \citet{scheuerer2015variogram}. 

\subsection{Quantile score}

For a given day $d$ of the testing set, a set of 99 quantiles (1, 2, ..., 99-th quantile) $\{ \mathbf{\hat{x}}^q_d\}_{q=1}^{99}$ are computed from the set of $M$ scenarios $\{ \mathbf{\hat{x}}_d^i \}_{i=1}^M$, with $q$ the quantile index ($q=0.01, \ldots, 0.99$).
For a given day $d$ of the testing set, the quantile score, per marginal, is defined by
\begin{equation}\label{eq:QS-marginal}
\rho_q(\hat{x}^q_{d,k}, x_{d,k}) =  \begin{cases} (1-q) \times (\hat{x}^q_{d,k} - x_{d,k})  &  x_{d,k} < \hat{x}^q_{d,k} \\
 q \times (x_{d,k} - \hat{x}^q_{d,k}) &    x_{d,k}  \geq \hat{x}^q_{d,k} \end{cases}.
\end{equation}
Then, it is averaged over all time periods and the testing set
\begin{equation}\label{eq:QS-all-marginals}
\text{QS}_q = \frac{1}{\#TS} \sum_{d \in TS} \frac{1}{T}\sum_{k=1}^T \rho_q(\hat{x}^q_{d,k}, x_{d,k}) .
\end{equation}
In Table \ref{tab:quality_average_scores}, $\text{QS}_q$ is averaged over all quantiles
\begin{align}\label{eq:QS-TEST-av}
\overline{\text{QS}} = \frac{1}{99} \sum_{q=1}^{99} \text{QS}_q.
\end{align}
 
\subsection{Classifier-based metric}

Modern binary classifiers can be easily turned into robust two-sample tests where the goal is to assess whether two samples are drawn from the same distribution \citep{lehmann2006testing}. In other words, it aims at assessing whether a generated scenario can be distinguished from an observation. 
To this end, the generator is evaluated on a held-out testing set that is split into a testing-train and testing-test subsets. The testing-train set is used to train a classifier, distinguishing generated scenarios from the actual distribution. Then, the final score is computed as the performance of this classifier on the testing-test set.

In principle, any binary classifier can be adopted for computing classifier two-sample tests (C2ST). A variation of this evaluation methodology is proposed by \citet{xu2018empirical} and is known as the 1-Nearest Neighbor (NN) classifier. The advantage of using 1-NN over other classifiers is that it requires no special training and little hyper-parameter tuning. This process is conducted as follows. Given two sets of observations $S_r$ and generated $S_g$ samples with the same size, \textit{i.e.}, $|S_r| = |S_g|$, it is possible to compute the leave-one-out (LOO) accuracy of a 1-NN classifier trained on $S_r$ and $S_g$ with positive labels for $S_r$ and negative labels for $S_g$. The LOO accuracy can vary from 0 \% to 100 \%. The 1-NN classifier should yield a 50 \% LOO accuracy when $|S_r| = |S_g|$ is large. It is achieved when the two distributions match. Indeed, the level 50 \% happens when a label is randomly assigned to a generated scenario. It means the classifier is not capable of discriminating generated scenarios from observations. If the generative model over-fits $S_g$ to $S_r$, \textit{i.e.}, $S_g = S_r$, and the accuracy would be 0 \%. On the contrary, if it generates widely different samples than observations, the performance should be 100 \%. Therefore, the closer the LOO accuracy is to 1, the higher the degree of under-fitting of the model. The closer the LOO accuracy is to 0, the higher the degree of over-fitting of the model. The C2ST approach using LOO with 1-NN is adopted by \citet{qi2020optimal} to assess the PV and wind power scenarios of a $\beta$ VAE.

However, this approach has several limitations. First, it uses the testing set to train the classifier during the LOO. Second, the 1-NN is very sensitive to outliers as it simply chose the closest neighbor based on distance criteria. This behavior is amplified when combined with the LOO, where the testing-test set is composed of only one sample. Third, the euclidian distance cannot deal with a context such as weather forecasts. Therefore, we cannot use a conditional version of the 1-NN using weather forecasts to classify weather-based renewable generation and the observations. Fourth, C2ST with LOO cannot provide ROC curve but only accuracy scores. 
An essential point about ROC graphs is that they measure the ability of a classifier to produce good relative instance scores. In our case, we are interested in discriminating the generated scenarios from the observations, and the ROC provides more information than the accuracy metric to achieve this goal. A standard method to reduce ROC performance to a single scalar value representing expected performance is to calculate the area under the ROC curve, abbreviated AUC. The AUC has an essential statistical property: it is equivalent to the probability that the classifier will rank a randomly chosen positive instance higher than a randomly chosen negative instance \citep{fawcett2004roc}.

To deal with these issues, we decided to modify this classifier-based evaluation by conducting the C2ST as follows: (1) the scenarios generated on the learning set are used to train the classifier using the C2ST. Therefore, the classifier uses the entire testing set and can compute ROC; (2) the classifier is an Extra-Trees classifier that can deal with context such as weather forecasts.

\begin{figure}[tb]
\centering
\includegraphics[width=0.5\linewidth]{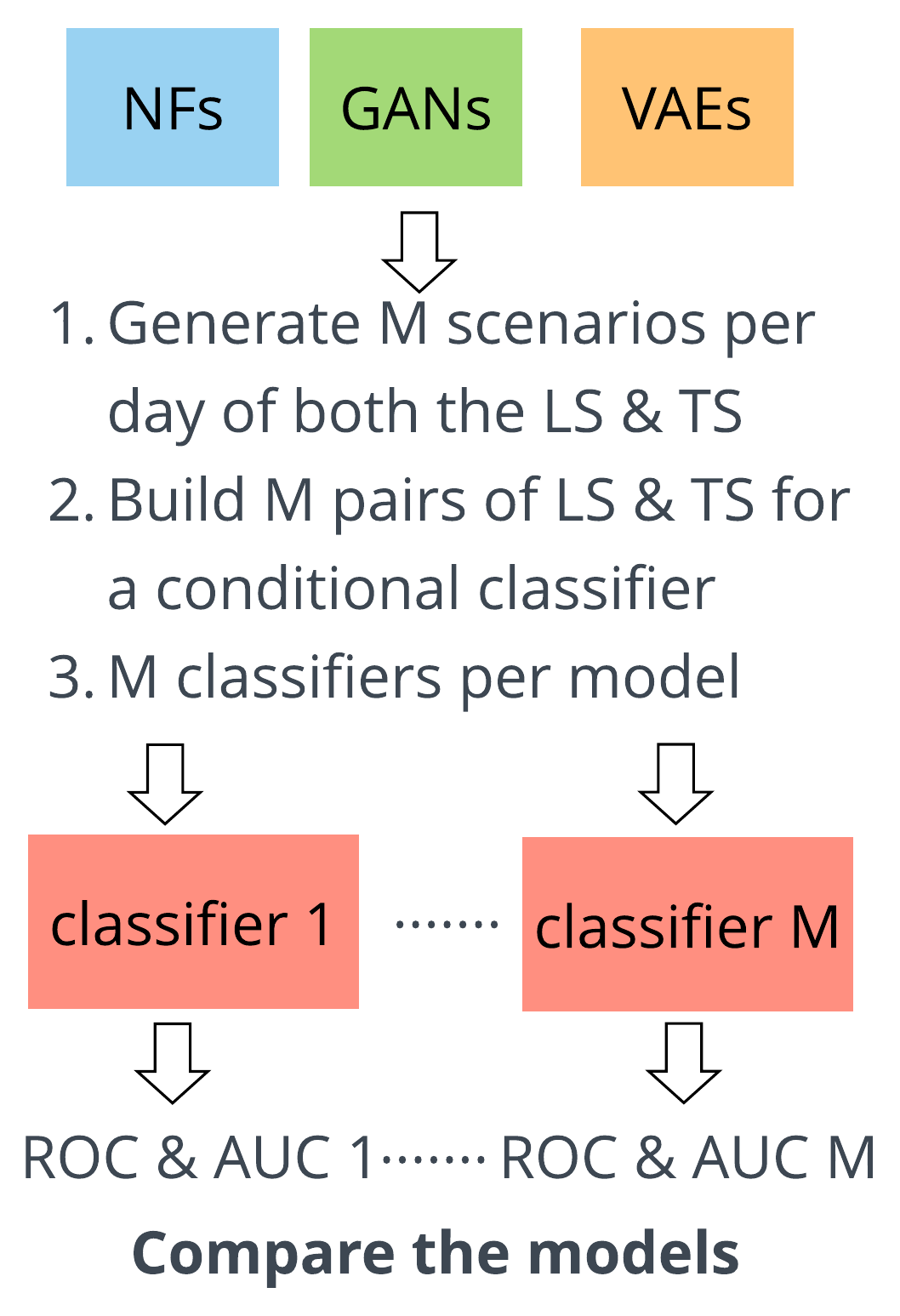}
\caption{Classifier-based metric methodology. \\ 
Each generative model generates $M$ scenarios per day of the learning and testing sets. They are used to build $M$ pairs of learning and testing sets for a conditional classifier by including an equal proportion of observations and weather forecasts. $M$ conditional classifiers, per model, are trained and make predictions. The $M$ ROC and AUC are computed per model, and the results are compared.
}
\label{fig:classifier-based}
\end{figure}

More formally, for a given generative model $g$ the following steps are conducted:
\begin{enumerate}
    \item Initialization step: the generative model $g$ has been trained on the LS and has generated $M$ weather-based scenarios per day of both the LS
     and TS: $\{  \mathbf{\hat{x}}_\text{LS}^i \}_{i=1}^M :=\cup_{d \in \text{LS}} \{  \mathbf{\hat{x}}_d^i \}_{i=1}^M$ and $\{  \mathbf{\hat{x}}_\text{TS}^i \}_{i=1}^M :=\cup_{d \in \text{TS}} \{  \mathbf{\hat{x}}_d^i \}_{i=1}^M$. For the sake of clarity the index $g$ is omitted, but both of these sets are dependent on model $g$.
    \item $M$ pairs of learning and testing sets are built with an equal proportion of generated scenarios and observations: $\mathcal{D}_\text{LS}^i := \bigg\{ \{\mathbf{\hat{x}}_\text{LS}^i, 0\} \cup \{    \{\mathbf{x}_\text{LS}^i, 1\}  \bigg\} $ and $\mathcal{D}_\text{TS}^i = \bigg\{ \{\mathbf{\hat{x}}_\text{TS}^i, 0\} \cup \{    \{\mathbf{x}_\text{TS}^i, 1\}  \bigg\} $. Note: $|\mathcal{D}_\text{LS}^i| = 2|\text{LS}|$ and $|\mathcal{D}_\text{TS}^i| = 2|\text{TS}|$.
    \item For each pair of learning and testing sets $\{\mathcal{D}_\text{LS}^i, \mathcal{D}_\text{TS}^i\}_{i=1}^M$ a classifier $d_g^i$ is trained and makes predictions.
    \item The $\text{ROC}_g^i$ curves and corresponding $\text{AUC}_g^i$ are computed for $i=1, \cdots, M$.
\end{enumerate}
This classifier-based methodology is conducted for all models $g$, and the results are compared. Figure \ref{fig:classifier-based} depicts the overall approach.
The classifiers $d_g^i$ are all Extra-Trees classifier made of $1000$ unconstrained trees with the hyper-parameters "$\text{max\_depth}$" set to "None", and ``$\text{n\_estimators}$" to 1 000.


\subsection{Diebold-Mariano test}

For a given day $d$ of the testing set, let $\epsilon_d \in \mathbb{R}$ be the error computed by an arbitrary forecast loss function of the observation and scenarios. The test consists of computing the difference between the errors of the pair of models $g$ and $h$ over the testing set
\begin{equation}
\label{eq:univariate-loss}
\Delta(g,h)_d  = \epsilon^g_d - \epsilon^h_d, \quad \forall d \in \text{TS},
\end{equation}
and to perform an asymptotic $z$-test for the null hypothesis that the expected forecast error is equal and the mean of differential loss series is zero $ \mathbb{E}[\Delta(g,h)_d] = 0$. It means there is no statistically significant difference in the accuracy of the two competing forecasts.
The statistic of the test is deduced from the asymptotically standard normal distribution as follows
\begin{equation}
\label{eq:DM-value}
\text{DM}(g,h) = \sqrt{\# \text{TS}}\frac{\hat{\mu}}{\hat{\sigma}},
\end{equation}
with $\# \text{TS}$ the number of days of the testing set, $\hat{\mu}$ and $\hat{\sigma}$ the sample mean and the standard deviation of $\Delta(g,h)$.
Under the assumption of covariance stationarity of the loss differential series $\Delta(g,h)_d$, the DM statistic is asymptotically standard normal.
The lower the $p$-value, \textit{i.e.}, the closer it is to zero, the more the observed data is inconsistent with the null hypothesis: $ \mathbb{E}[\Delta(g,h)_d] < 0$ the forecasts of the model $h$ are more accurate than those of model $g$. If the $p$-value is less than the commonly accepted level of 5 \%, the null hypothesis is typically rejected. It means that the forecasts of model $g$ are significantly more accurate than those of model $h$.

When considering the ES or VS scores, there is a value per day of the testing set $\text{ES}_d$ or $\text{VS}_d$. In this case, $\epsilon_d = \text{ES}_d$ or $\epsilon_d = \text{VS}_d$.
However, when considering the CRPS or QS, there is a value per marginal and per day of the testing set $\text{CRPS}_{d,k}$ or $\text{QS}_{d,k}$. 
A solution consists of computing 24 independent tests, one for each hour of the day. Then, to compare the models based on the number of hours for which the predictions of one model are significantly better than those of another. Another way consists of a multivariate variant of the DM-test with the test performed jointly for all hours using the multivariate loss differential series.
In this case, for a given day $d$, $\mathbf{\epsilon}^g_d = [\epsilon^g_{d,1}, \ldots,  \epsilon^g_{d,24}]^\intercal$, $\mathbf{\epsilon}^h_d = [\epsilon^h_{d,1}, \ldots,  \epsilon^h_{d,24}]^\intercal$ are the vectors of errors for a given metric of models $g$ and $h$, respectively. Then the multivariate loss differential series
\begin{equation}
\label{eq:multivariate-loss}
\Delta(g,h)_d  = \Vert \mathbf{\epsilon}^g_d\Vert_1 - \Vert \mathbf{\epsilon}^h_d\Vert_1,
\end{equation}
defines the differences of errors using the $\Vert \cdot \Vert_1$ norm. Then, the $p$-value of two-sided DM tests is computed for each model pair as described above.
The univariate version of the test has the advantage of providing a deeper analysis as it indicates which forecast is significantly better for which hour of the day.
The multivariate version enables a better representation of the results as it summarizes the comparison in a single $p$-value, which can be conveniently visualized using heat maps arranged as chessboards. In this study, we decided to adopt the multivariate DM-test for the CRPS and QS.

\section{Value assessment}\label{annex:value-assessment}

\subsection{Notation}

\subsubsection*{Sets and indexes}
\begin{supertabular}{l p{0.7\columnwidth}}
	Name & Description \\
	\hline
	$t$ & Time period index. \\
	$\omega$ & Scenario index. \\
	$T$ & Number of time periods per day. \\
	$\#\Omega$ & Number of scenarios. \\
	$\mathcal{T}$ & Set of time periods, $\mathcal{T}= \{1,2, \ldots, T\}$. \\
	$\Omega$ & Set of scenarios, $\Omega= \{1,2, \ldots, \#\Omega\}$. \\
\end{supertabular}

\subsubsection*{Parameters}
\begin{supertabular}{l p{0.65\columnwidth} }
	Name & Description  \\
	\hline
	$e_t^{min}$, $e_t^{max}$ & Minimum/maximum day-ahead bid [MWh]. \\
	$y_t^{min}$, $y_t^{max}$ & Minimum/maximum retailer net position [MWh]. \\
	$y^\text{dis}_\text{max}$, $y^\text{cha}_\text{max}$ & BESS maximum (dis)charging power [MW].  \\
	$\eta^\text{dis}$, $\eta^\text{cha}$ & BESS (dis)charging efficiency [-]. \\
	$s^\text{min}$, $s^\text{max}$ & BESS minimum/maximum capacity [MWh]. \\
	$s^\text{ini}$, $s^\text{end}$ & BESS initial/final state of charge [MWh].  \\
	$\pi_t$ & Day-ahead price [\euro/MWh].   \\
	$\bar{q}_t$, $\bar{\lambda}_t$ & Negative/positive imbalance price [\euro/MWh].   \\
	$\Delta t$ & Duration of a time period [hour].   \\
	
\end{supertabular}

\subsubsection*{Variables}
\noindent For the sake of clarity the subscript $\omega$ is omitted.
\begin{supertabular}{l l p{0.6\columnwidth}}
	Name & Range & Description \\
	\hline
	$e_t$ & $ [e_t^{min},e_t^{max}]$ & Day-ahead bid [MWh]. \\
	$y_t$ & $ [y_t^{min},y_t^{max}]$ & Retailer net position [MWh]. \\
	$\PVgeneration_t $ & $ [0, 1]$ & PV generation [MW].  \\
	$\Wgeneration_t $ & $ [0, 1]$ & Wind generation [MW].  \\
	$\Charge_t $ & $ [0, y^\text{cha}_\text{max}]$ & Charging power [MW].    \\
	$\Discharge_t $ & $ [0, y^\text{dis}_\text{max}]$ & Discharging power [MW].   \\
	$\Soc_t$ & $ [s^\text{min}, s^\text{max}]$ & BESS state of charge [MWh]. \\
	$d^-_t$, $d^+_t$  & $ \mathbb{R}_+$ & Short/long deviation [MWh].  \\
	$y_t^b$ & $ \{0, 1\}$ & BESS binary variable [-]. \\
	
\end{supertabular}

\subsection{Problem formulation}

The mixed-integer linear programming (MILP) optimization problem to solve is 
\begin{subequations}
\label{eq:S_obj}	
\begin{align}
	\max_{e_t \in \mathcal{X}, y_{t, \omega} \in \mathcal{Y}(e_t)}  & \sum_{\omega \in \Omega} \alpha_\omega  \sum_{t\in \mathcal{T}} \bigg[  \pi_t e_t -  \bar{q}_t  d^-_{t, \omega}  - \bar{\lambda}_t d^+_{t, \omega} \bigg], \\
	\mathcal{X} =  & \bigg \{ e_t : e_t \in [e_t^{min}, e_t^{max}] \bigg \}, \\
	\mathcal{Y}(e_t) = & \bigg \{ y_{t, \omega} : (\ref{eq:short_dev})- (\ref{eq:BESS_dyn_last_period})  \bigg \}.
\end{align}
\end{subequations}
The optimization variables are $e_t$, day-ahead bid of the net position, $\forallw$, $y_{t,\omega}$, retailer net position in scenario $\omega$, $d^-_{t, \omega}$, short deviation, $d^+_{t, \omega}$, long deviation, $\PVgeneration_{t,\omega}$, PV generation, $\Wgeneration_{t,\omega}$, wind generation, $\Charge_{t,\omega}$, battery energy storage system (BESS) charging power, $\Discharge_{t,\omega}$, BESS discharging power, $\Soc_{t,\omega}$, BESS state of charge, and $y^b_{t,\omega}$ a binary variable to prevent from charging and discharging simultaneously. 
The imbalance penalty is modeled by the constraints (\ref{eq:short_dev})-(\ref{eq:long_dev}) $\forallw$, that define the short and long deviations variables $d^-_{t, \omega}, d^+_{t, \omega}  \in \mathbb{R}_+$. The energy balance is provided by (\ref{eq:energy_balance})  $\forallw$. The set of constraints that bound $\PVgeneration_{t,\omega}$ and $\Wgeneration_{t,\omega}$  variables are (\ref{eq:PV_cst})-(\ref{eq:W_cst}) $\forallw$ where $\PVforecast_{t,\omega}$ and $\Wforecast_{t,\omega}$ are PV and wind generation scenarios. The load is assumed to be non-flexible and is a parameter (\ref{eq:L_cst}) $\forallw$ where $\Loadforecast_{t,\omega}$ are load scenarios. The BESS constraints are provided by (\ref{eq:BESS_charge})-(\ref{eq:BESS_max_soc}), and the BESS dynamics by  (\ref{eq:BESS_dyn_first_period})-(\ref{eq:BESS_dyn_last_period}) $\forallw$.
\begin{subequations}
	\label{eq:dispatch_constraint}	
	\begin{align}
	- d^-_{t, \omega} & \leq     - ( e_t - y_{t,\omega} )   , \forallt \label{eq:short_dev}	\\
	- d^+_{t, \omega} & \leq    -  ( y_{t,\omega} -e_t )   , \forallt  \label{eq:long_dev}	\\
	\frac{y_{t,\omega}}{\Delta t}&  =  \PVgeneration_{t,\omega} +  \Wgeneration_{t,\omega} -  \Load_{t,\omega} \notag \\
	& + \Discharge_{t,\omega} - \Charge_{t,\omega}, \forallt \label{eq:energy_balance}	\\
	\PVgeneration_{t,\omega} & \leq \PVforecast_{t,\omega}, \forallt \label{eq:PV_cst}		\\
	\Wgeneration_{t,\omega} & \leq \Wforecast_{t,\omega}, \forallt \label{eq:W_cst}		\\
	\Load_{t,\omega} & = \Loadforecast_{t,\omega}, \forallt \label{eq:L_cst} \\
	\Charge_{t,\omega} & \leq y^b_{t,\omega} y^\text{cha}_\text{max}, \forallt \label{eq:BESS_charge}		\\
	\Discharge_{t,\omega} & \leq (1-y^b_{t,\omega}) y^\text{dis}_\text{max}, \forallt \label{eq:BESS_discharge}		\\
	-\Soc_{t,\omega} & \leq -s^\text{min}, \forallt \label{eq:BESS_min_soc}		\\
	\Soc_{t,\omega} & \leq s^\text{max}, \forallt \label{eq:BESS_max_soc}		\\
	\frac{\Soc_{1,\omega} - s^\text{ini}}{\Delta t} & =  \eta^\text{cha} \Charge_{1,\omega} - \frac{\Discharge_{1,\omega}}{\eta^\text{dis}},  \label{eq:BESS_dyn_first_period}		\\
	\frac{\Soc_{t,\omega} - \Soc_{t-1,\omega}}{\Delta t}& =  \eta^\text{cha} \Charge_{t,\omega} - \frac{\Discharge_{t,\omega}}{\eta^\text{dis}}, \forallt \setminus \{1\} \label{eq:BESS_dyn_all_period}		\\
	\Soc_{T,\omega}& = s^\text{end} = s^\text{ini} \label{eq:BESS_dyn_last_period} .
	\end{align}
\end{subequations}
Notice that if $\bar{\lambda}_t <0$, the surplus quantity is remunerated with a non-negative price. 
In practice, such a scenario could be avoided provided that the energy retailer has curtailment capabilities, and $(\bar{q}_t , \bar{\lambda}_t)$ are strictly positive in our case study.
%
The deterministic formulation with perfect forecasts, the oracle (O), is a specific case of the stochastic formulation by considering only one scenario where $\PVgeneration_{t,\omega}$, $\Wgeneration_{t,\omega}$, and $\Load_{t,\omega}$ become the actual values of PV, wind, and load $\forallt$. The optimization variables are $e_t$, $y_t$, $d_t^-$, $d_t^+$, $\PVgeneration_t$, and $\Wgeneration_t$, $\Charge_t$, $\Discharge_t$, $\Soc_t$, and $y^b_t$.

\subsection{Dispatching}

Once the bids $e_t$ have been computed by the planner, the dispatching consists of computing the second stage variables given observations of the PV, wind power, and load. The dispatch formulation is a specific case of the stochastic formulation with $e_t$ as parameter and by considering only one scenario where  $\PVgeneration_{t,\omega}$, $\Wgeneration_{t,\omega}$, and $\Load_{t,\omega}$ become the actual values of PV, wind, and load $\forallt$. The optimization variables are $y_t$, $d_t^-$, $d_t^+$, $\PVgeneration_t$, and $\Wgeneration_t$, $\Charge_t$, $\Discharge_t$, $\Soc_t$, and $y^b_t$.

\clearpage
\section{Quality results}\label{annex:assessment_results}

%
\begin{figure}[htb]
	\centering
	\begin{subfigure}{.25\textwidth}
		\centering
		\includegraphics[width=\linewidth]{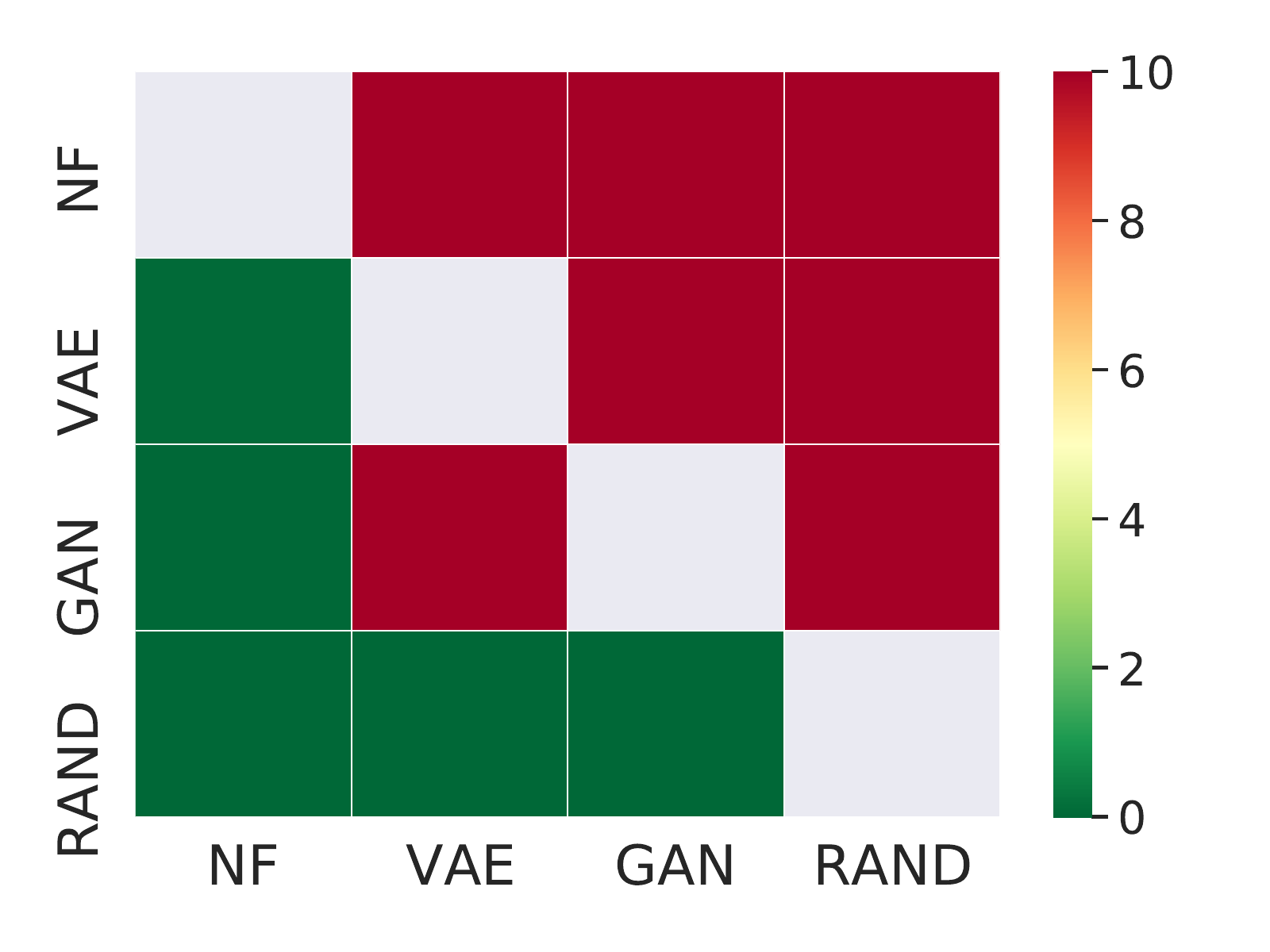}
		\caption{PV CRPS DM.}
	\end{subfigure}%
	\begin{subfigure}{.25\textwidth}
		\centering
		\includegraphics[width=\linewidth]{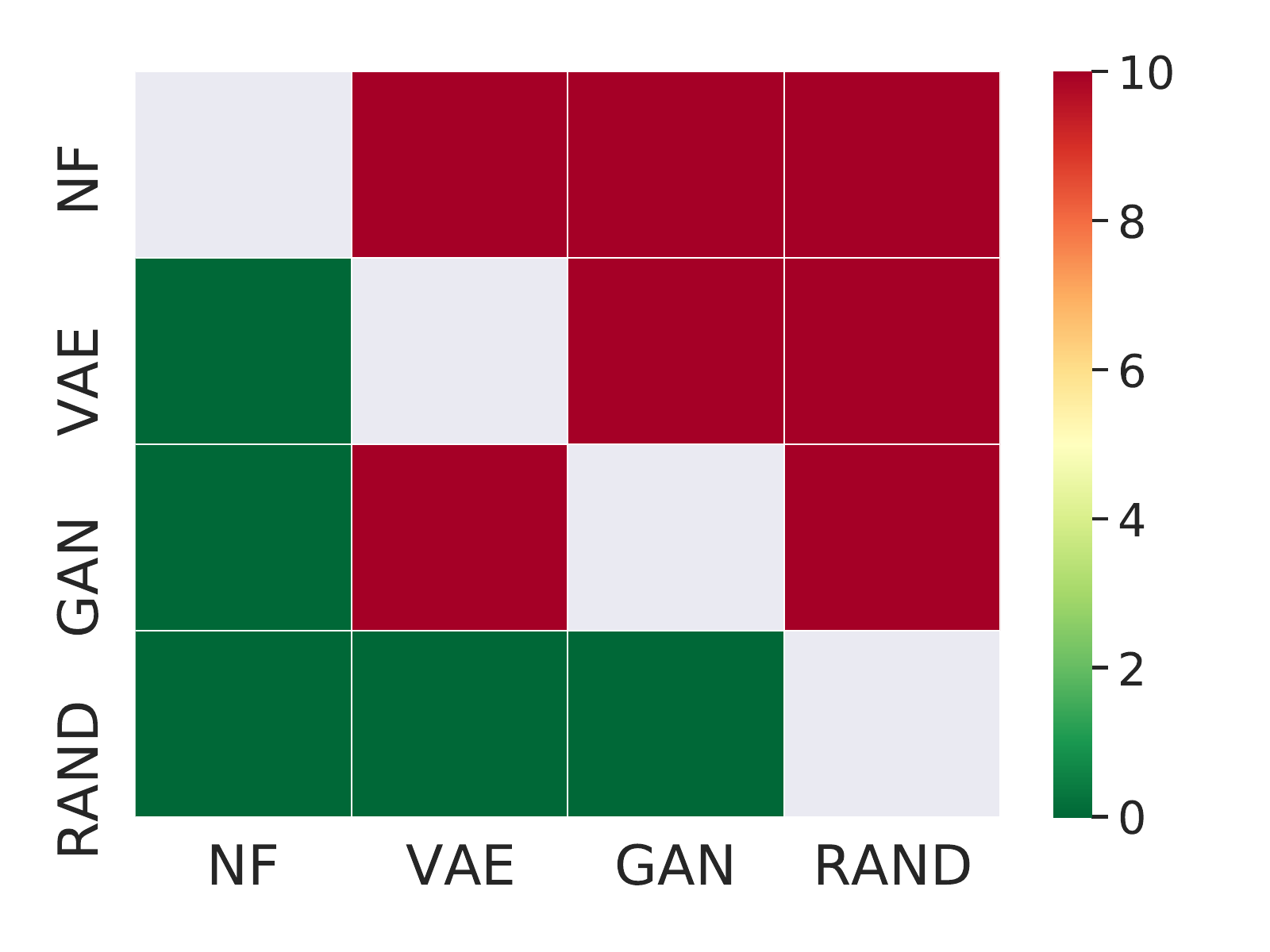}
		\caption{Load CRPS DM.}
	\end{subfigure}
	\begin{subfigure}{.25\textwidth}
		\centering
		\includegraphics[width=\linewidth]{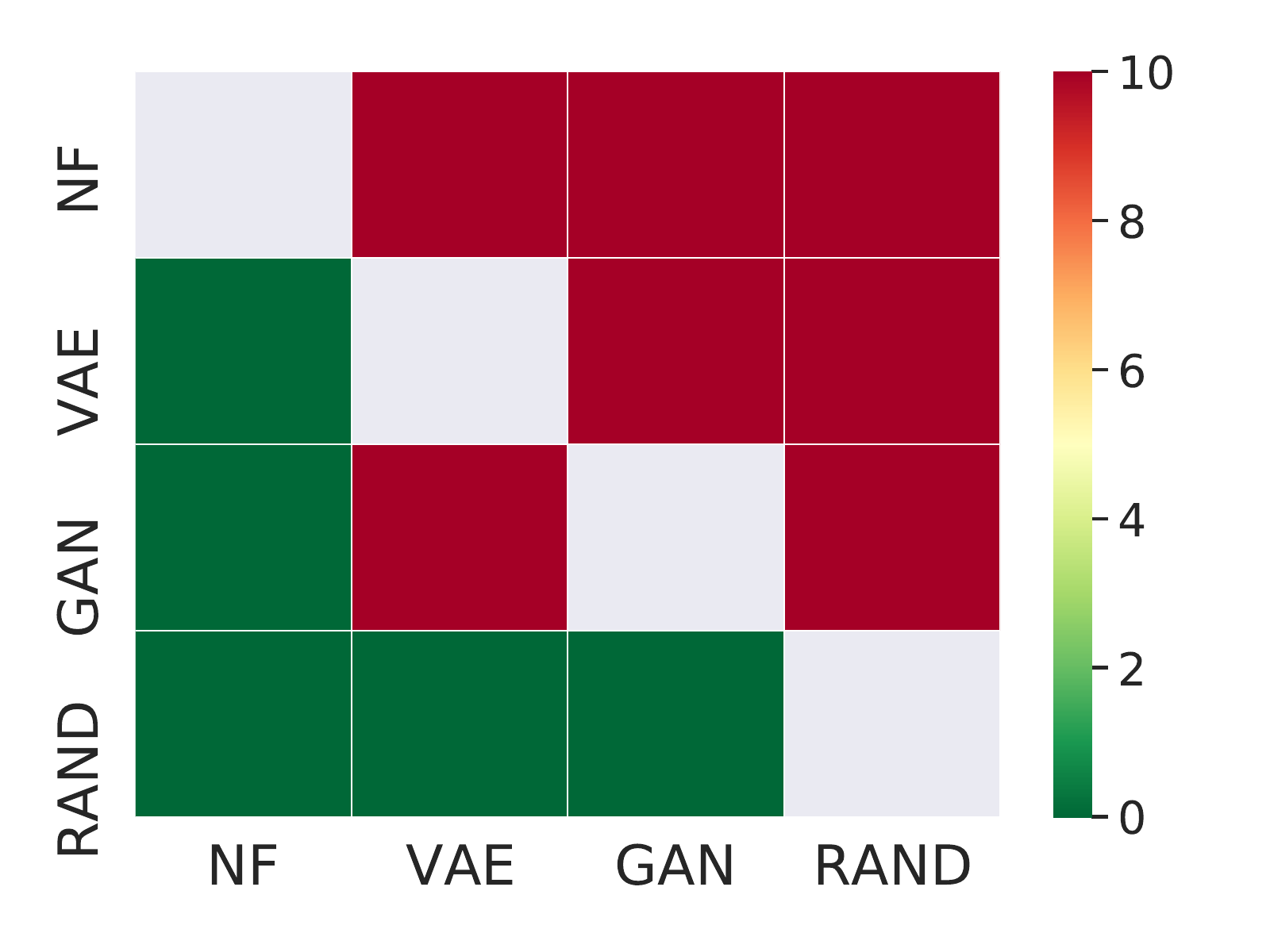}
		\caption{PV QS DM.}
	\end{subfigure}%
	\begin{subfigure}{.25\textwidth}
		\centering
		\includegraphics[width=\linewidth]{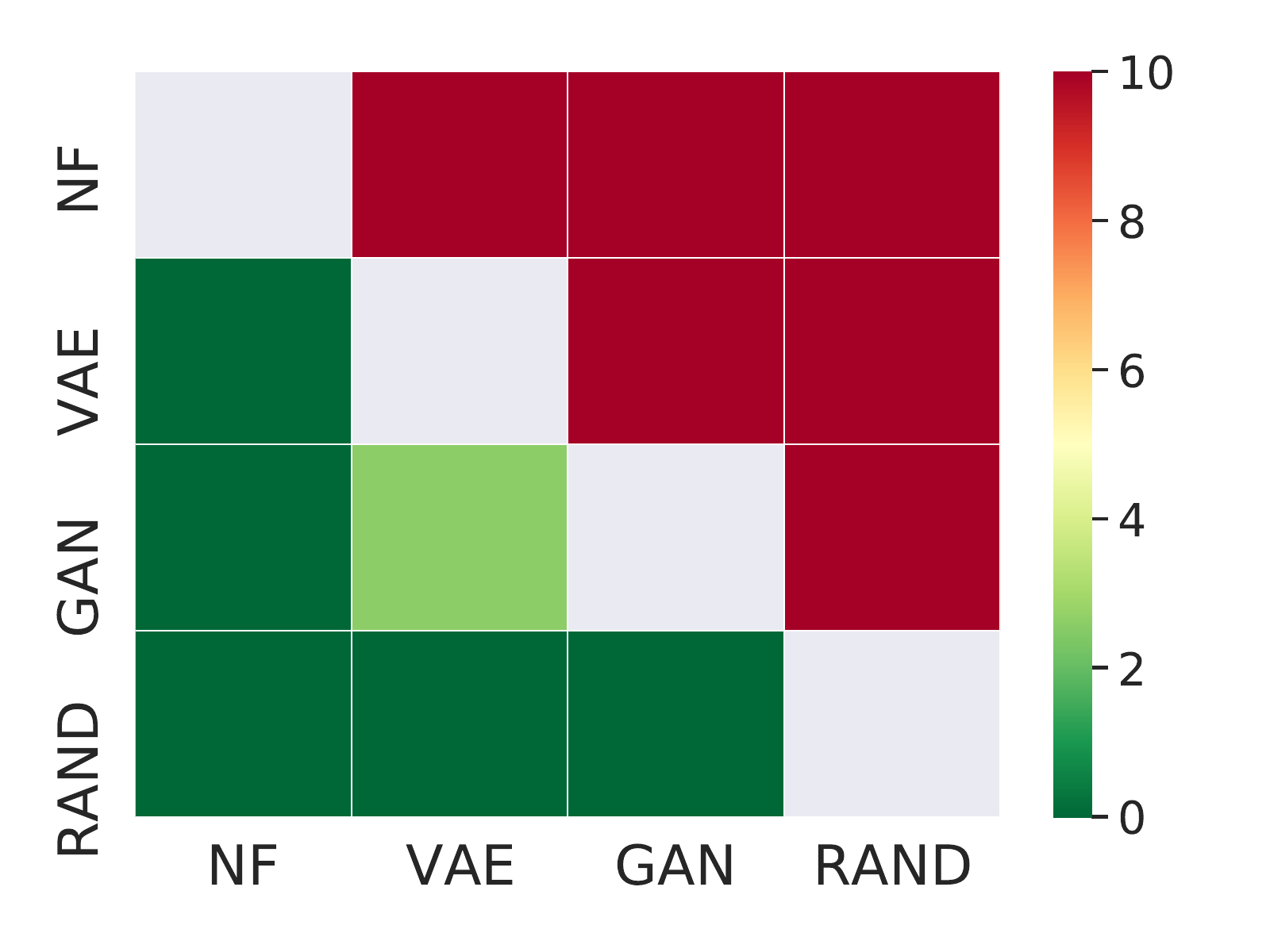}
		\caption{Load QS DM.}
	\end{subfigure}
	\begin{subfigure}{.25\textwidth}
		\centering
		\includegraphics[width=\linewidth]{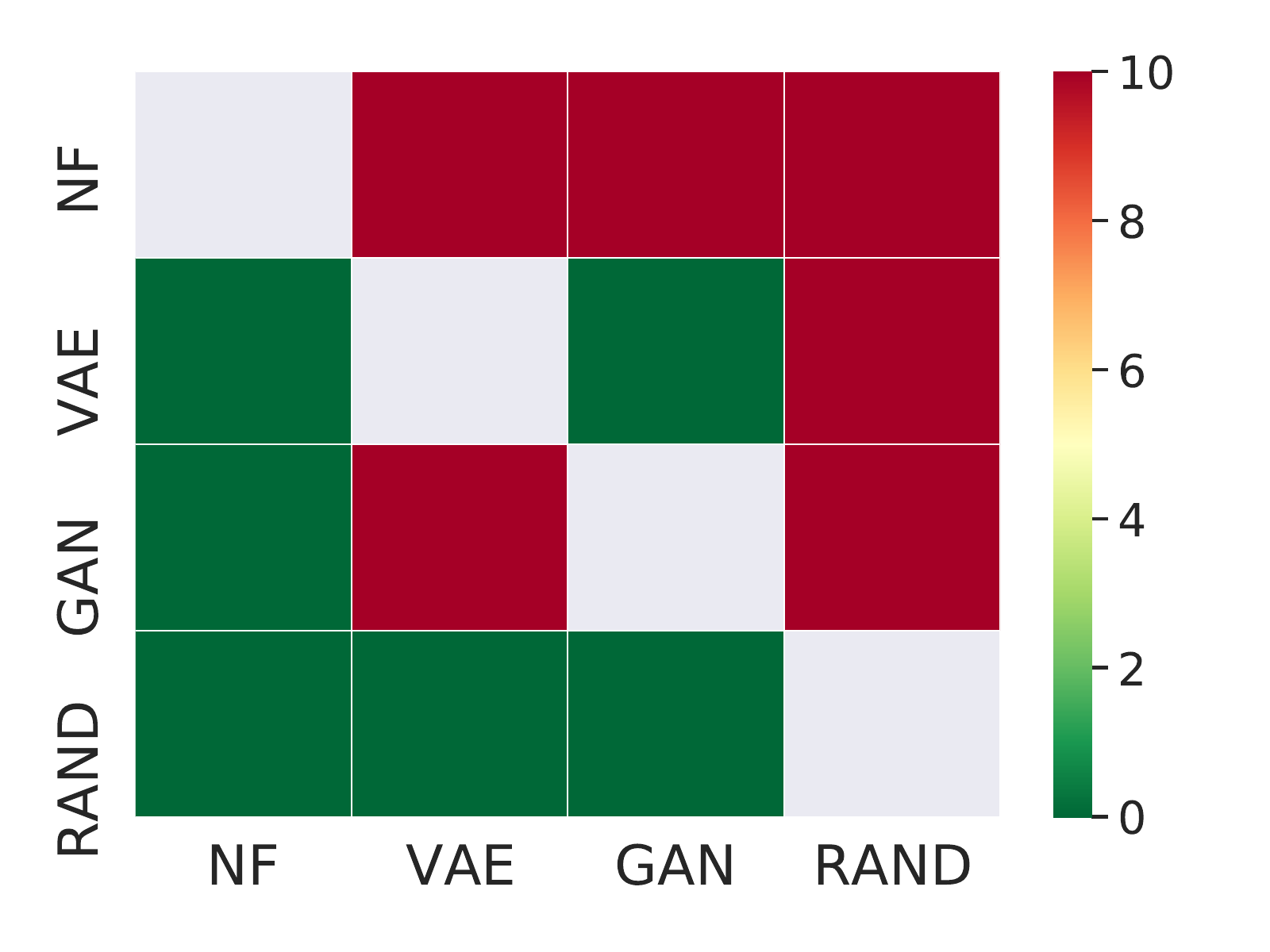}
		\caption{PV ES DM.}
	\end{subfigure}%
	\begin{subfigure}{.25\textwidth}
		\centering
		\includegraphics[width=\linewidth]{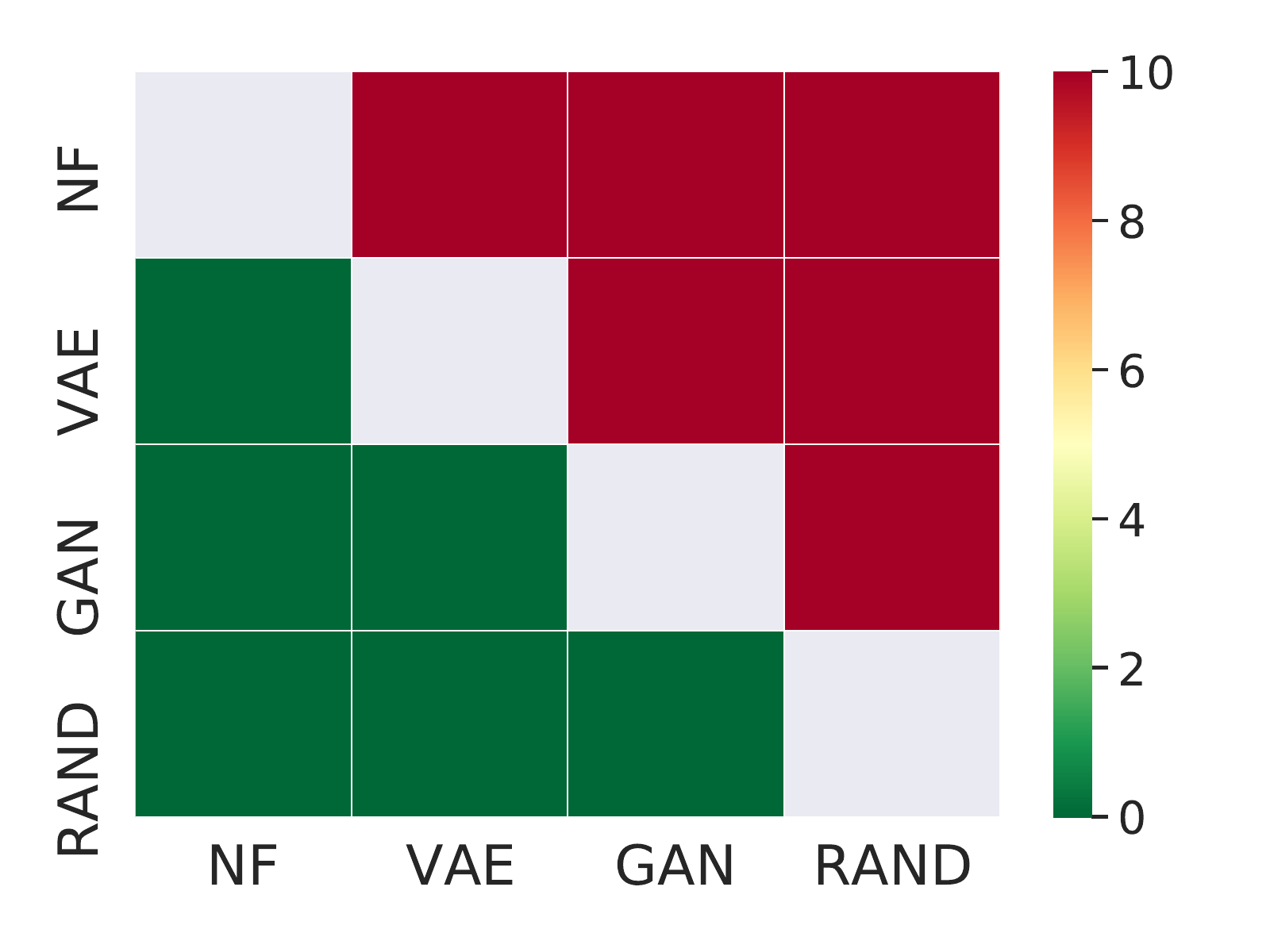}
		\caption{Load ES DM.}
	\end{subfigure}
	\begin{subfigure}{.25\textwidth}
		\centering
		\includegraphics[width=\linewidth]{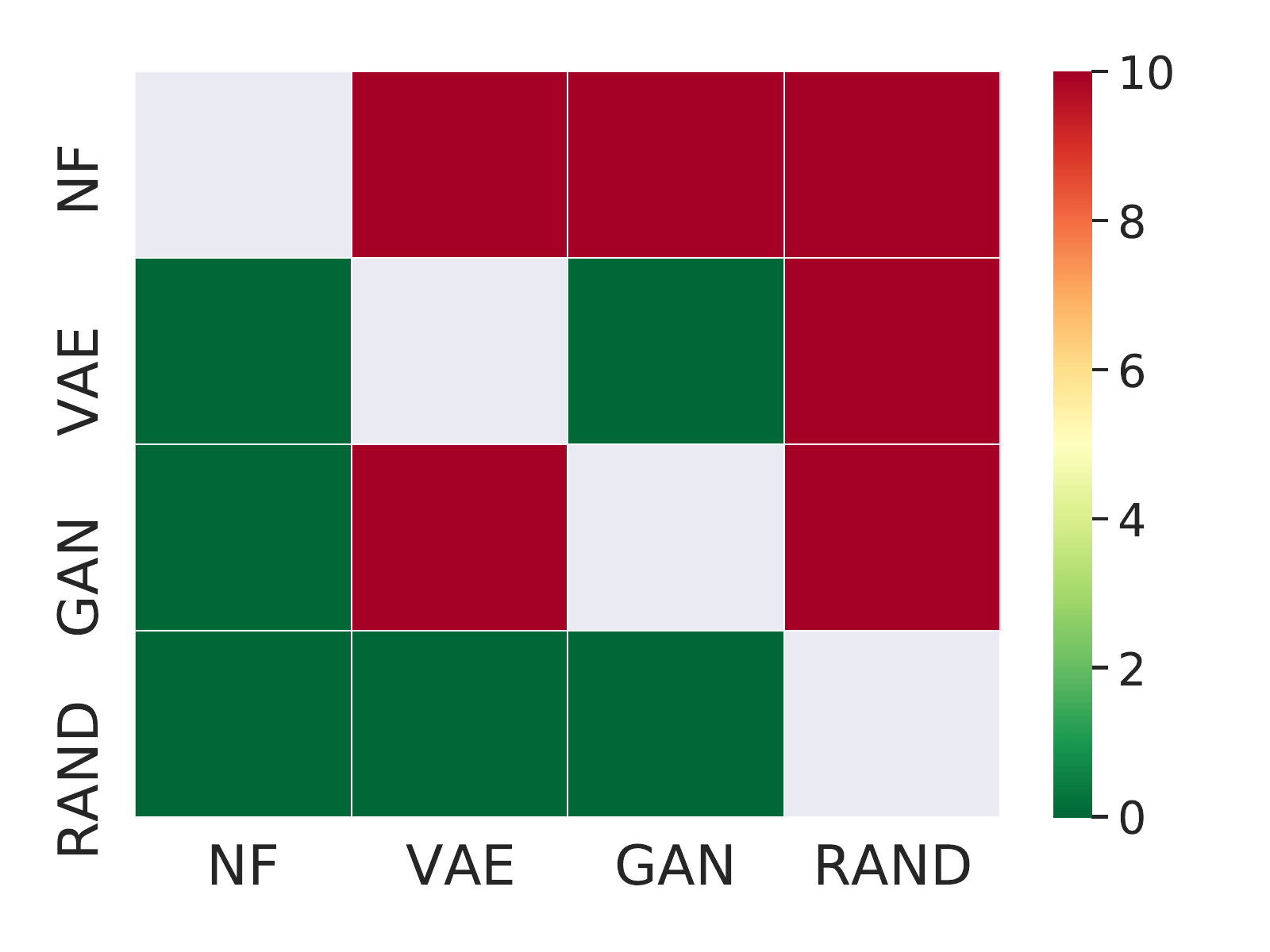}
		\caption{PV VS DM.}
	\end{subfigure}%
	\begin{subfigure}{.25\textwidth}
		\centering
		\includegraphics[width=\linewidth]{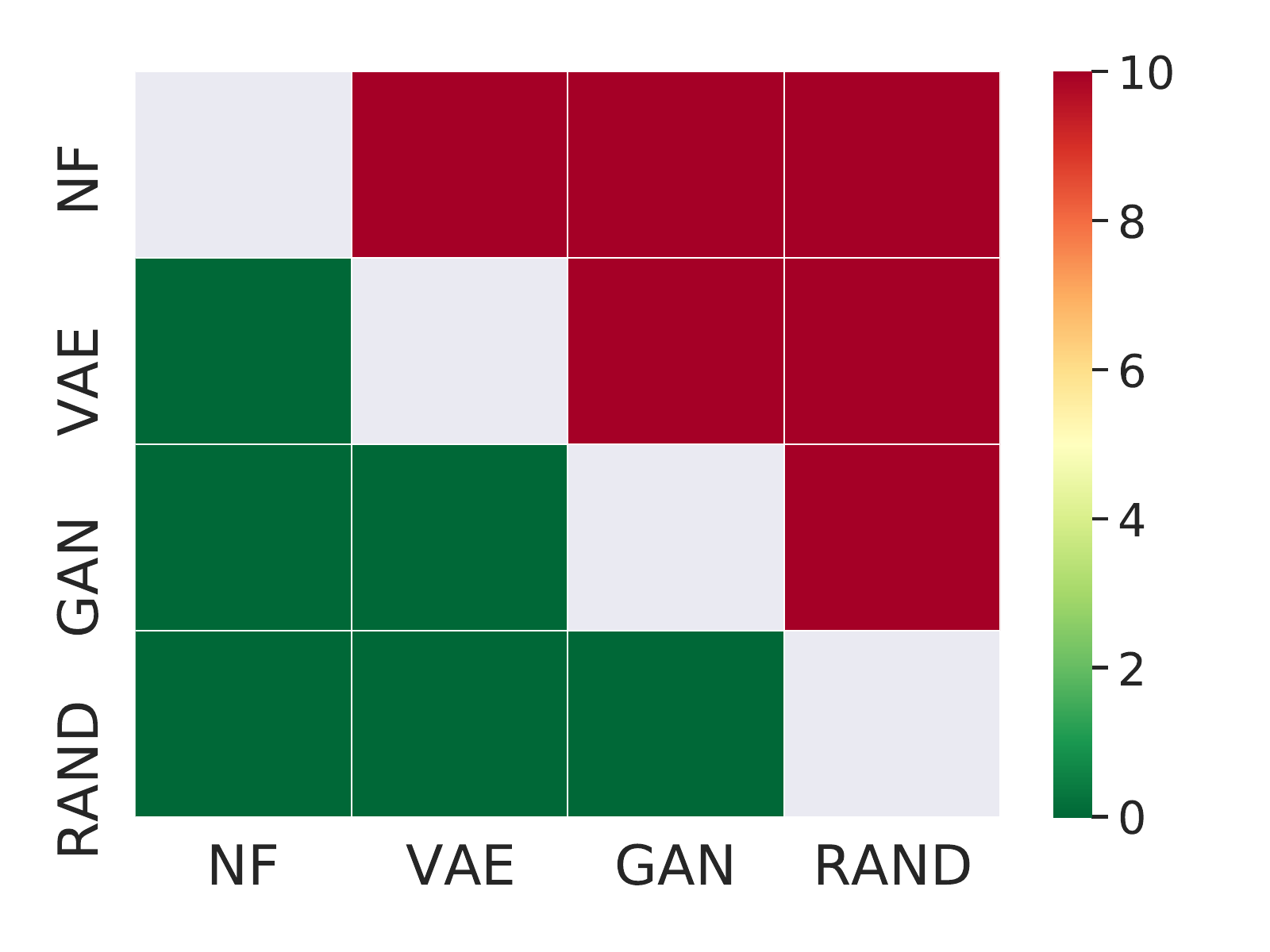}
		\caption{Load VS DM.}
	\end{subfigure}
	\caption{PV and load tracks Diebold-Mariano tests. \\
	The Diebold-Mariano tests of the CRPS, QS, ES, and VS demonstrate the NF outperforms both the VAE and GAN. Note: the GAN outperforms the VAE for both the ES and VS for the PV track. However, the VAE outperforms the GAN on this dataset for both the CRPS and QS.}
	\label{fig:DM-test-pv-load}
\end{figure}
\begin{figure}[htb]
	\centering
	\begin{subfigure}{.25\textwidth}
		\centering
		\includegraphics[width=\linewidth]{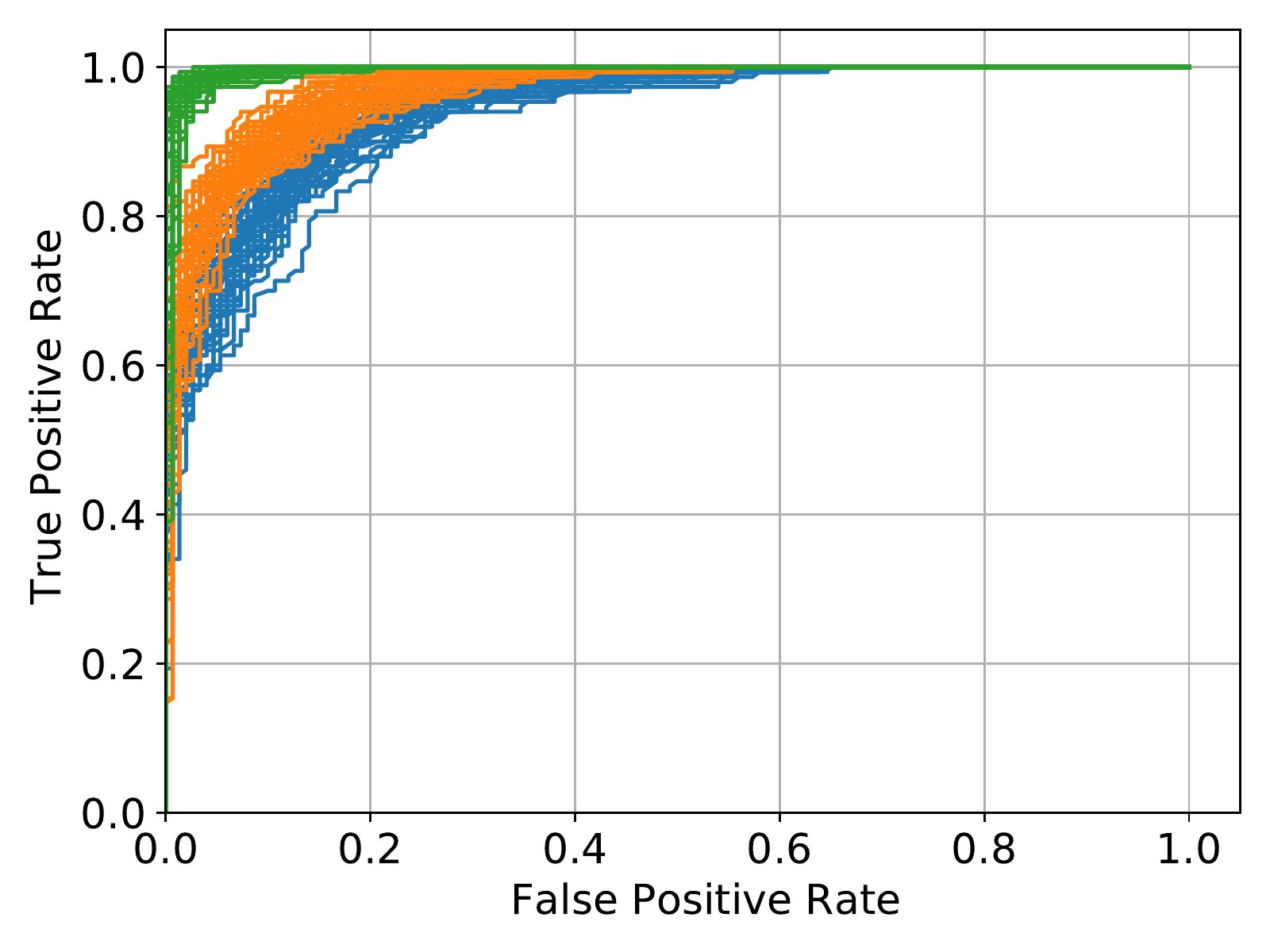}
		\caption{ROC PV.}
	\end{subfigure}%
		\begin{subfigure}{.25\textwidth}
		\centering
		\includegraphics[width=\linewidth]{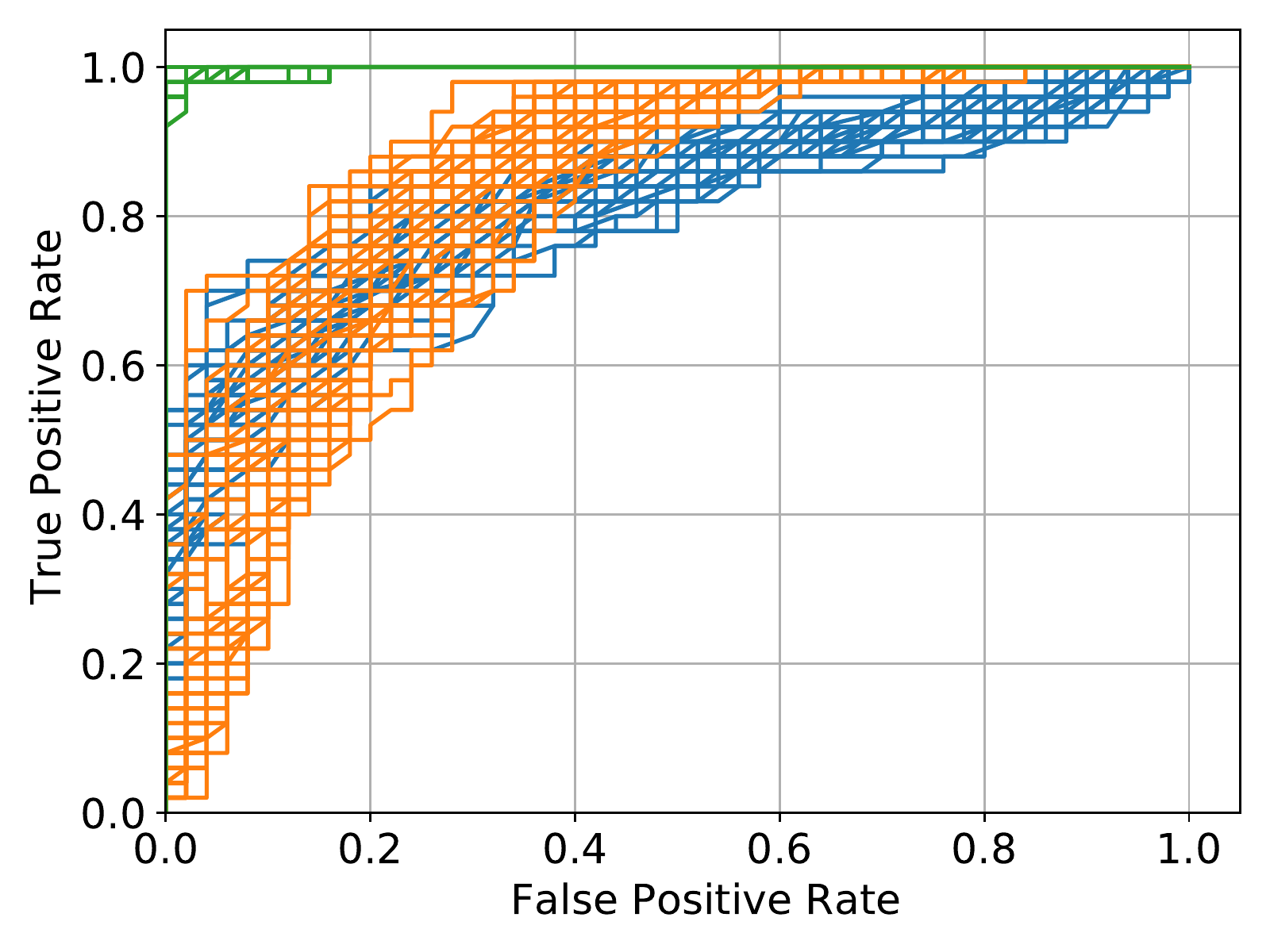}
		\caption{ROC load.}
	\end{subfigure}
	\caption{Classifier-based metric for both the PV and load tracks. \\
The NF (blue) is the best to fake the classifier, followed by the VAE (orange), and the GAN (green).}
	\label{fig:classifier-pv-load}
\end{figure}
\begin{figure}[tb]
	\centering
		\begin{subfigure}{.45\textwidth}
		\centering
		\includegraphics[width=\linewidth]{figs/legend_scenarios}
	\end{subfigure}
	\begin{subfigure}{.25\textwidth}
		\centering
		\includegraphics[width=\linewidth]{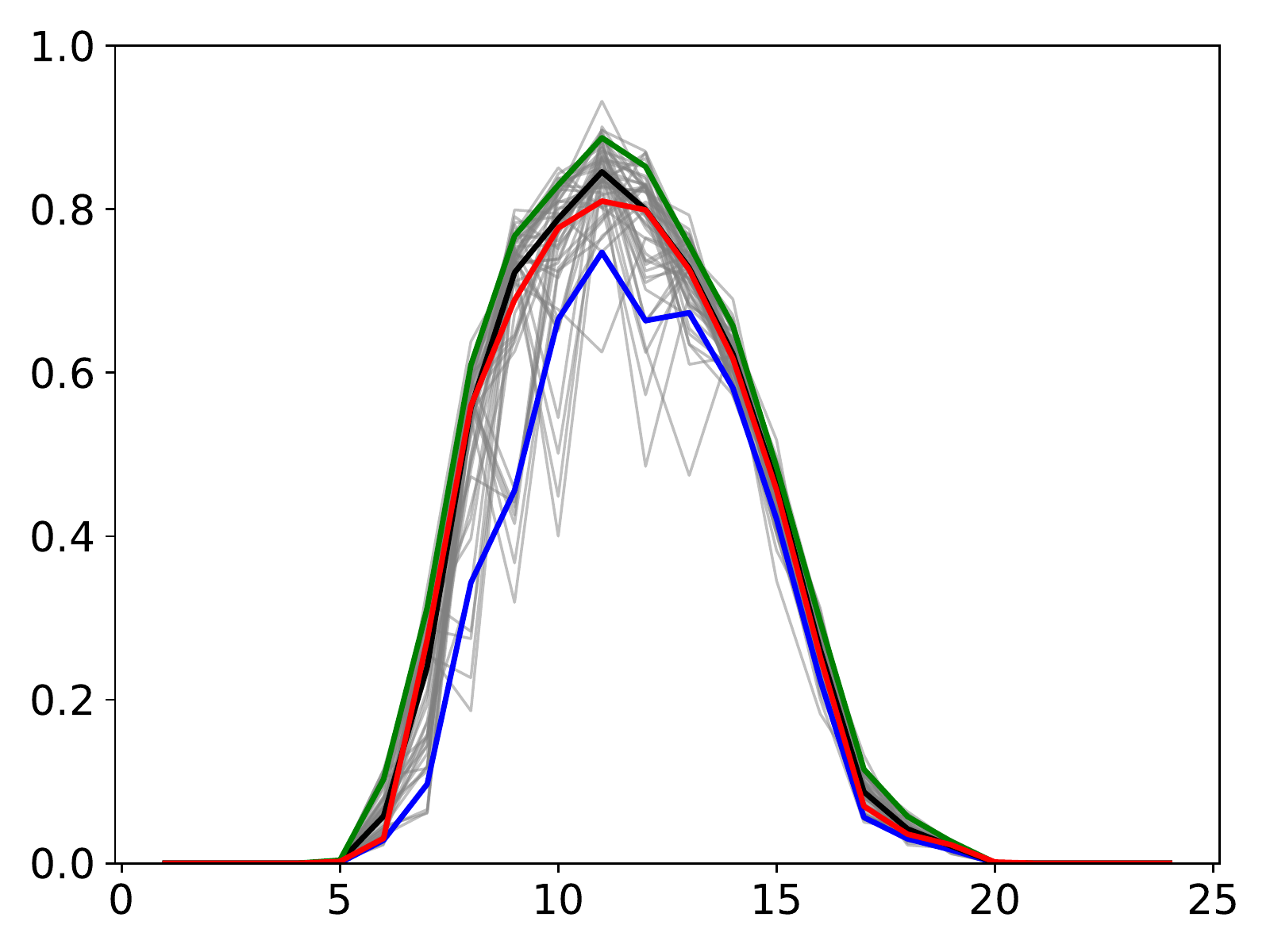}
		\caption{NF.}
	\end{subfigure}%
	\begin{subfigure}{.25\textwidth}
		\centering
		\includegraphics[width=\linewidth]{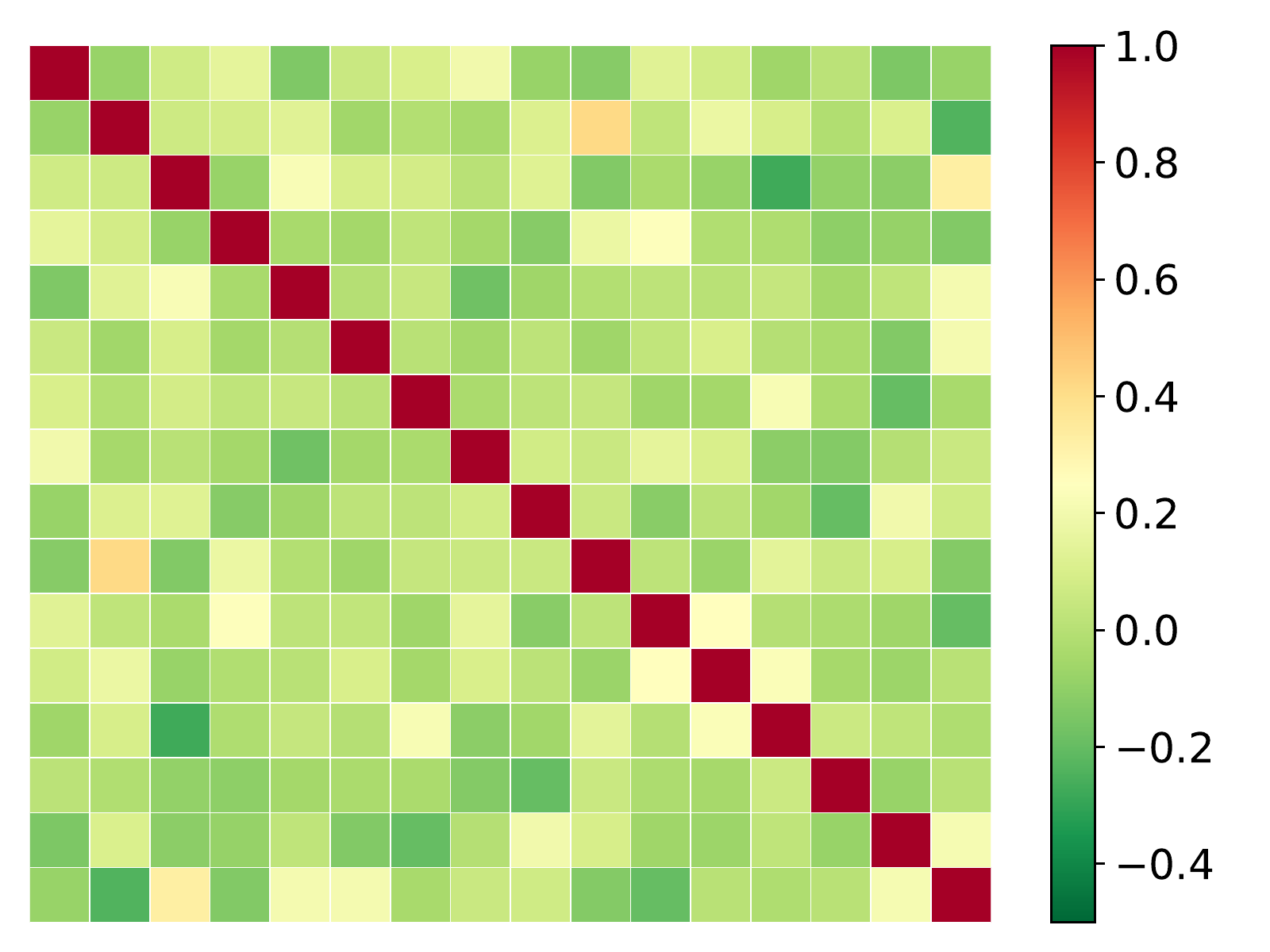}
		\caption{NF.}
	\end{subfigure}
	\begin{subfigure}{.25\textwidth}
		\centering
		\includegraphics[width=\linewidth]{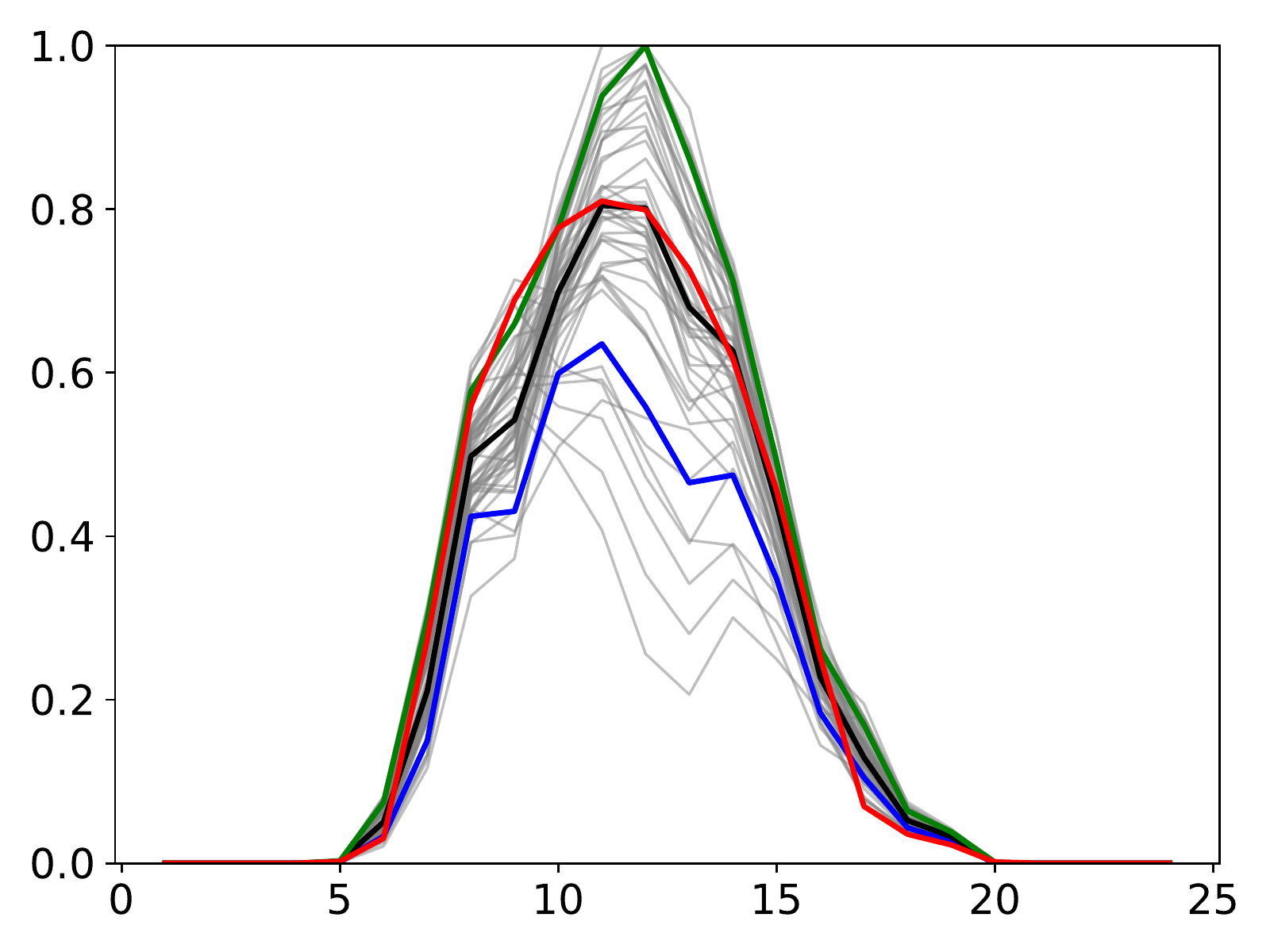}
		\caption{GAN.}
	\end{subfigure}%
	\begin{subfigure}{.25\textwidth}
		\centering
		\includegraphics[width=\linewidth]{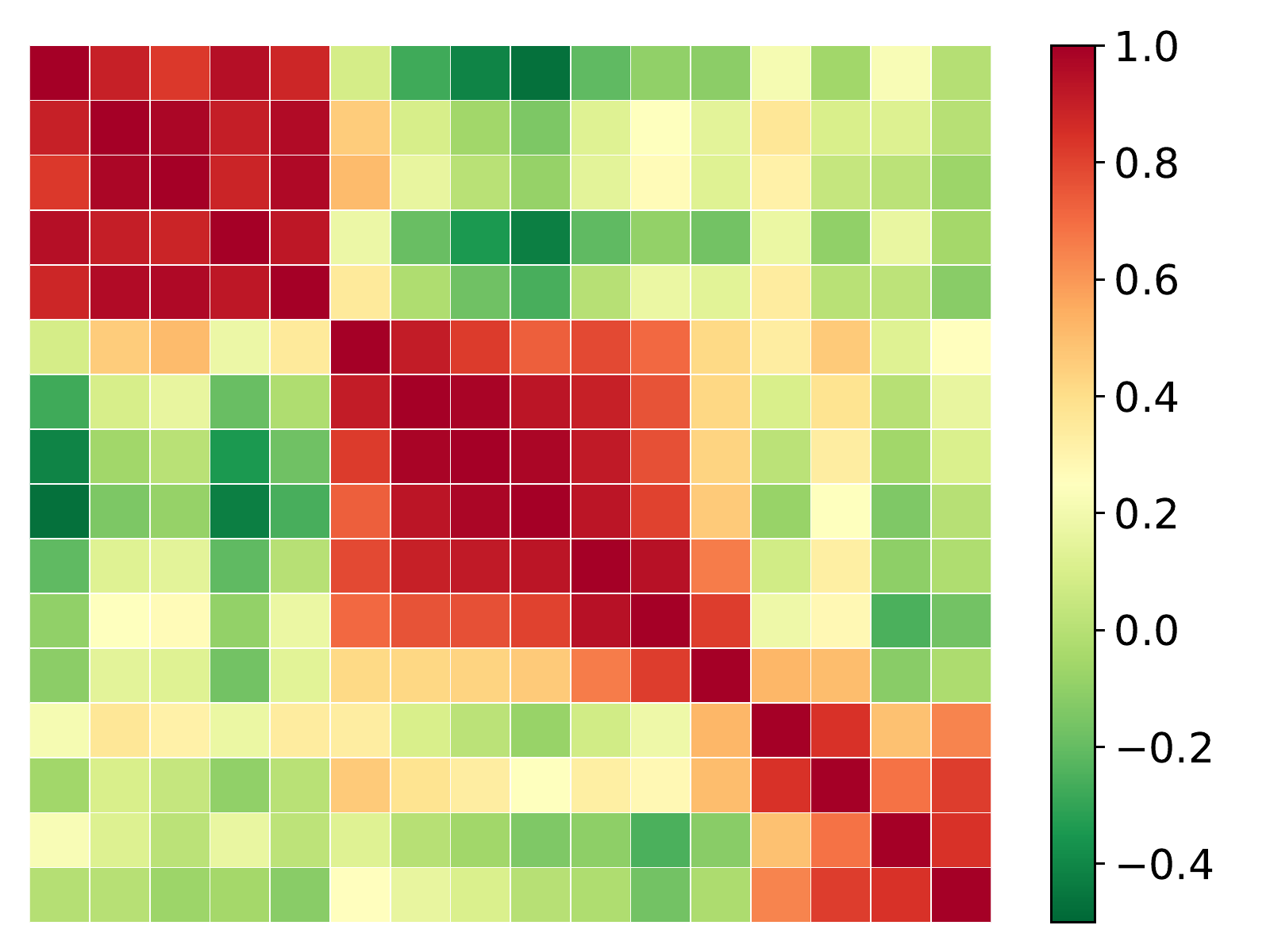}
		\caption{GAN.}
	\end{subfigure}
	\begin{subfigure}{.25\textwidth}
		\centering
		\includegraphics[width=\linewidth]{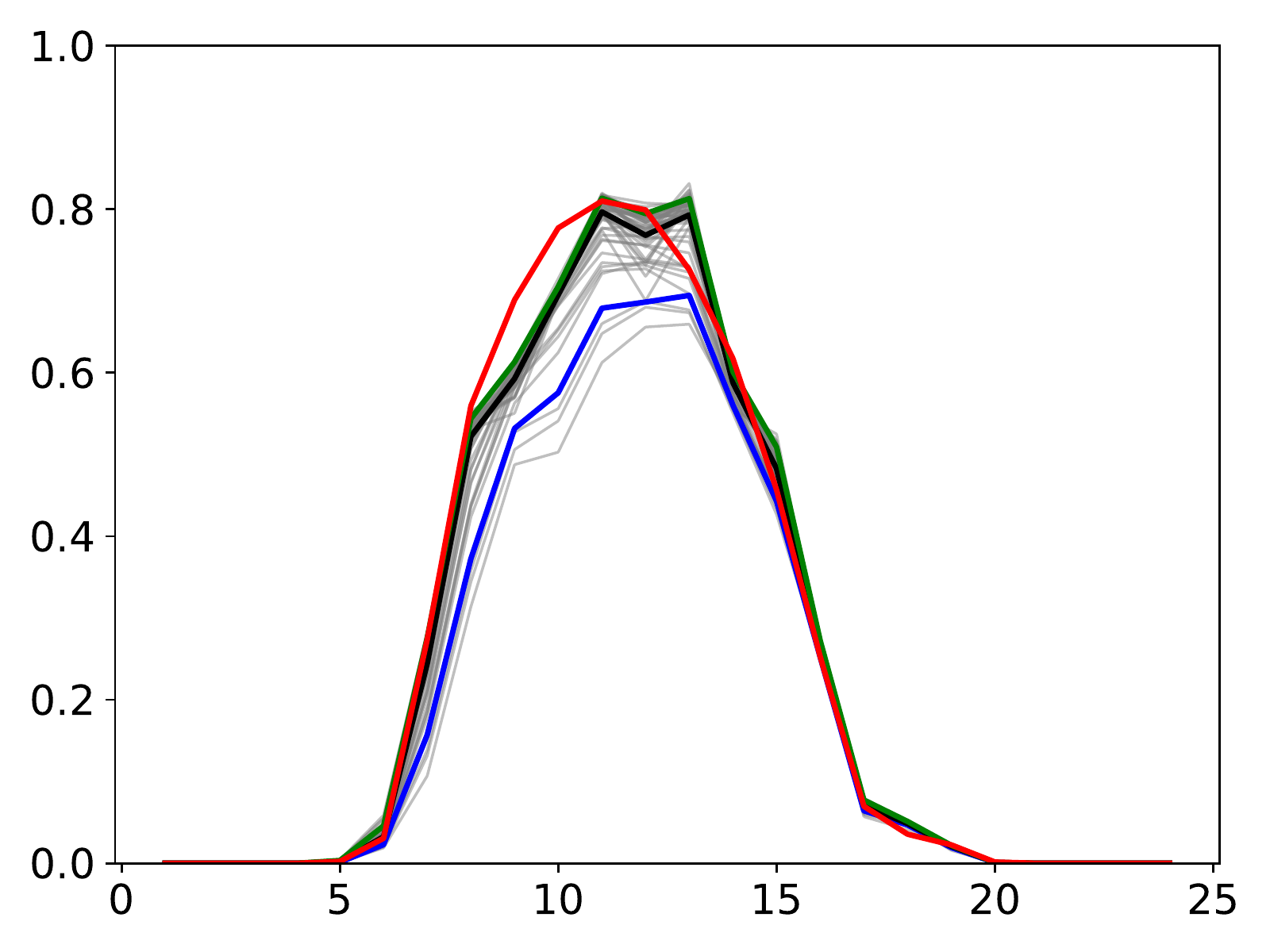}
		\caption{VAE.}
	\end{subfigure}%
	\begin{subfigure}{.25\textwidth}
		\centering
		\includegraphics[width=\linewidth]{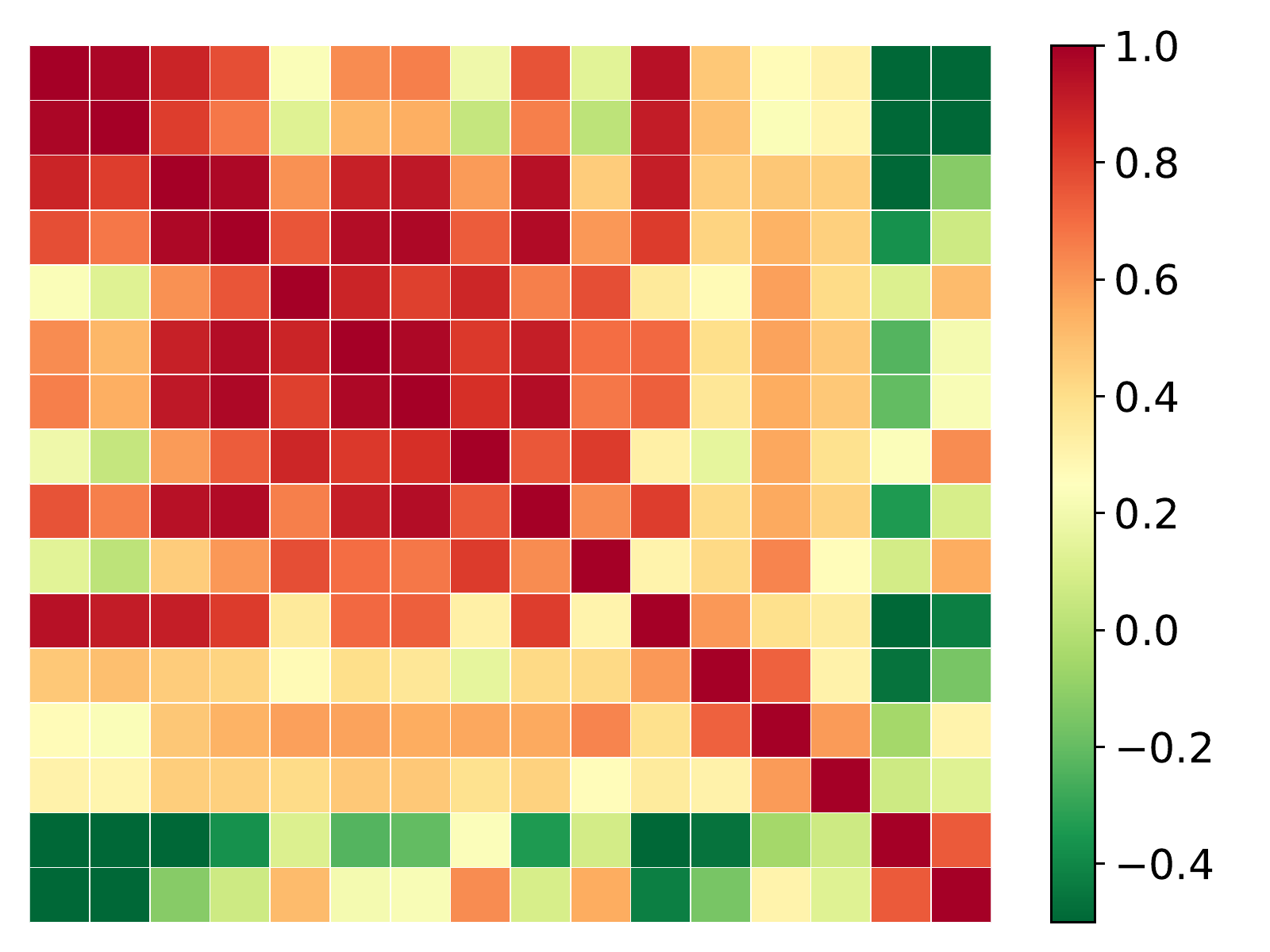}
		\caption{VAE.}
	\end{subfigure}
	\caption{PV scenarios shape comparison and analysis. \\
	Left part (a) NF, (c) GAN, and (e) VAE: 50 PV scenarios (grey) of a randomly selected day of the testing set along with the 10 \% (blue), 50 \% (black), and 90 \% (green) quantiles, and the observations (red). Right part (b) NF, (d) GAN, and (f) VAE: the corresponding Pearson time correlation matrices of these scenarios with the time periods as rows and columns.
	Similar to wind power and load scenarios, NF tends to exhibit no time correlation between scenarios. In contrast, the VAE and GAN tend to be partially time-correlated over a few time periods. 
	}
	\label{fig:pv_scenarios}
\end{figure}
\begin{figure}[tb]
\centering
	\begin{subfigure}{.45\textwidth}
		\centering
		\includegraphics[width=\linewidth]{figs/legend_scenarios}
	\end{subfigure}
	\begin{subfigure}{.25\textwidth}
		\centering
		\includegraphics[width=\linewidth]{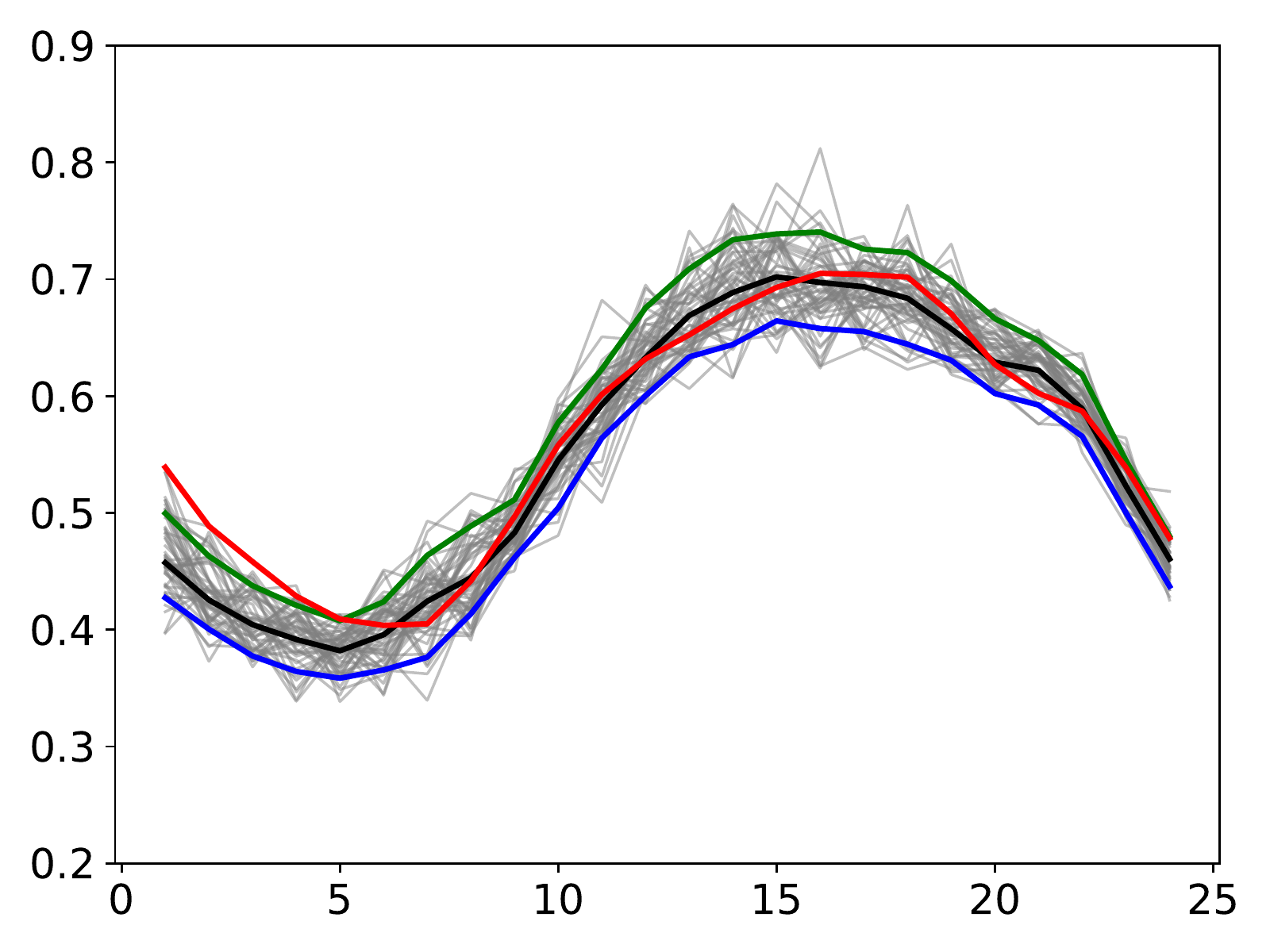}
		\caption{NF.}
	\end{subfigure}%
	\begin{subfigure}{.25\textwidth}
		\centering
		\includegraphics[width=\linewidth]{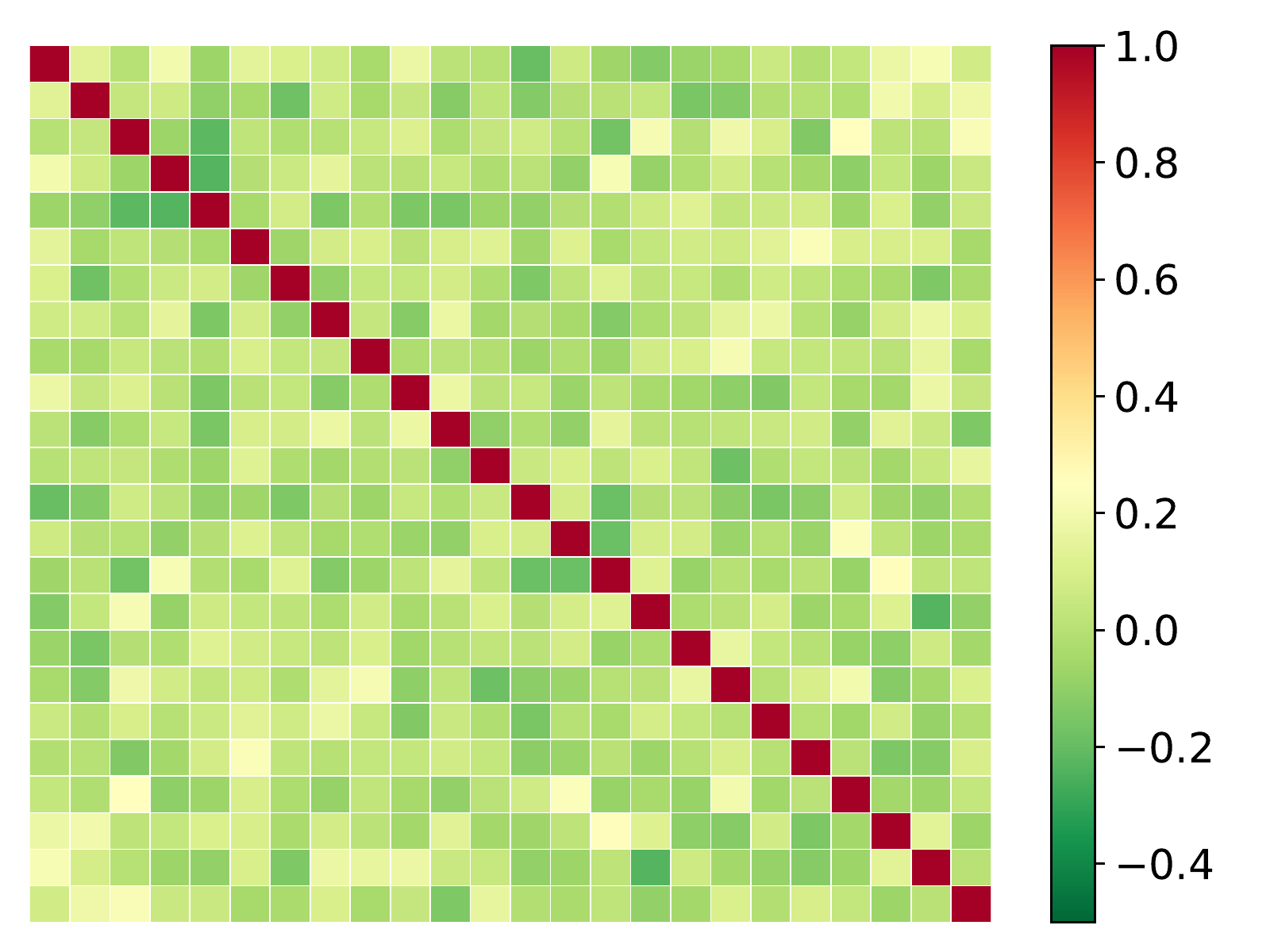}
		\caption{NF.}
	\end{subfigure}
	\begin{subfigure}{.25\textwidth}
		\centering
		\includegraphics[width=\linewidth]{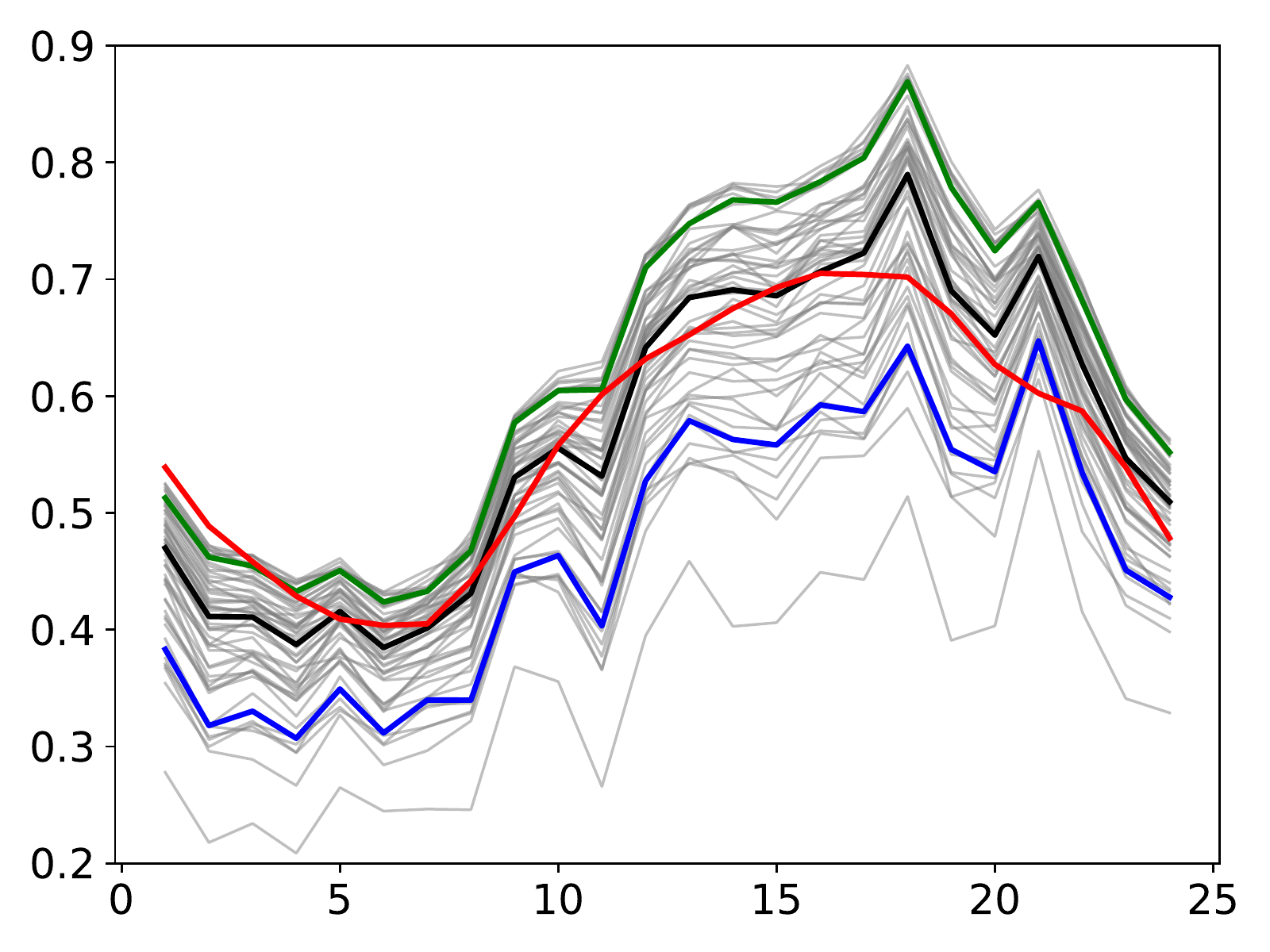}
		\caption{GAN.}
	\end{subfigure}%
	\begin{subfigure}{.25\textwidth}
		\centering
		\includegraphics[width=\linewidth]{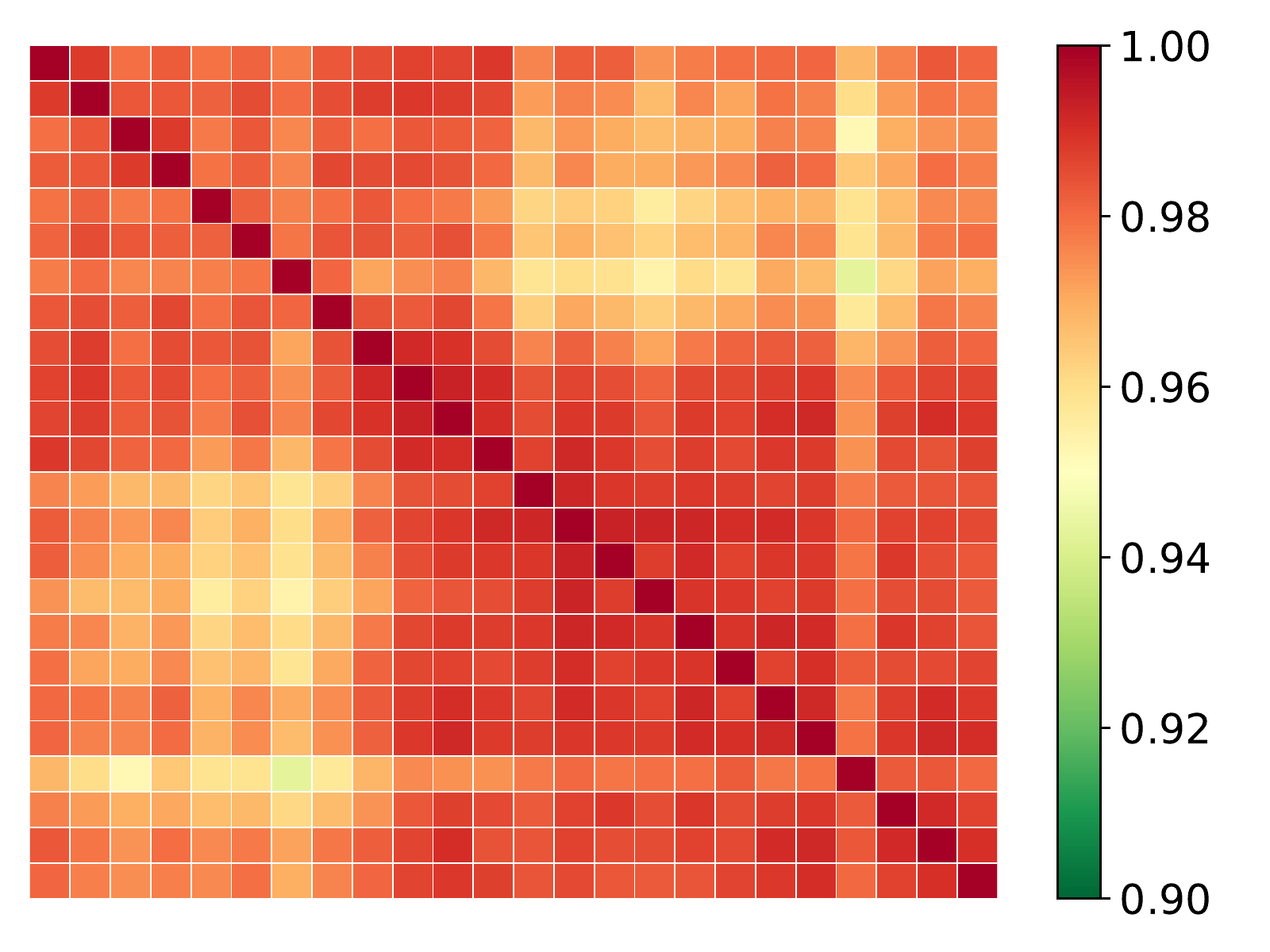}
		\caption{GAN.}
	\end{subfigure}
	\begin{subfigure}{.25\textwidth}
		\centering
		\includegraphics[width=\linewidth]{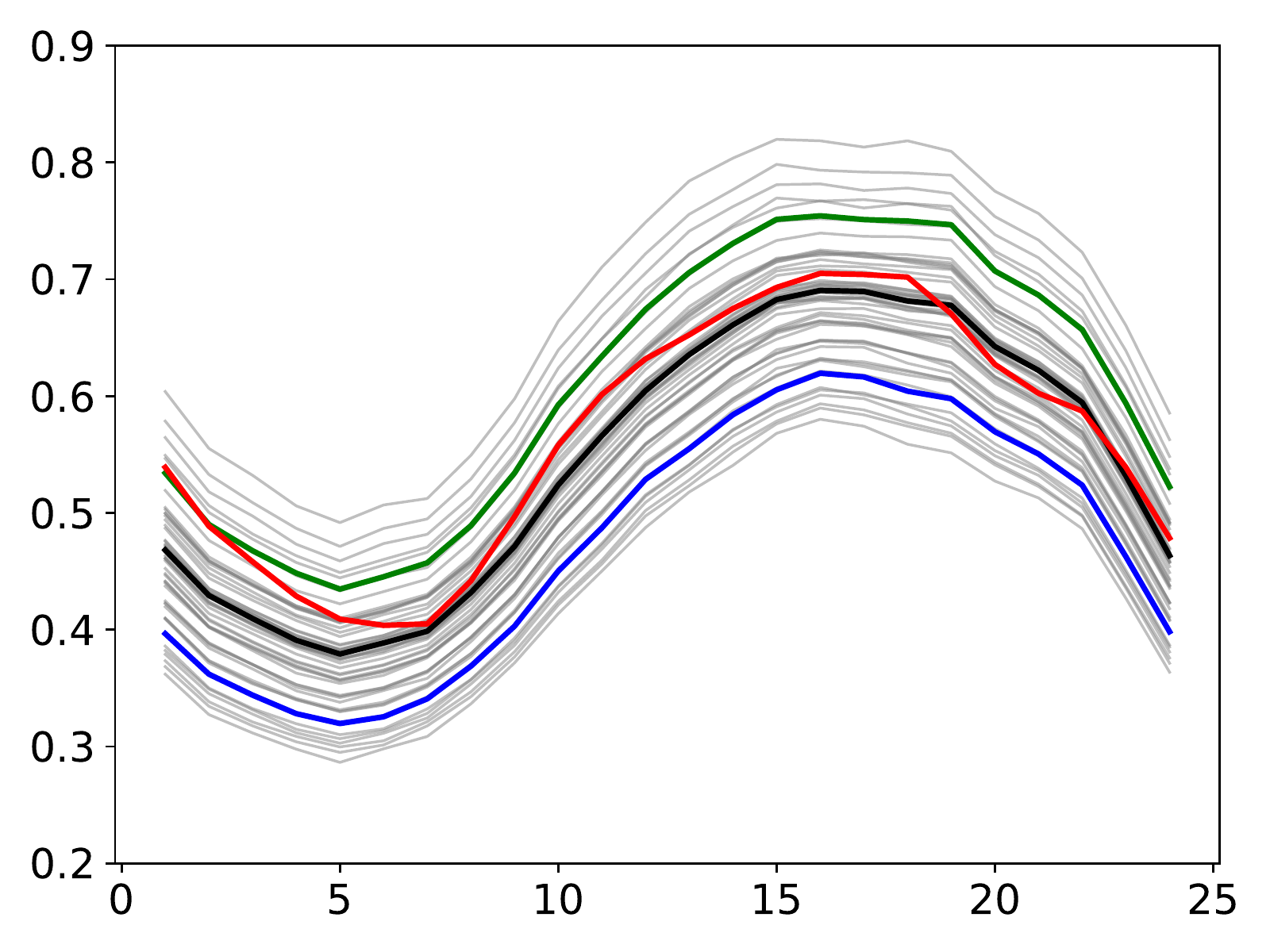}
		\caption{VAE.}
	\end{subfigure}%
	\begin{subfigure}{.25\textwidth}
		\centering
		\includegraphics[width=\linewidth]{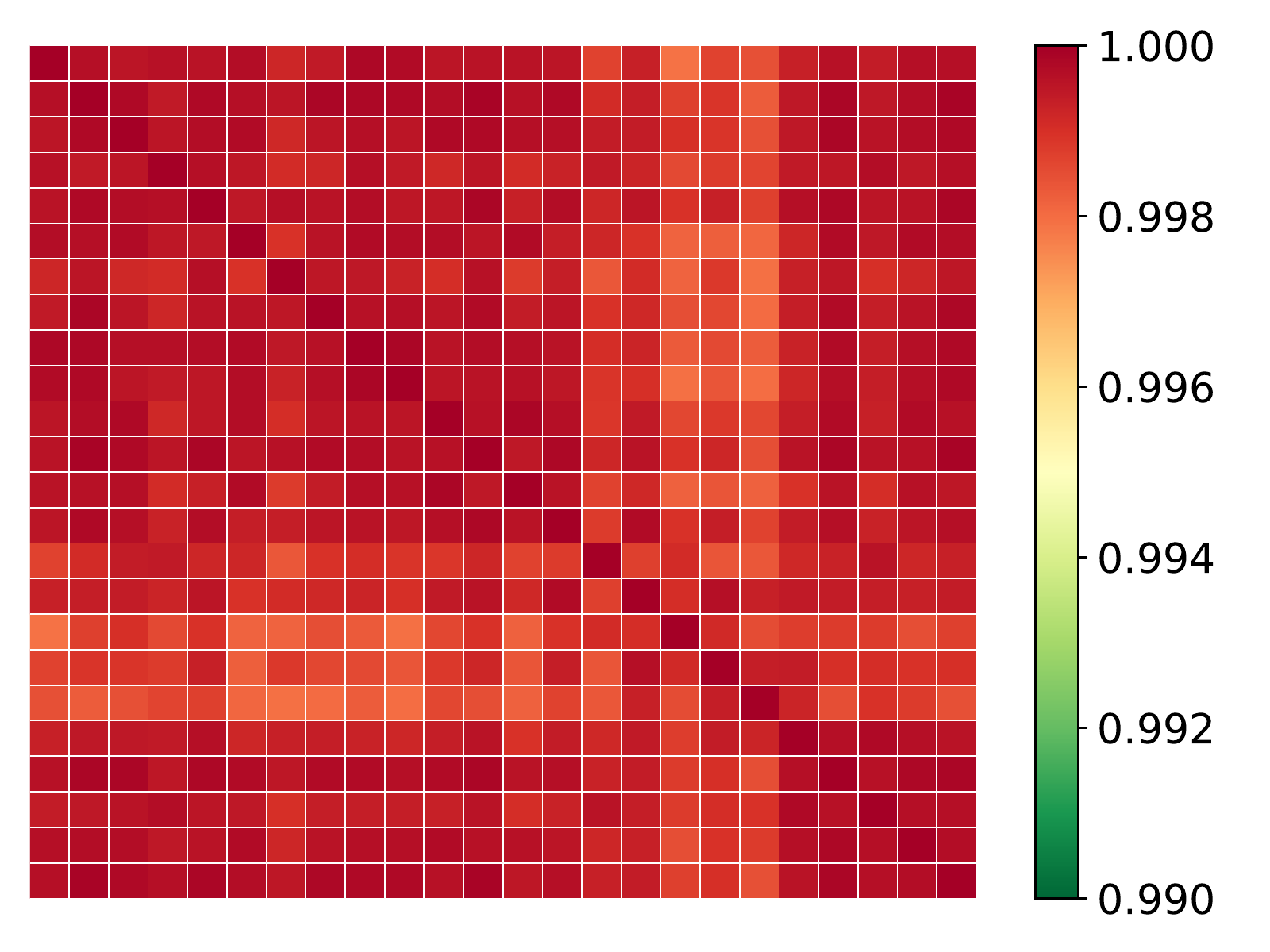}
		\caption{VAE.}
	\end{subfigure}
	\caption{Load scenarios shape comparison and analysis. \\
	Left part (a) NF, (c) GAN, and (e) VAE: 50 load scenarios (grey) of a randomly selected day of the testing set along with the 10 \% (blue), 50 \% (black), and 90 \% (green) quantiles, and the observations (red). Right part (b) NF, (d) GAN, and (f) VAE: the corresponding Pearson time correlation matrices of these scenarios with the time periods as rows and columns.
	Similar to PV and wind power scenarios, NF tends to exhibit no time correlation between scenarios. In contrast, the VAE and GAN tend to be highly time-correlated. 
	}
	\label{fig:load_scenarios}
\end{figure}
%
%
\begin{figure}[tb]
	\centering
	\begin{subfigure}{.16\textwidth}
		\centering
		\includegraphics[width=\linewidth]{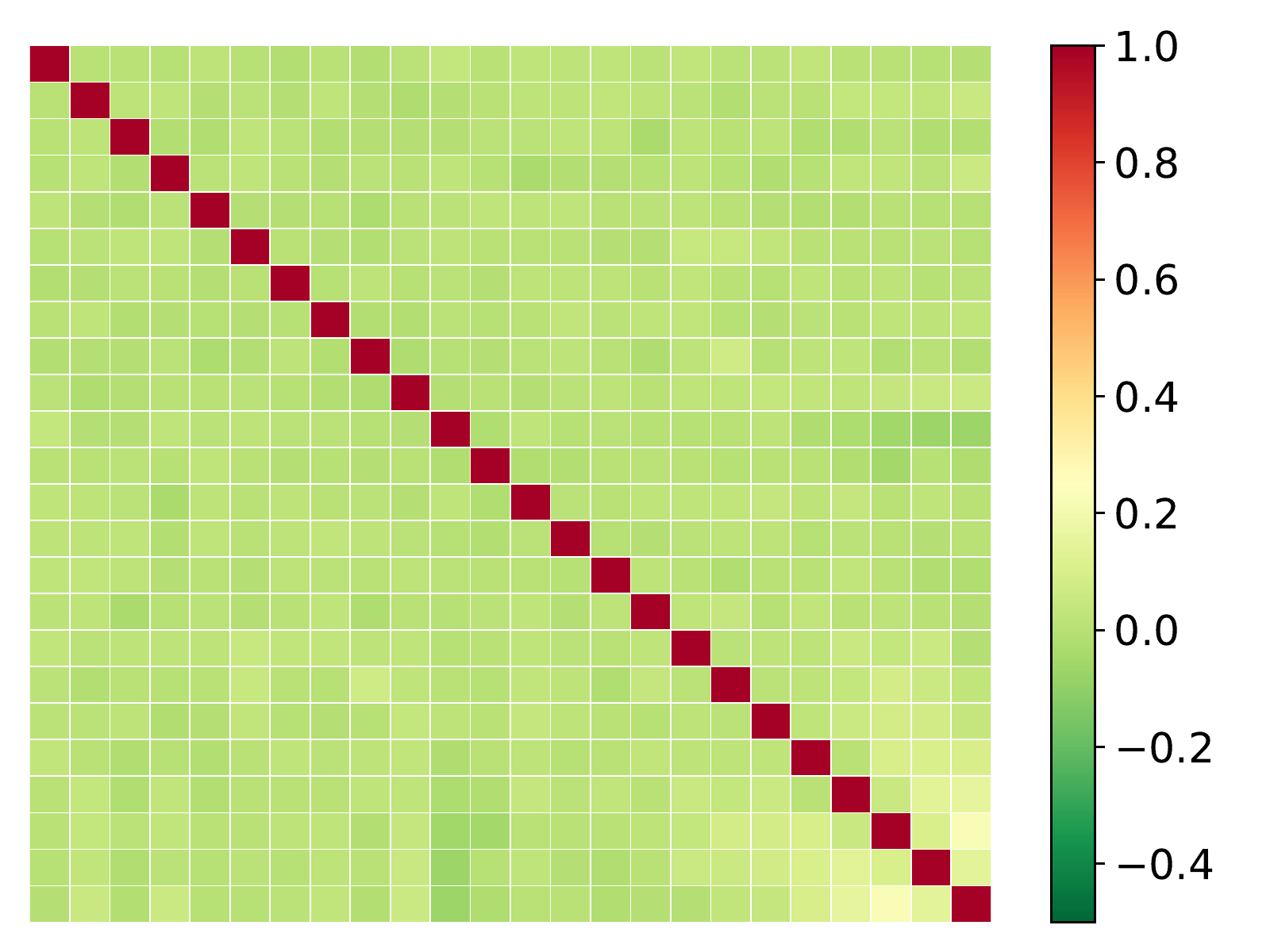}
		\caption{Wind-NF.}
	\end{subfigure}%
	\begin{subfigure}{.16\textwidth}
		\centering
		\includegraphics[width=\linewidth]{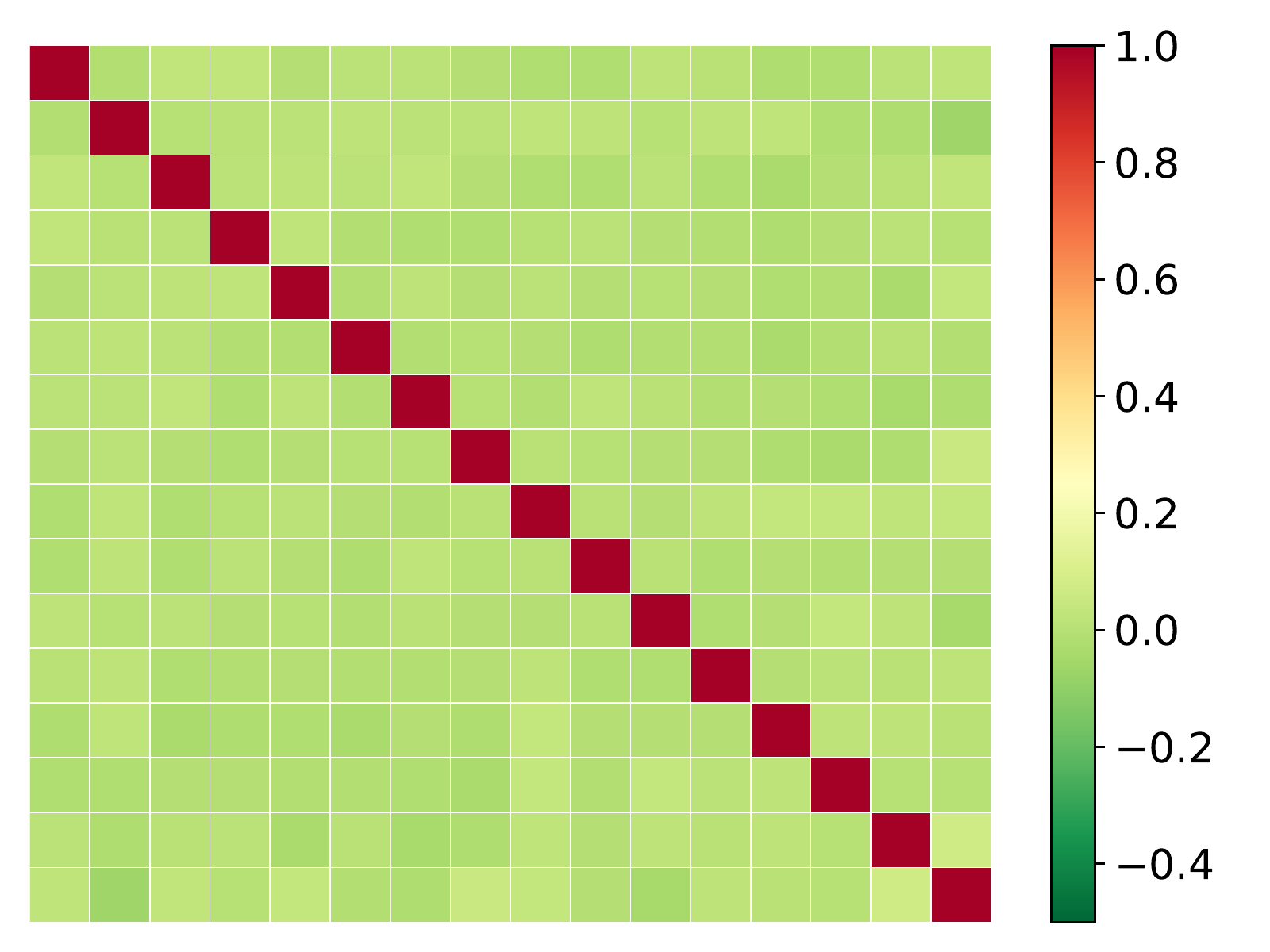}
		\caption{PV-NF.}
	\end{subfigure}%
	\begin{subfigure}{.16\textwidth}
		\centering
		\includegraphics[width=\linewidth]{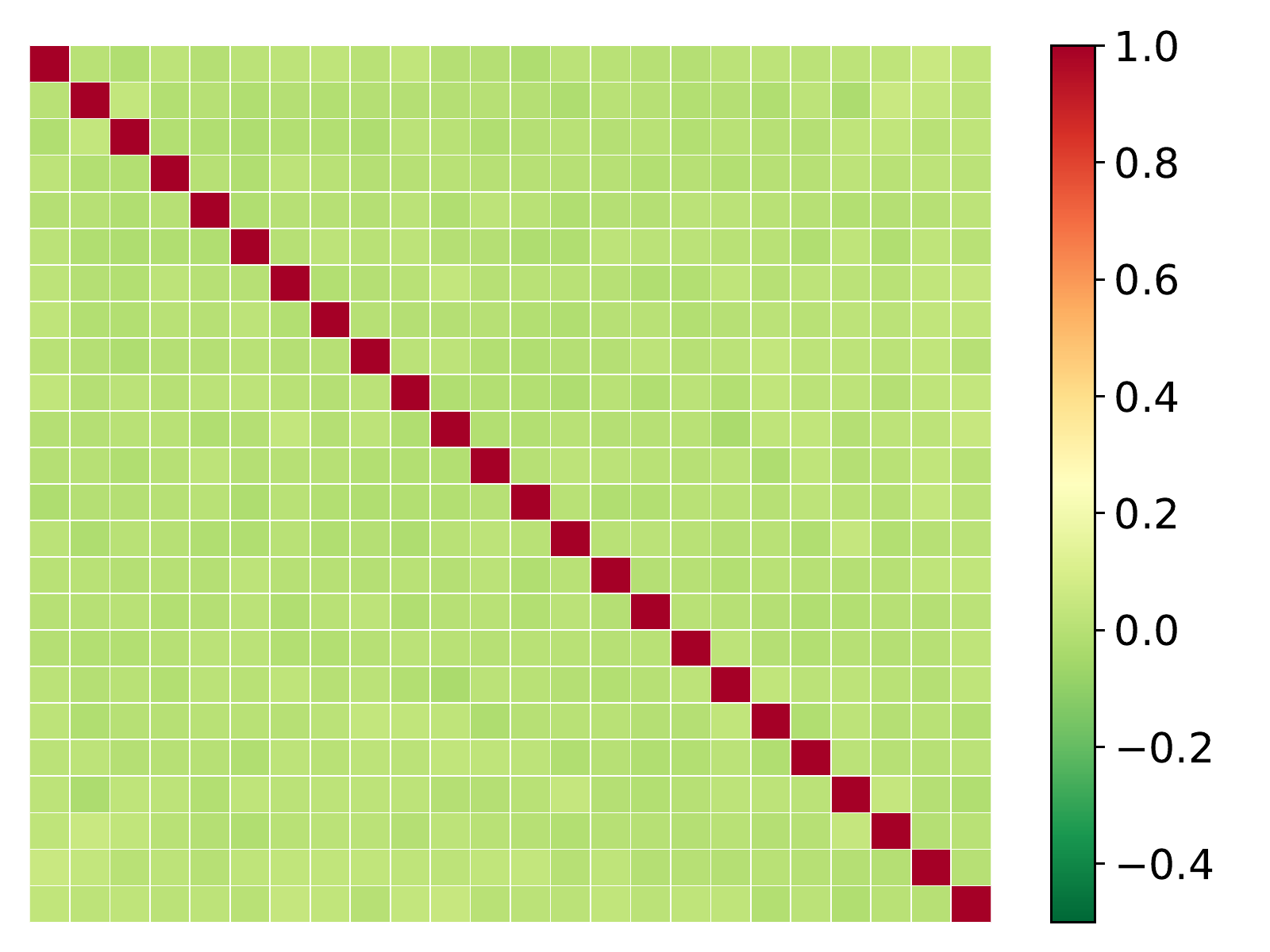}
		\caption{NF-load.}
	\end{subfigure}
	\begin{subfigure}{.16\textwidth}
		\centering
		\includegraphics[width=\linewidth]{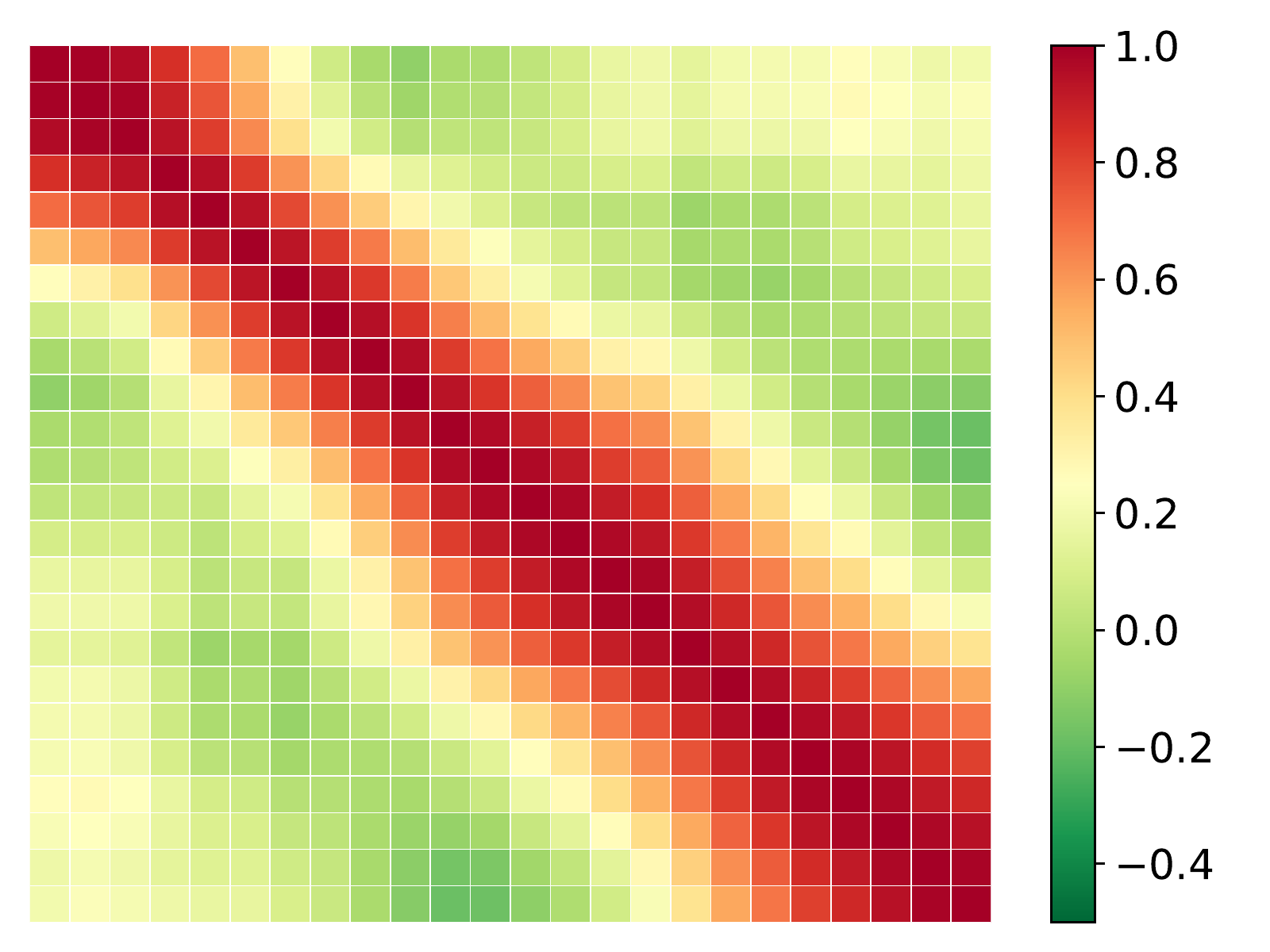}
		\caption{Wind-GAN.}
	\end{subfigure}%
	\begin{subfigure}{.16\textwidth}
		\centering
		\includegraphics[width=\linewidth]{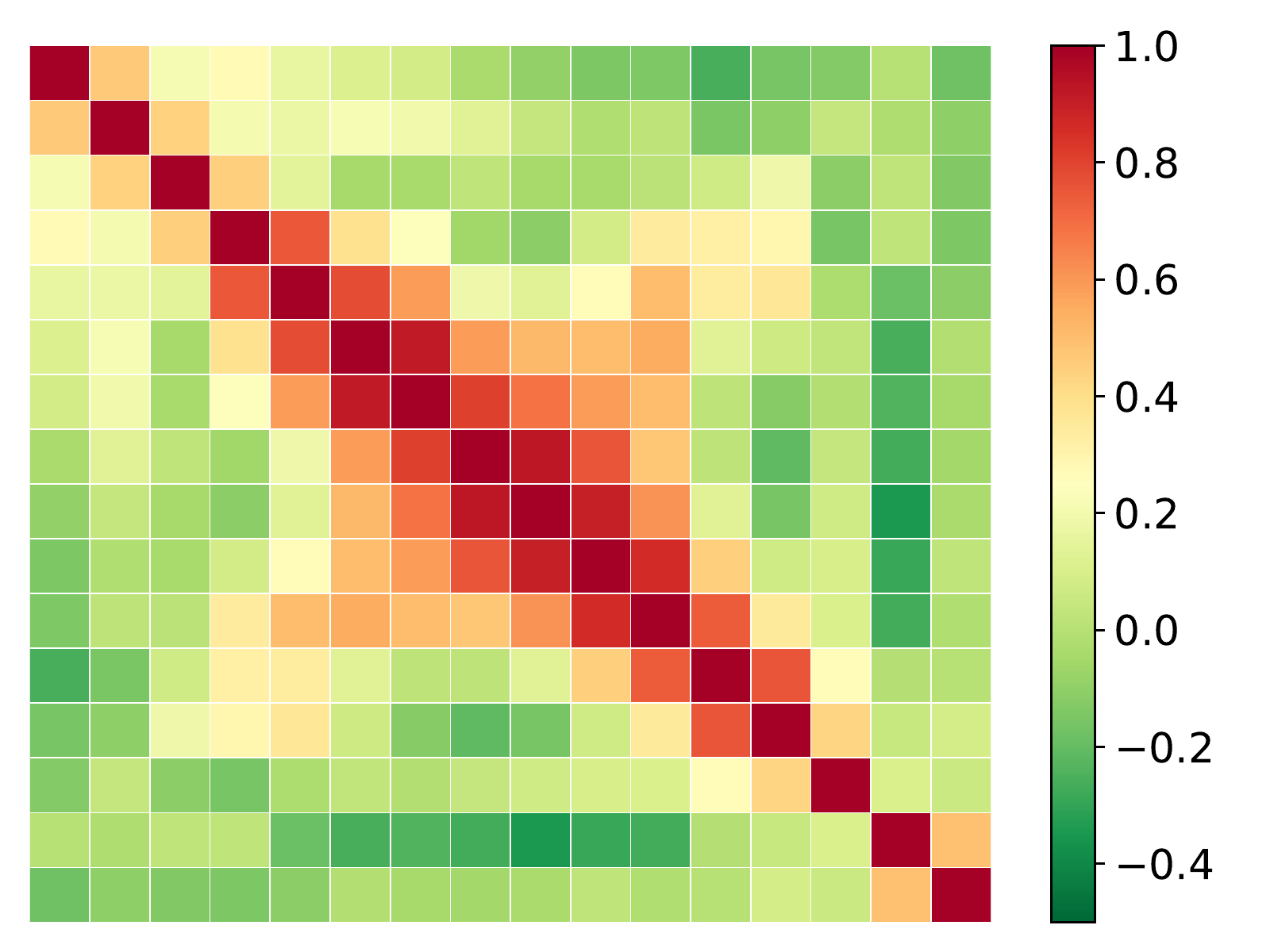}
		\caption{PV-GAN.}
	\end{subfigure}%
	\begin{subfigure}{.16\textwidth}
		\centering
		\includegraphics[width=\linewidth]{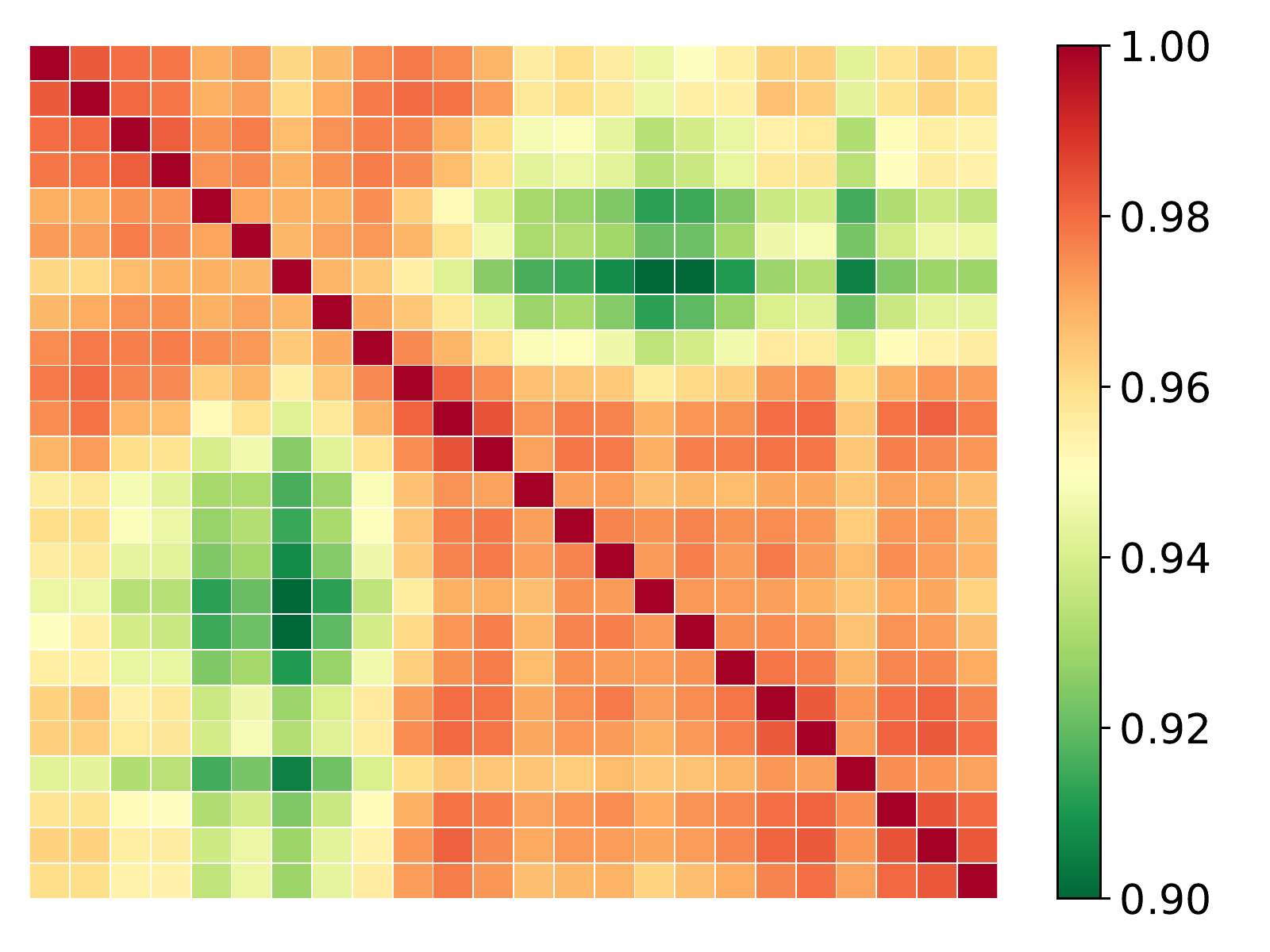}
		\caption{GAN-load.}
	\end{subfigure}
	\begin{subfigure}{.16\textwidth}
		\centering
		\includegraphics[width=\linewidth]{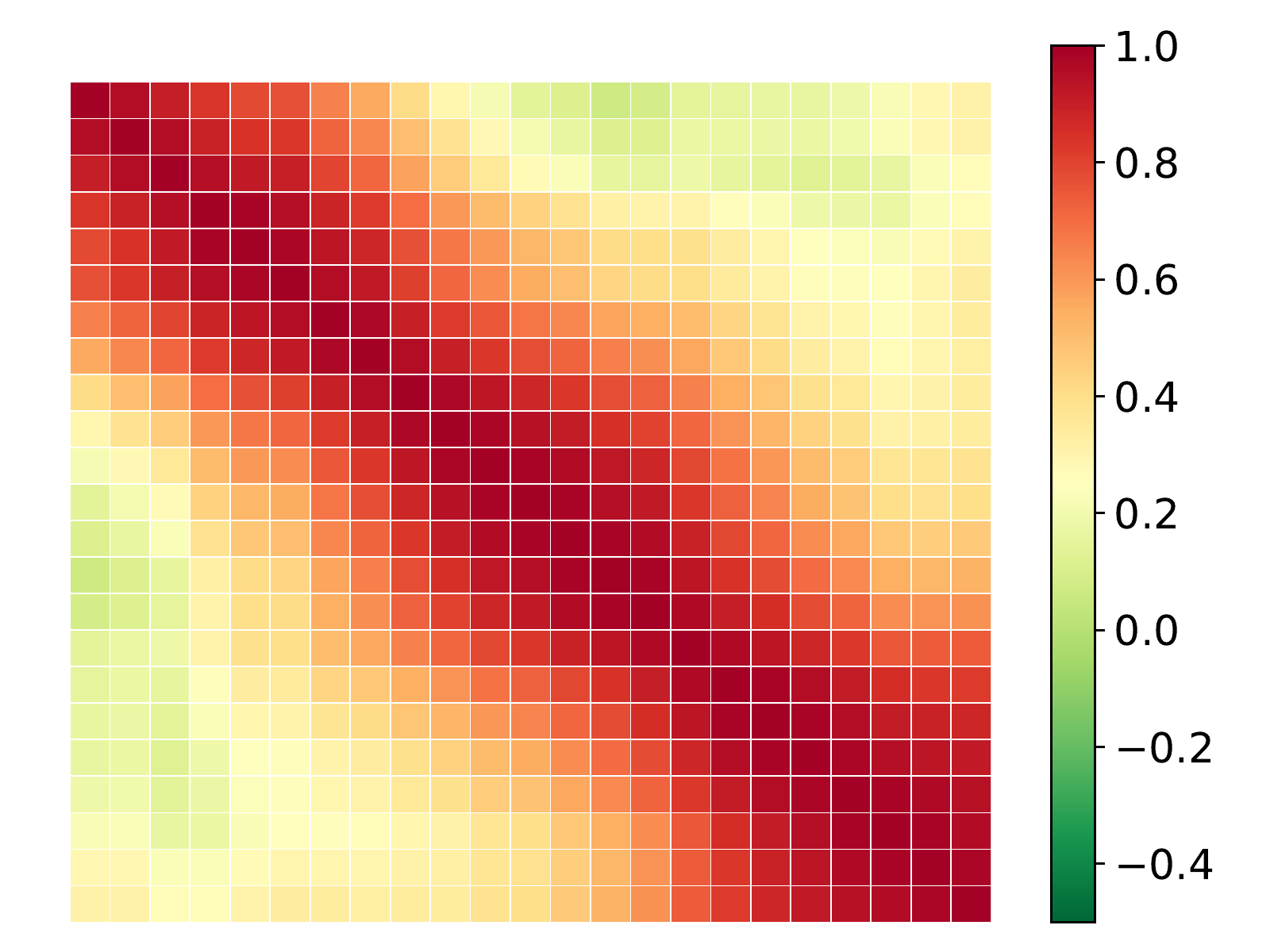}
		\caption{Wind-VAE.}
	\end{subfigure}%
	\begin{subfigure}{.16\textwidth}
		\centering
		\includegraphics[width=\linewidth]{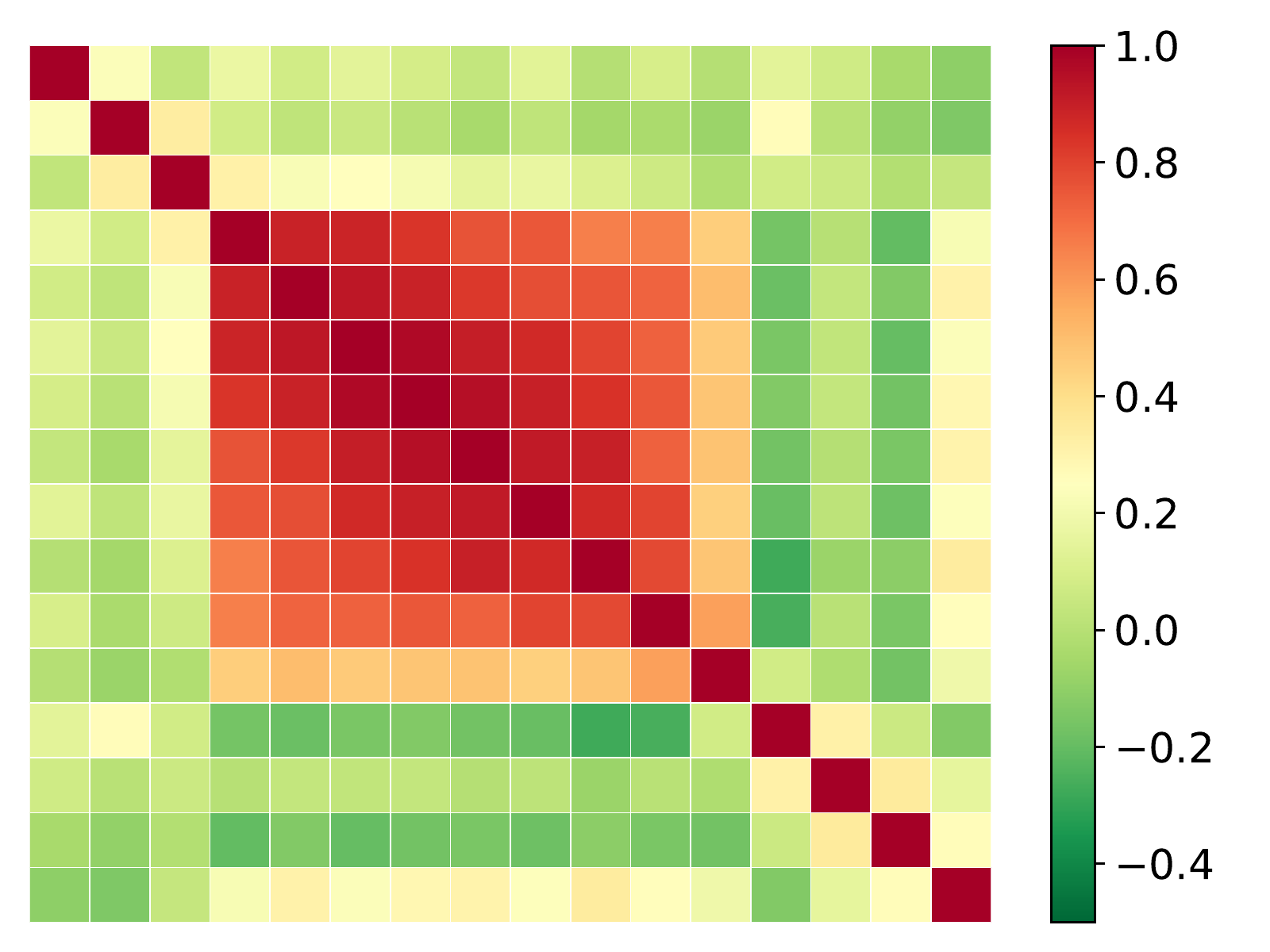}
		\caption{PV-VAE.}
	\end{subfigure}%
	\begin{subfigure}{.16\textwidth}
		\centering
		\includegraphics[width=\linewidth]{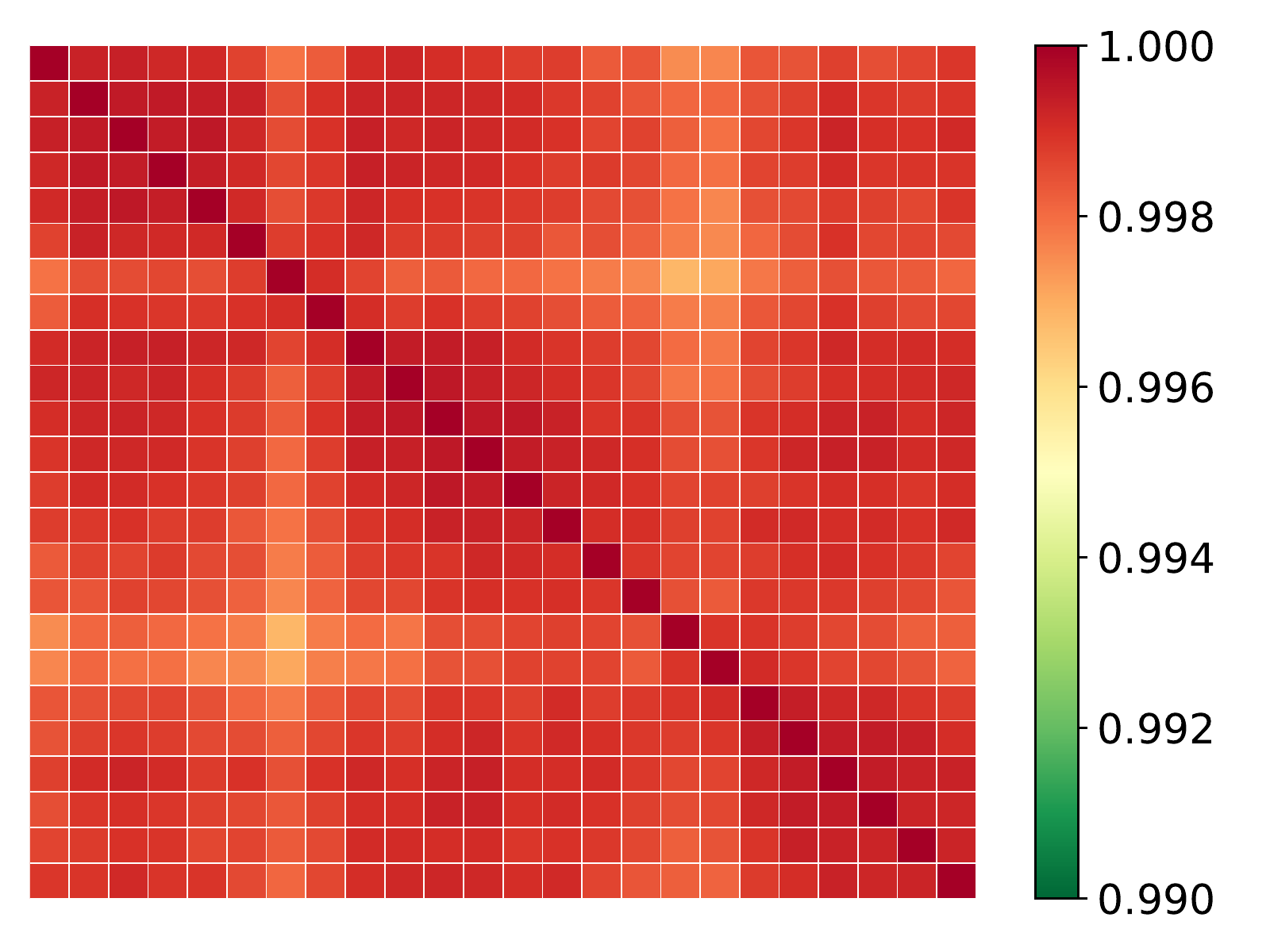}
		\caption{VAE-load.}
	\end{subfigure}
	\caption{Average of the correlation matrices over the testing set for the three datasets. \\
    Left: wind power; center: PV; right:load. The trend in terms of time correlation is observed on each day of the testing set for all the datasets. The NF scenarios are not correlated. In contrast, the VAE and GAN scenarios tend to be time-correlated over a few periods. In particular, the VAE generates highly time-correlated scenarios for the load dataset.
	}
	\label{fig:average-correlation}
\end{figure}

\end{document}